\documentclass[lettersize,journal]{IEEEtran}
\usepackage{amsmath,amsfonts}
\usepackage{algorithmic}
\usepackage{algorithm}
\usepackage{array}
\usepackage{textcomp}
\usepackage{stfloats}
\usepackage{url}
\usepackage{verbatim}
\usepackage{graphicx}
\usepackage{cite}
 \usepackage{multirow}
 \usepackage{setspace}
 \usepackage{booktabs}
 
\usepackage{pifont}
\usepackage{caption,subcaption} 
 
\usepackage{amssymb}
\usepackage{tabularx}
\usepackage{makecell}
\usepackage{footmisc}

\usepackage{placeins}  
 
\usepackage{xcolor}

\newtheorem{theorem}{Theorem}[section]
\newtheorem{proposition}{Proposition}[section]

\newenvironment{proof}{{\noindent\it Proof}\quad}{\hfill $\square$\par}

\begin{document}

\title{Boosting the Generalization Ability for Hyperspectral Image Classification using Spectral-spatial Axial Aggregation Transformer }

\author{Enzhe Zhao,~\IEEEmembership{Student Member,~IEEE,} 
	Zhichang Guo, Shengzhu Shi, Yao Li, Jia Li and Dazhi Zhang
\thanks{Corresponding author: Shengzhu Shi (E-mail: mathssz@hit.edu.cn)}
\thanks{Enzhe Zhao, Zhichang Guo, Shengzhu Shi, Yao Li, Jia Li and Dazhi Zhang are with the Department of Computational Mathematics, School of Mathematics, Harbin Institute of Technology, Harbin 150001, China.}}



\maketitle

\begin{abstract}

In the hyperspectral image classification (HSIC) task, the most commonly used model validation paradigm is partitioning the training-test dataset through pixel-wise random sampling. By training on a small amount of data, the deep learning model can achieve almost perfect accuracy. However, in our experiments, we found that the high accuracy was reached because the training and test datasets share a lot of information. On non-overlapping dataset partitions, well-performing models suffer significant performance degradation. To this end, we propose a spectral-spatial axial aggregation transformer model, namely SaaFormer, that preserves generalization across dataset partitions. SaaFormer applies a multi-level spectral extraction structure to segment the spectrum into multiple spectrum clips, such that the wavelength continuity of the spectrum across the channel are preserved. For each spectrum clip, the axial aggregation attention mechanism, which integrates spatial features along multiple spectral axes is applied to mine the spectral characteristic. The multi-level spectral extraction and the axial aggregation attention emphasize spectral characteristic to improve the model generalization. The experimental results on five publicly available datasets demonstrate that our model exhibits comparable performance on the random partition, while significantly outperforming other methods on non-overlapping partitions. Moreover, SaaFormer shows excellent performance on background classification.

\end{abstract}


\section{Introduction}
\IEEEPARstart{H}{yperspectral} remote sensing images are a type of satellite or airborne images that capture data across a wide range of wavelengths, which is concerned with the extraction of meaningful information based on the radiance acquired by the sensor at short or long distances \cite{refh1}. Unlike multispectral images that capture data in a limited number of wavelength bands, HS images record information across hundreds of narrow bands. In contrast to other remote sensing images, HS image enables a detailed and precise analysis of Earth's surface due to variations in how different materials and surfaces reflect light across the electromagnetic spectrum \cite{refs1,ref25}. Consequently, a profound understanding and thorough modeling are critical for HS image analysis.

In the early stages of HS image analysis, researchers modeled the spectra for individual pixels, traditional classifiers are commonly employed. They assumed that spectra from the same type of area followed the same distribution \cite{refz1,refz2}, were geometrically close to each other \cite{refz3}, or shared the same latent feature space \cite{ref1}. Typically, these techniques involve a two-stage process. Initially, they reduce the dimensionality of the HSI data and extract informative features. Subsequently, spectral classifiers utilize these features for classification purposes\cite{refh2}, such as the k-nearest neighbor\cite{reff1}, the Bayesian estimation method\cite{reff2}, the multinomial logistic regression\cite{ref4} and the support vector machine (SVM)\cite{reff3}. Furthermore, various approaches have been devised to accomplish dimension reduction and extract spectral information. These include techniques such as principal component analysis (PCA)\cite{reff4}, independent component analysis (ICA) \cite{reff5}, and linear discriminant analysis (LDA)\cite{reff6}. However, these methods ignore the spatial correlation among the pixels in spatial dimension and due to factors such as changes in illumination or surface conditions, the spectra can vary, diminishing the generalizability of modeling spectra from a single pixel to HS images. To enhance the extraction of spatial features from images, researchers have introduced several mathematical morphological approaches, including extended MP (EMP) \cite{reff7}, and extended multiattribute profile (EMAP) \cite{reff8}.



The rapid advancement of deep learning technology has hastened the progress of image processing techniques across various domains and has also contributed to the technical innovation of remote sensing image processing\cite{ref7}. For HSI, a multitude of diverse classification methods employing deep models have been put forward. such as 1D convolutional neural networks (CNNs) (1D-CNNs)\cite{ref8}, 2D-CNNs \cite{ref13,ref14,ref15, refh4}, and 3D-CNNs\cite{ref16, ref17} and recurrent neural network (RNN) \cite{ref9}. Hu \textit{et al.} \cite{ref8} introduced a 1D-CNN Classifier to identify and incorporate local dependencies between adjacent spectral bands enhancing the overall comprehension and interpretation of HS images. He \textit{et al.} \cite{ref17} presented the M3D-CNN model for HS image classification, allowing for joint modeling of multiple spectral bands and enabling the integration and fusion of diverse features across different scales. Roy \textit{et al.}\cite{reff9} introduced a hierarchical network structure that combines 3D and 2D CNN characteristics to efficiently extract spatial-spectral features, reducing computational complexity and improving classification accuracy. 

With the prevalence of residual networks in the field of image classification, Zhong \textit{et al.} \cite{refq1} introduced a spatial-spectral residual network for HSI classification, leveraging features from multiple layers to enhance feature utilization. Meanwhile, in \cite{refq2}, a pyramid residual network was devised, enhancing feature map dimensionality and extracting information efficiently through grouped residual blocks. Roy \textit{et al.} \cite{refh8} investigated an improved spectral-spatial residual network for joint feature extraction. Besides, they utilized a lightweight paradigm, employing the squeeze-and-excitation ResNet for spatial and spectral feature extraction, and integrating a bag-of-features learning mechanism to achieve accurate final classification results \cite{reff10}. Meanwhile, in order to fully leverage convolutional neural networks, various complex structures have been introduced to achieve efficient classification, such as Rotation equivariant CNNs \cite{reff12}, gradient centralized convolutions\cite{reff13,reff14}. Although CNNs are effective at capturing spatial contextual information, they struggle to integrate long- and middle-term dependencies sequentially. They may be more suitable for handling local spatial features rather than spectral features, thus failing to fully capture the subtle differences and key features in the spectral information. Consequently, their performance in classifying HSI might suffer, especially when dealing with classes sharing similar spectral signatures, which complicates the extraction of distinguishing spectral attributes.

On the other hand, Recursive neural networks (RNNs) can be employed to model the spectral signatures in HSIs, which can construct a sequence model to effectively simulate the relationship between adjacent spectral bands \cite{ref9}. Yet, for longer hyperspectral sequences, RNNs may encounter the issue of vanishing or exploding gradients, resulting in training difficulties and reduced computational efficiency. Thus, we consider the aforementioned limitations by rethinking HSI classification using Transformers\cite{ref18}, which as leading-edge backbone architectures, transformers utilize self-attention mechanisms to optimize the processing and analysis of sequential data. Its integration seamlessly fuses spatial and spectral information through self-attention, enriching the classification model's understanding of holistic features. Hong \textit{et al.} \cite{ref19} developed a new model, SpectralFormer, which can learn spectral representation information from group-wise neighboring bands and construct a cross-layer transformer encoder module. He \textit{et al.} \cite{ref19_1} proposed a bidirectional encoder representation for a transformer that incorporates flexible and dynamic input regions for pixel-based classification of HSIs.  Roy \textit{et al.} \cite{ref19_3} introduced morphFormer, leveraging learnable spectral and spatial morphological convolution operations combined with attention mechanisms for improved performance. Sun \textit{et al.} \cite{ref19_4} implemented spatial and spectral tokenization in the encoder to extract local spatial information and establish long-range relations between neighboring sequences. While transformer-based models are computationally expensive when dealing with long sequential data due to the self-attention across the entire sequence. This complexity leads to high computational costs and memory consumption. Additionally, the sparsity and redundancy of HSIs  pose a challenge for the self-attention mechanism to capture crucial information. Consequently, these models may struggle to extract features and preserve spectral information.


Furthermore, recent advances in RS technology have increased the availability of multi-sensor data, allowing for multiple representations of the same geographical region. Multimodal data integration has also attracted significant attentions \cite{refh15, refh16, refh18, refh19}. Hong \textit{et al.} \cite{refh17, refh20} proposed a general and unified multimodal deep learning framework focusing on RS image classification. This framework integrates pixel-level labeling guided by an FC design and spatial-spectral joint classification with a CNN framework. Roy \textit{et al.} \cite{refh16} developed a multimodal fusion transformer to extract features from HSIs and fuse them with a CLS token derived from light detection and ranging (LiDAR) data to enhance the joint classification performance. They further \cite{ref20_2, ref20_3} contributed to the benchmark dataset, transfer learning, plug-and-play techniques for the multimodal remote sensing.  Recently, with the significant surge in foundation models based on pre-training techniques \cite{ref20_4}, SpectralGPT \cite{ref20_5} pioneered an universal RS base model utilizing 3D Generative Pre-trained Transformer (GPT).


Meanwhile, with the development of artificial intelligence and the emergence of large-scale remote sensing image datasets, the trustworthy artificial intelligence (AI) in remote sensing(RS) has become a focal point of widespread attention\cite{reff15}. Currently, the development of RS image data presents multimodal features, such as multi-angle, multi-temporal, multi-platform and multi-scale characteristics\cite{reff16, ref6_1}. Multimodal data-driven RS intelligent interpretation algorithms (RSIIA) have been widely used in various fields of RS image processing. At present, data-driven AI has advanced notably in RS, yet it has introduced challenges regarding interpretability and trustworthy. To address this, researchers have focused on investigating trust at two levels—data and model—within the RS field. On one hand, to enhance the trustworthiness of  RS at the data level, key strategies involve improving data quality, integrating multimodal data, updating data, and promoting data sharing. Spatial-temporal-spectral data fusion\cite{reff17} facilitates obtaining RS imagery with high temporal, spatial, and spectral resolutions. The fusion of multisource and multimodal data\cite{reff18} provides additional information gain for RS intelligent interpretation. In terms of the AI models, given their black-box characteristics, research on model interpretability is crucial in assessing the credibility of the models. Kakogeorgiou \textit{et al.}\cite{reff19} explored interpretable AI methods applied to the task of multi-label classification of RS. Moreover, assessing model uncertainty is a crucial approach to improve the reliability of AI. He \textit{et al.}\cite{reff20} addressed data and model uncertainty in HSI classification, leading to the development of a reliable HSI classifier through the evaluation and mitigation of these uncertainties. Wang \textit{et al.} \cite{reff21} introduced a technique for quantifying the uncertainty of multimodal data in order to enhance the precision and trustworthiness of data fusion. In summary, trustworthy research in the data-driven RS is extensive, emerging a series of AI methods for measuring, evaluating, and enhancing the trustworthiness of RS interpretation.

Based on the above analysis and discussion, a new spectral-spatial fusion transformer model is introduced in this work to take advantage of the transformer’s ability to obtain local spatial semantic information and model the relationship between adjacent sequences. The objective of our model, SaaFormer is to extract and fuse spectral-spatial features across multiple scales along the spectral dimension from the patch embeddings of HSI, as well as to ensure efficient computation and grasp intricate spectral nuances, generating comprehensive feature representations, bolstering the generalization and credibility of the model.


\begin{figure*}[ht!]
	\centering
	\includegraphics[width=0.8\textwidth]{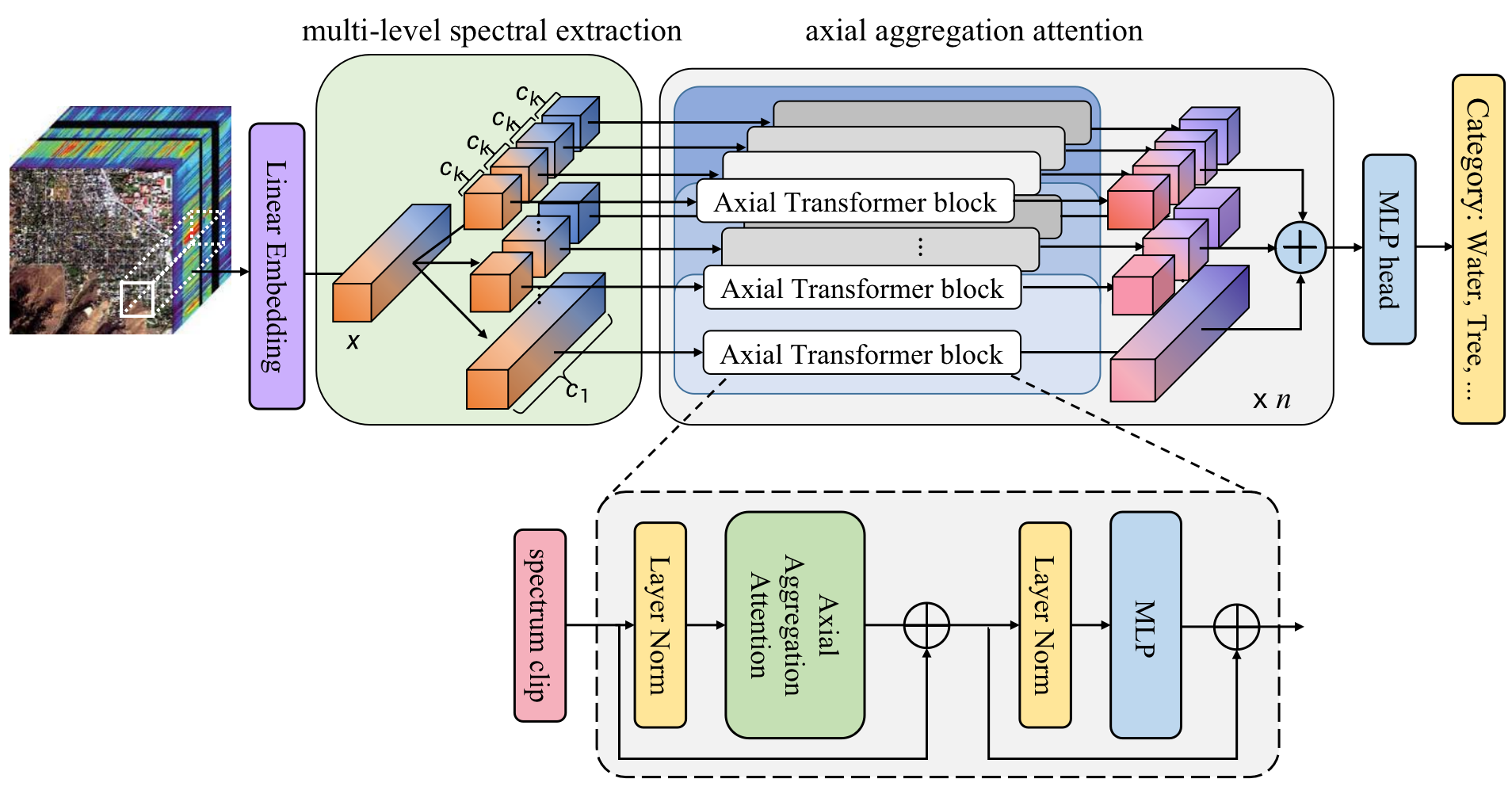}
	\caption{Overview Architecture of SaaFormer for HS image classification task. The model architecture consists of two primary components: the multi-level spectral extraction structure and the axial aggregation attention.} 
	\label{Fig:saanet} 
\end{figure*}

The primary contributions of this paper can be outlined as follows:
%
%
%

	\begin{itemize}
		\item We provide a spectral-spatial axial aggregation transformer framework, namely SaaFormer, which emphasizes fine-grained feature extraction through multi-spectral extraction modules and efficiently aggregates global spatial and spectral feature information with axial transformer modules thereby reducing computational complexity while significantly improving the handling of redundancy in HSIs and enhancing the capability to extract and maintain spectral features. Consequently, it has achieved state-of-the-art performance across five public datasets.
		\item We have raised significant concerns regarding the generalization of HSI classification. Through comprehensive theoretical and experimental analyses of information sharing, we discovered that excessive overlap between the training set and test set tends to yield overly optimistic model performance. And our model emphasizes central pixels with axial transformer modules, thereby mitigating interference from related pixels more effectively and improving generalization capabilities, as validated by various partitioning methods. 
		\item The proposed SaaFormer achieves exceptional performance in the background classification task, which further supports its superior generalization capabilities and perform well across diverse datasets and varying conditions. 
	\end{itemize}

The remaining structure of the paper is outlined as follows: Section II presents the model structure we have proposed along with its key details. In Section III, we discuss several dataset partitioning methods aimed at mitigating data share information issues. In Section IV, we conduct extensive experiments, offering specific experimental data and visual results. Lastly, Section V offers comprehensive conclusions based on a thorough analysis.

\section{Methodology}

\begin{figure*}[htbp]
	\centering
	{\includegraphics[width=0.6\textwidth]{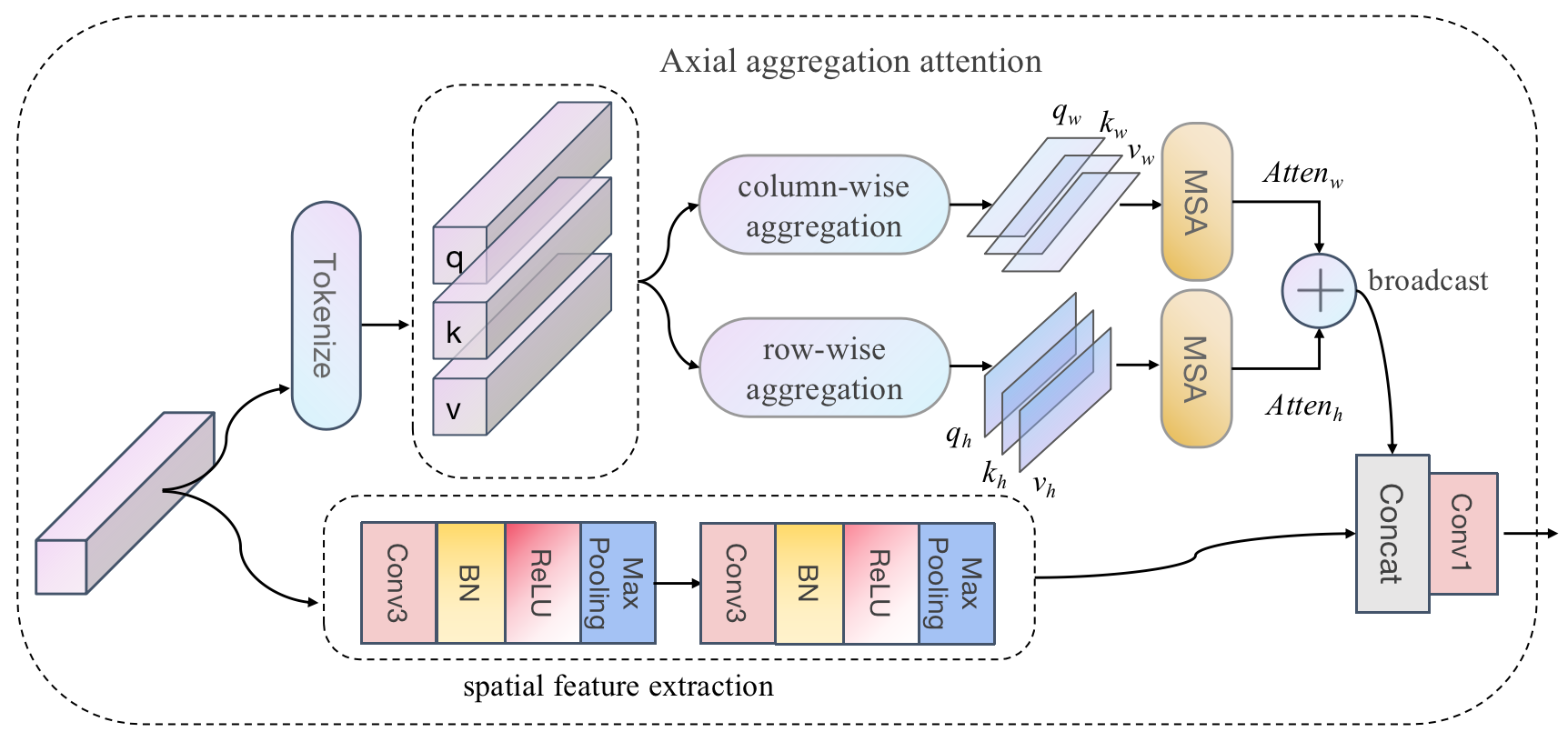}}
	\caption{Diagram of the Axial Transformer Block in the proposed SaaFormer. The attention process compresses the feature map either vertically or horizontally and calculates self-attention separately for each axis. }
	\label{Fig:attenb}
\end{figure*}

\subsection{Architecture of SaaFormer}

The architecture of the proposed SaaFormer is shown in Fig. \ref{Fig:saanet}. SaaFormer consists of two primary components: the multi-level spectral extraction structure and the axial aggregation attention mechanism. The multi-level spectral extraction structure conducts $K$ different-levels segmentation plans to the linear embedding $\mathbf{x} \in \mathbb{R}^{h\times w\times c}$ of the sample $\mathbf{x}_0 \in \mathbb{R}^{h\times w\times c}$ such that $\mathbf{x}$ is segmented into clips with length $c_1,c_2,...,c_K$. Segmenting the spectra into clips instead of weighting each spectra channel can preserve the physical continuity of the spectra wavelength across channels. Extracting features in the unit of spectra clips regroups the information in each single spectra channel and improves model generalization ability. The multi-level segmentation plans ensure that the multi-level features can be observed from clips in different granular. 

For each spectrum clip, the axial aggregation transformer block is applied to mine the spectral characteristic. The axial transformer block integrates row and column-wise spatial features along multiple spectral axes to emphasize spectral characteristics. Then, the features mined from each spectrum clip are integrated as the nonlinear embedding of the sample. The axial attention mechanism applies the spatial information as the auxiliary of the spectra information analysis task, such that the model can have a deep understanding of the spectra instead of simply memorizing spatial features, to improve the model generalization. The implementation details of the multi-level spectral extraction structure and the axial aggregation attention mechanism are described in the following sections.

\subsection{Multi-level Spectral Extraction}

From the view of the absorption or reflection characteristics of different materials for different wavelengths of light, the absorption peaks and the reflection peaks are usually interspersed occasionally. Thus, we propose to analysis the spectra of HS images with multi-level spectra clips. In the multi-level spectral-spatial extraction structure, the linear embeddings of the samples are segmented into non-overlapping continuous spectral segments of varying lengths, which are considered as spectral clips. As illustrated in Fig. \ref{Fig:saanet}, the feature map $x$ is divided into spectral clips of different lengths (i.e. $c_1,c_2,...,c_K$) in spectral dimensions. The multi-level segmentation plans ensure that the multi-level features are observed from clips at different levels of granularity. The spectral characteristic of each spectrum clip is mined using the axial aggregation transformer block.

\subsection{Axial Transformer Block}

The axial transformer block is applied to each spectra clip for deep understanding of the spectra segmentation. It utilizes spatial information as an auxiliary by project the spatial information onto spectral axes. To reduce the structure redundancy, we project the row-wise spatial information and the column-wise spatial information separately to the spectral axes. Then, the spatial-enhanced spectral features are integrated through the multi-head self-attention mechanism (MSA), which provides adaptive attention to different spectral channels. The axial self-attention mechanism considers the interrelationships among spectral bands at each pixel location and aggregating features in the spatial domain to improve the model generalization capability. The axial attention mechanism has also been used in various tasks \cite{ref20, ref21, ref21a, ref21b} as lightweight modifications to improve the self-attention mechanism \cite{refk1, refk2, refk3}. Hence, in addressing the challenge of HSIC, we improve \cite{ref21b}.  Due to its imperfect alignment with the specific characteristics of high-dimensional spectral data, we have undertaken further improvements by designing the spatial-spectral axial aggregation attention mechanism.

The structure details are shown in Fig. \ref{Fig:attenb}, firstly, the query, key, and value of the spectral clip $\mathbf{x} \in \mathbb{R}^{ h\times w \times c}$ is calculated through linear projections, i.e., $\mathbf{q}=\mathbf{x}\mathbf{W}_q$, $\mathbf{k}=\mathbf{x}\mathbf{W}_k$ and $\mathbf{v}=\mathbf{x}\mathbf{W}_v$ where $\mathbf{W}_q, \mathbf{W}_k, \mathbf{W}_v\in  \mathbb R^{c\times d}$ are all learnable matrices, and $d$ is the given hyperparameter. Then, we achieve horizontal and vertical spatial information projection by taking the maximum value of the query feature map in the row and column directions, $\bold q_{h}={\rm max}_w(\bold q), \bold q_{w}={\rm max}_h(\bold q)$. The aggregation operation consolidates global information by compressing it into a single axis, and the  operation on $\mathbf{q}$ also repeats on $\mathbf{k}$ and $\mathbf{v}$ to obtain $\mathbf q_{h}, \mathbf k_{h}, \mathbf v_{h}\in \mathbb{R}^{h\times d}$, $\mathbf q_{w}, \mathbf k_{w}, \mathbf v_{w}\in \mathbb{R}^{w\times d}$. The last addition in the Eq. (\ref{key3}) is realized by the broadcast operation, i.e. $\mathbf{z}\in \mathbb{R}^{ h\times w \times d}$

\begin{align}\label{key3}
		\operatorname{Atten}_h&=\operatorname{SoftMax}\left(\mathbf{q}_{h} \mathbf{k}_{h}^{T}/\sqrt{d}\right) \mathbf{v}_{h},\notag\\
		\operatorname{Atten}_w&=\operatorname{SoftMax}\left(\mathbf{q}_{w} \mathbf{k}_{w}^{T}/\sqrt{d}\right) \mathbf{v}_{w},\notag\\
		\mathbf{z} =& \operatorname{Broadcast}(\operatorname{Atten}_h+\operatorname{Atten}_w)
\end{align}

To introduce the position information \cite{ref21b, ref21c, refk1, ref21}, or equivalently the wavelength information, to the features, we include a learnable relative position encode to each head which are gained from learnable parameters $ Pe\in \mathbb{R}^{N\times d}$, $N$ changes with the input parameters. And apply the learnable relative position encode as $\bold p_{h}^q, \bold p_{h}^k,\bold p_{h}^v\in \mathbb R^{h\times d}$ directly adding to the $\bold q_{h}, \bold k_{h}, \bold v_{h}$, and $\bold p_{w}^q, \bold p_{w}^k, \bold p_{w}^v \in \mathbb R^{w\times d}$ are also applied to  $\bold q_{w}, \bold k_{w}, \bold v_{w} \in \mathbb R^{w\times d}$ in the same way. Thus, the positional-aware aggregation attention can be expressed as 

\begin{align}\label{key4}
	\operatorname{Atten}_h =&\operatorname{SoftMax}\left(\frac{(\mathbf{q}_{h}+\mathbf{p}_{h}^q) (\mathbf{k}_{h}+\mathbf{p}_{h}^k)^{T} }{\sqrt{d}}\right)(\mathbf{v}_{h}+\bold p_{h}^v),\notag\\
	\operatorname{Atten}_w=&\operatorname{SoftMax}\left(\frac{(\mathbf{q}_{w}+\mathbf{p}_{w}^q) (\mathbf{k}_{w}+\mathbf{p}_{w}^k)^{T}}{\sqrt{d}}\right)
	(\mathbf{v}_{w}+\bold p_{w}^v),\notag\\
	\mathbf{z} &= \operatorname{Broadcast}(\operatorname{Atten}_h+\operatorname{Atten}_w)
\end{align}

As is shown in the upper path of Fig. \ref{Fig:attenb}, the attention process compresses the feature map either vertically or horizontally and calculates self-attention separately for each axis. By incorporating horizontal and vertical axial attention modules, a global receptive field is achieved. However, this axial attention mechanism prioritizes essential spectral features but may overlook spatial local details. To address this, we incorporate an additional spatial feature extraction module comprising $3\times 3$ convolution and batch normalization. This module supplements spatial details and preserves the sample's spatial structure.


\section{Generalization Validation}

\begin{figure}[htbp]
	\centering
	\includegraphics[width=0.4\textwidth]{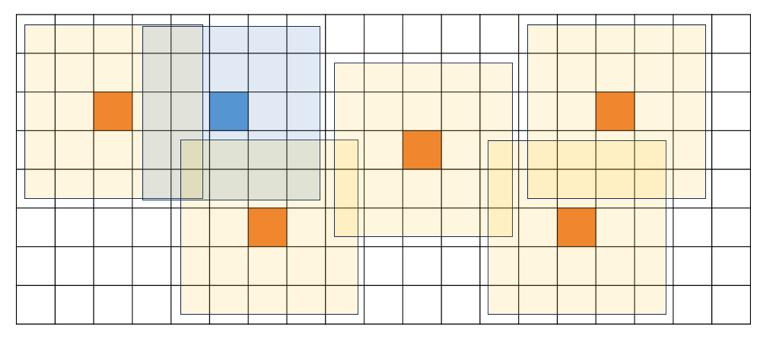}
	\caption{The pixel-wise random sampling dataset partition. Five percents of the pixels are selected as the training dataset and marked in orange. The rest pixels (blue and white) are samples in the test dataset.} 
	\label{Fig:biaoge} 
\end{figure}

In the deep learning model validation paradigm, the HS image is partitioned as training dataset and test dataset. The training dataset usually contains a small portion of pixels. For each pixel, its neighbor pixels (usually in a 7$\times$7 patch) are also fed into the model to provide spatial and contextual information. The pixels in the training dataset are randomly selected from the entire HS image. It has been demonstrated that with only 5\% pixels, deep learning models \cite{ref13,ref14,ref15,ref16,ref17,ref18,ref19} can reach almost 100\% classification accuracy on the test dataset which constructed by the rest 95\% of the HS image pixels. However, the patches in the test dataset actually share many pixels with some patches in the training dataset. As shown in Fig. \ref{Fig:biaoge}, the pixels and its neighbor pixels in the training dataset cover a large portion of the HS image although the training dataset contains only 5\% of the HS image pixels. For an arbitrary sample in the test dataset, for example the pixel in blue in Fig. \ref{Fig:biaoge}, there is a large chance that the sample shares pixels with a sample in the training dataset. If the high classification accuracy is reached because of the shared pixels, or rather shared information, instead of the model generalization, the samples which share more pixels with the training dataset will have higher probability to be classified correctly comparing to those sample share less pixels with the training dataset. This phenomenon is observed in our following experiment. 

\begin{figure}[htbp]
	\centering
	\includegraphics[width=0.48\textwidth]{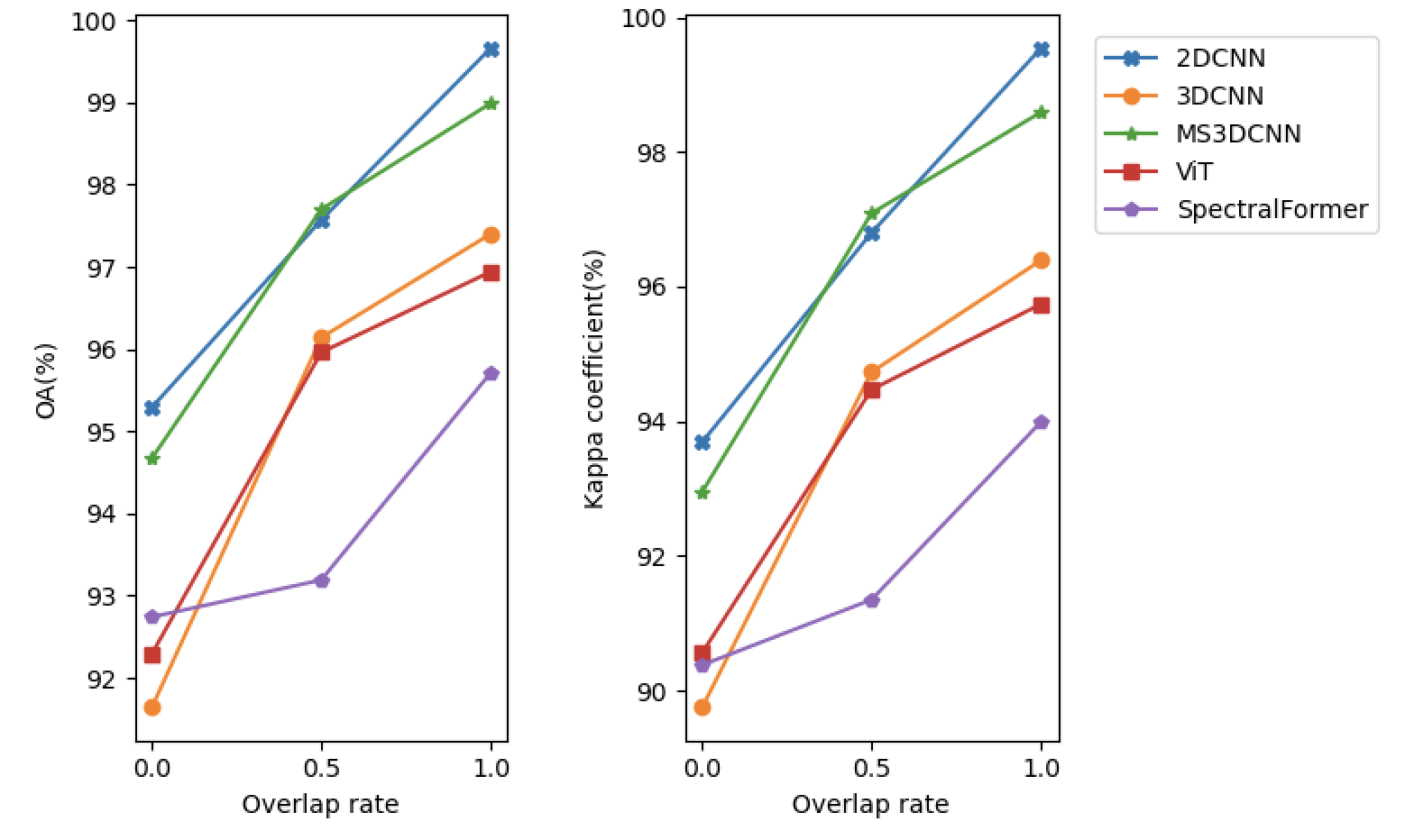}
	\caption{Comparison of the accuracy of the overlap rate of training samples and test samples for different models.} 
	\label{Fig:duibi} 
\end{figure}

We divide all the samples in the test dataset into three subsets according to their maximum overlap rates with samples in the training dataset. The three subsets are 1) sample without overlap, 2) samples with overlap rate between 0\%-50\%, and 3) samples with overlap rate is higher than 50\%. The classification accuracies are calculated for each subset on various popular models. As can be seen in Fig. \ref{Fig:duibi}, the test samples with higher overlap rate exhibit significant higher accuracy than samples with lower overlap rate. Therefore, the high classification accuracy is reached because of the shared pixels instead of the model generalization.

As we discussed before, the samples in the training dataset and test dataset share information between each other if random sampling is applied. The shared information can lead to overly optimistic model performance. In this section, we introduce several dataset partition methods to provide a more comprehensive model generalization evaluation. If the model has good performance across different sampling methods, instead of merely the random sampling, then we conclude that the model has good generalization capability. We further give mathematical analysis of the information share for each dataset partition method.

\begin{figure}[htbp]
	\centering
	\includegraphics[width=0.48\textwidth]{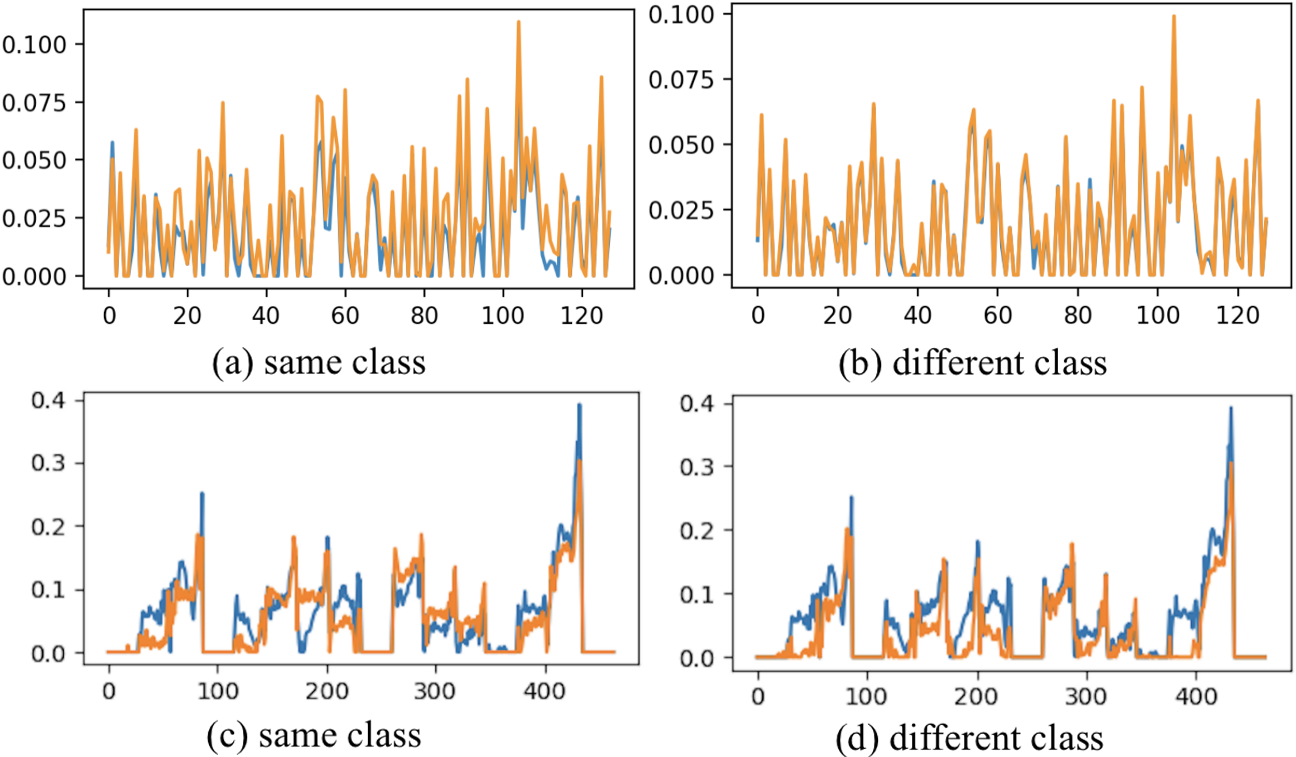}
	\caption{Spectral features visualizations: (a) and (b) are spectral features extracted by the CNN model \cite{ref15}, while (c) and (d)  are spectral features extracted by  the 3DCNN model \cite{ref17}. The blue and oranges lines are spectral features of two samples from the same class, for instance (a) and (c), or different classes, such as (b) and (d). } 
	\label{Fig:cnntongji} 
\end{figure}

Furthermore, we visualize the spectral features extracted by 2DCNN \cite{ref15} and 3DCNN \cite{ref17} models for further analysis. The 2DCNN model achieved 99.35\% accuracy and the 3DCNN model achieved 97.36\% accuracy on the PaviaU dataset. As shown in  Fig.  \ref{Fig:cnntongji} (a) and (b), for the 2DCNN model, the spectral patterns are very similar for the samples from the same class as while as the samples from the different classes. For the 3DCNN model (Fig. 3(c) and (d)), the spectral patterns are not very similar for the samples from both the same class and the different classes. However, for hyperspectral image analysis, the spectral information is the most important and should be distinguishable between classes.

Therefore, based on the above experiment and analysis, models trained under random sampling paradigm can achieve good performance on the test dataset. However, the performance dependents on the overlapping rate between the training samples and the test samples. Moreover, the spectral patterns extracted by the models can't effectively differentiate between samples from different classes. 

\subsection{Dataset Partitions}
\subsubsection{Random Sampling}
The most commonly used model validation paradigm for the HS image classification is the pixel-wise random sampling. It random selects a small portion of pixels (5\%) as the training dataset and leave the rest of the pixels as the validation dataset. Each sample consists of a central pixel and its surrounding region ($5\times 5$ pixels patch) to provide spatial information. However, as we discussed above, random sampling can result in a high overlap between samples in the training dataset and the validation dataset, leading to a significant similarity between the training and the validation dataset.

\subsubsection{Checkerboard Sampling}
To reduce the overlap between the samples in the training dataset and the samples in the validation dataset, we propose the checkerboard sampling which partitions the training-test datasets through non-adjacent checkerboards. Comparing to the random sampling, samples fall in the inner area of each checkerboard of the validation dataset will not overlap with the samples of the training dataset.

\subsubsection{Block-wise Sampling}
To further explore the generalization ability of the model, we propose the block-wise sampling which divides the training and validation datasets into non-overlapping horizontal or vertical blocks.

\subsubsection{K-means sampling}
In addition, we employ k-means algorithm to cluster each class as $k$ disjoint datasets and random select $k/2$  datasets as the training dataset while leaving the rest as the validation dataset.

\begin{figure}[htbp]
	\centering
	\subcaptionbox{trainset \\random \label{train_r}}
	{\includegraphics[width=0.11\textwidth]{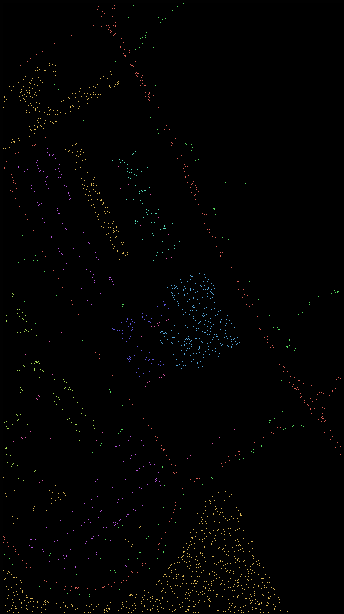}}
	\hskip 1.5mm 
	\subcaptionbox{trainset  \\checkerboard \label{train_squ}}
	{\includegraphics[width=0.11\textwidth]{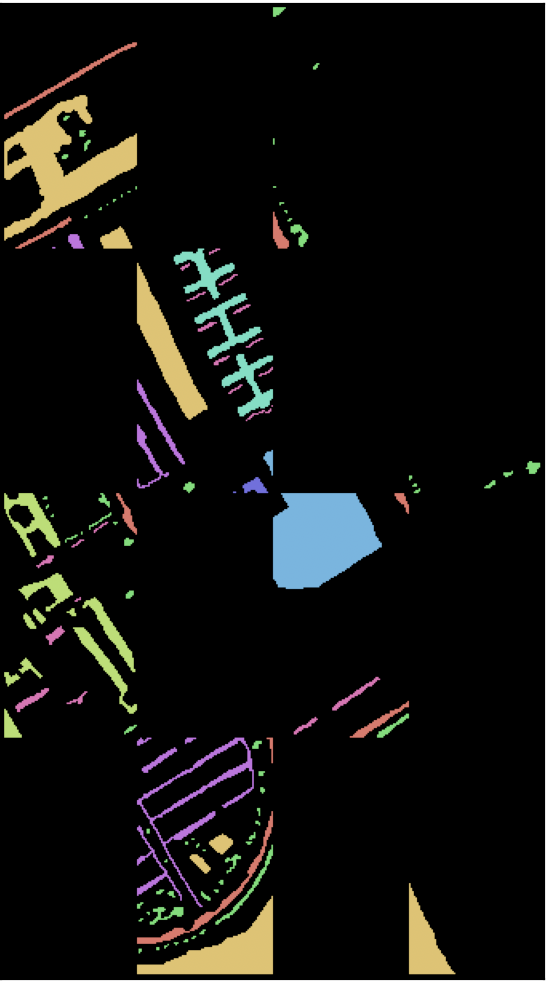}}
	\hskip 1.5mm 
	\subcaptionbox{trainset \\block-wise\label{train_8}}
	{\includegraphics[width=0.11\textwidth]{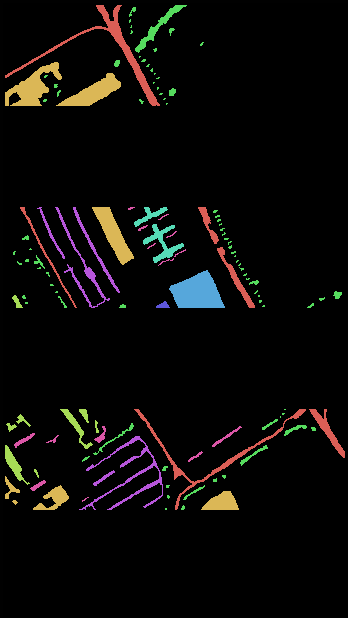}}
	\hskip 1.5mm 
	\subcaptionbox{trainset \\k-mean\label{train_k}}
	{\includegraphics[width=0.11\textwidth]{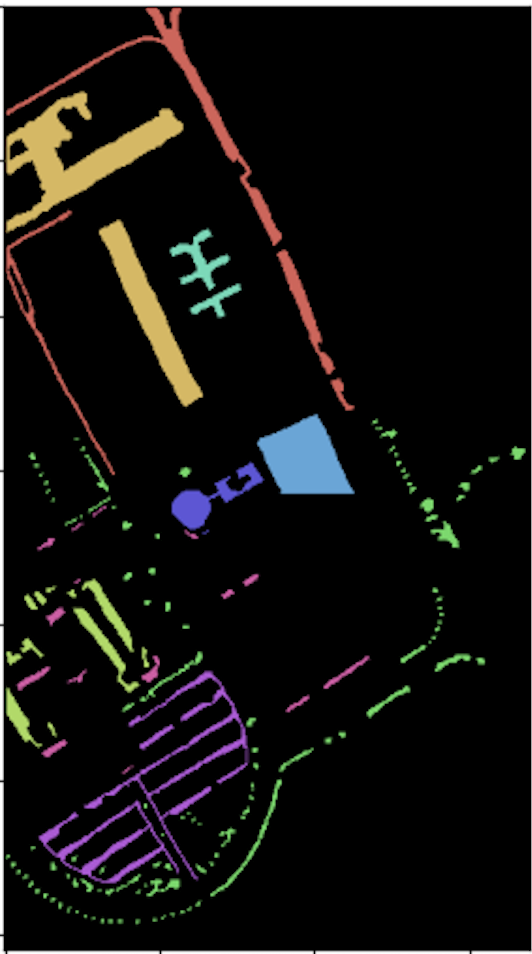}}
	\\ \vspace{1mm} 
	\subcaptionbox{testset  \\random\label{test_r}}
	{\includegraphics[width=0.11\textwidth]{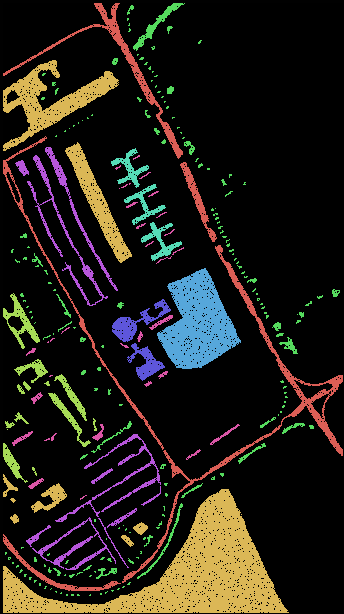}}
	\hskip 1.5mm 
	\subcaptionbox{testset  \\checkerboard\label{test_squ}}
	{\includegraphics[width=0.11\textwidth]{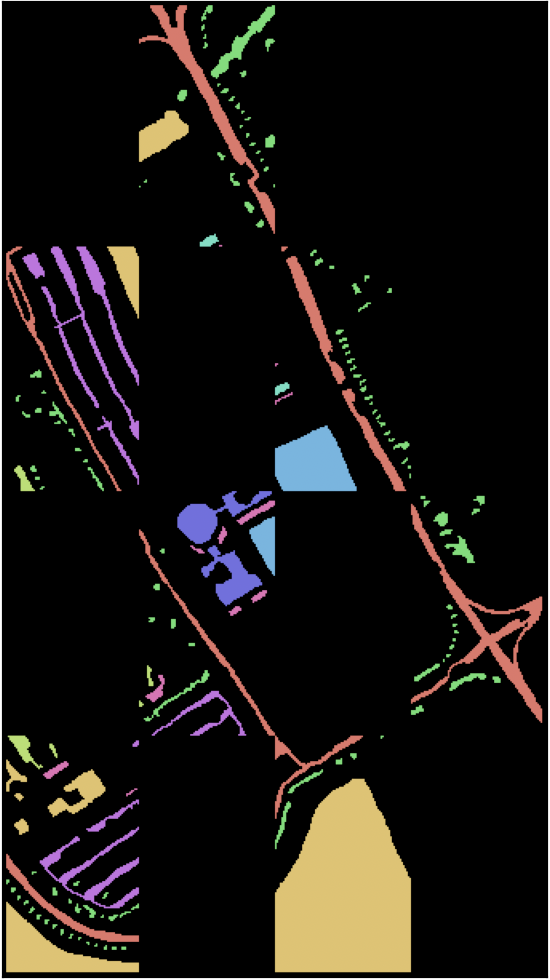}}
	\hskip 1.5mm
	\subcaptionbox{testset  \\block-wise\label{test_8}}
	{\includegraphics[width=0.11\textwidth]{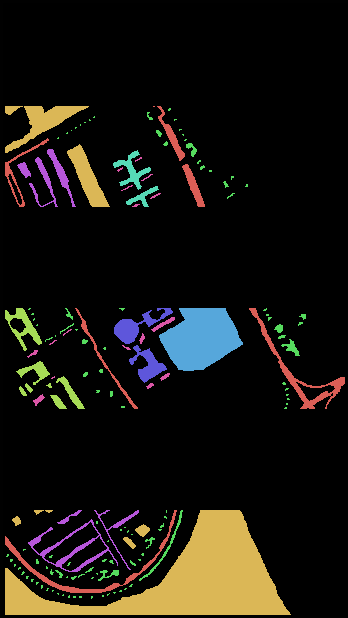}}
	\hskip 1.5mm 
	\subcaptionbox{testset \\k-mean\label{test_k}}
	{\includegraphics[width=0.11\textwidth]{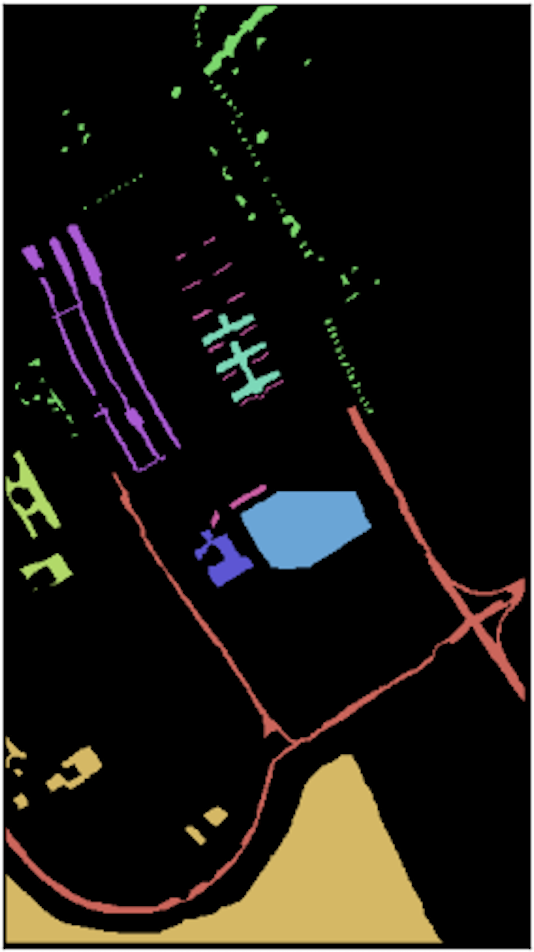}}
	\caption{Comparison diagram of PaviaU dataset by four different methods to sample}
	\label{Fig:division}
\end{figure}

The checkerboard, block-wise and k-means samplings ensures that almost all the samples in the training dataset has no overlap with the samples in the validation dataset.  While samples in the same class may gather in a small area in the entire HS image. The block-wise sampling can fail to separate all the classes equally. In this case, the k-means sampling can provide better sample partition. For the model generalization validation, if the model has good performance across different sampling methods, instead of merely the random sampling, then we conclude that the model has proper generalization capability. The diagram Fig. \ref{Fig:division}. illustrates the aforementioned partitions.

\subsection{Information Share of Dataset Partitions}
Firstly, we interpret the classification using a set of randomly selected training samples as a random classification problem \cite{ref24}. Specifically, we assume $ \Omega$ denote the entire sample space, i.e. the hyperspectral image and $N$ is the total number of data samples. The sample space $\Omega$ is divided into mutually exclusive training dataset $\Omega_T$ and test dataset $\Omega_t$ through random permutation. Let $\alpha$ indicates the sampling ratio(SR) of the training dataset. Then, the size of $\Omega_T$  and $\Omega_t$  are $|\Omega_T|=\alpha N$ and $|\Omega_t|=(1-\alpha)N$, respectively, where $|\Omega|=N$. By the distribution of the random permutation, the probability that a sample $S(x,y)$ belongs to the training dataset or the test dataset are
\begin{equation}
	P\{S\in \Omega_T\}=\alpha,P\{S\in \Omega_t\}=1-\alpha
\end{equation}

The following result shows that the random sampling method causes a high probability of information shared between the training and test dataset, which could lead to an over-estimated performance for the classifier. 

\begin{theorem}\label{th1}
	The dataset is divided into mutually exclusive training and testing sets through random permutation, where the training set accounts for $\alpha$, and the testing set accounts for $1- \alpha$. If the sample size is $n\times n$, for an arbitrary test sample $S$, the probability it shares information with the training dataset is
	\begin{equation}
		p_I(S) = 1-(1-\alpha)^{4n^2-1}
	\end{equation}
	The expected information benefit from the training dataset is
	\begin{equation}
		E[I(S)] = (n^2-1)\alpha H(S)
	\end{equation}
	where $H(S)$ is the entropy of $S$.
\end{theorem}

\begin{proof}
	If any sample within the square centered at $S(x,y)$ of size $2n$ belongs to the training dataset, the two samples will have pixel overlap between each other, so as their information. Only if all the samples in the square belongs to the test dataset, $S$ doesn’t share any information with the training dataset. Thus, by the independency of the samples, the probability that $S$ shares information with the training dataset is
	$$p_I(S) = 1 - P\{S \in \Omega_t\}^{(4n^2-1)}$$
	where \(\Omega_t\) is the test set. By the distribution of the random permutation, \(P\{S \in \Omega_t\} = 1 - \alpha\). Thus,
	\[p_I(S) = 1 -(1 -\alpha)^{(4n^2-1)} \]
	As for the expected information benefit, we first define the information share ratio (ISR) between two samples $S(x, y)$ and $S^\prime(x^\prime, y^\prime)$ as the ratio of the information they share to the information of $S$ as
	\begin{equation}
		\gamma(S,S^\prime):=\dfrac{I(S,S^\prime)}{H(S)}
	\end{equation}
    where $I(S, S^\prime)$ is the mutual information. Assuming each pixel contains the same amount of information, the ISR can be represented by the distance of the two samples as
    \begin{equation}
    	\gamma(S,S^\prime):=\dfrac{I(S,S^\prime)}{H(S)}=\frac{i_xi_y}{n^2}
    \end{equation}
    where \(i_x = (n - |x-x^\prime|)_+, i_y = (n - |y - y^\prime|)_+\), and \((x)_+ = x\mathbf{1}_{x>0}.\)
	The expected information benefit of $S$ is
	\begin{equation}
		\begin{aligned}
			E[I(S)] &= \sum_{S^\prime}E_{S'}[\gamma(S,S^\prime)H(S)]=\sum_{S^\prime\in \Omega_T}\gamma(S,S^\prime)\alpha H(S) \\
			&= ((\sum_{x^\prime=x-n}^{x+n}\sum_{y^\prime=y-n}^{y+n}\frac{i_xi_y}{n^2})-1)\alpha H(S) \\
			&= (n^2-1)\alpha H(S) \\
		\end{aligned}
	\end{equation}
    where $\Omega_T$ is the training set.
\end{proof}

By Theorem \ref{th1}, if we take a commonly used setting, i.e., dividing $\alpha$ = 5\% data as the training dataset and applying $n = 5$ pixels patch for each sample, the probability that an arbitrary sample in the test dataset shares information with the training dataset is $p_I(S) > 0.993$
and the expected information benefit is $1.24H(S)$. Thus, the samples in the test dataset can take a lot of information benefit through the sample overlapping, which contains even more information than the sample itself. Due to the information benefit, the information in the test dataset is already included in the training dataset. Hence, it’s not surprising to see good performance on the test dataset. However, the classifier can fail to generalize on samples it doesn’t share or share less information with the training dataset.

\begin{theorem}\label{th2}
	The dataset is divided through block-wise sampling, where the height of each block is $h$ for row-wise (or the width of each block is $w$ for column-wise). If the sample size is $n \times n,$ for an arbitrary test sample $S$, the expected information benefit the training dataset is
	\begin{equation}
		E(I(S))=\frac{(n^2-n)}{h}H(S)
	\end{equation}
\end{theorem}

\begin{proof}
	\begin{equation}
		\begin{aligned}
			E[I(S)] &= \sum_{S^\prime}E_{S^\prime}[\gamma(S,S^\prime)H(S)] \\
			&= ((\sum_{y^\prime=y-n}^{y-1}\frac{i_yn}{n^2} + \sum_{y^\prime=y+1}^{y+n}\frac{i_yn}{n^2}))\frac{n}{h} H(S) \\
			&= \frac{n^2-n}{h} H(S) \\
		\end{aligned}
	\end{equation}
\end{proof}

\begin{proposition}\label{th3}
	Assume that the sample space $\Omega$  divided training dataset $\Omega_T$ and test dataset $\Omega_t$ through checkerboard sampling, i.e. non-adjacent checkerboards. If the sample size is $n\times n$ for an arbitrary test sample $S$, and the height of each checkerboards is $h$ and the width $w$. The expected information benefit for the checkerboard sampling is
	\begin{equation}
		E[I(S)]\leq (n^2-n)(\frac{1}{h}+\frac{1}{w})H(S)
	\end{equation}
\end{proposition}

\begin{proposition}\label{th4}
	The expected information benefit for the k-means sampling is bounded by
	\begin{align}
		&E[I(S)] \geq (n^2-n)(\min(\frac{1}{h},\frac{1}{w}))H(S) \notag\\ 
		&E[I(S)]\leq (n^2-n)(\frac{1}{h}+\frac{1}{w})H(S)
	\end{align}
	where $h$ and $w$ are the total height and width of the image divided by $k$.
\end{proposition}

The Proposition \ref{th3} and \ref{th4} are intuitive to proof. Taking the University of Pavia dataset ($610 \times 340$ pixels) as example, if the dataset is divided into 8 rows, the expected information benefit is $0.258H(S)$, which is about one-fifth of the information benefit of the 5\% random sampling. If the 8-by-8 checkerboard sampling is taken, the expected information benefit is $0.519H(S)$. The expected information benefit of the k-means sampling can be upper bounded by checkerboard sampling and lower bounded by block-wise sampling. If $k=4$, the expected information benefit is between $0.131H(S)\leq E[I(S)]\leq 0.365H(S)$.

From the specific values of the expected information benefit, it can be seen that the three proposed partition methods can make the test set less polluted by the training set than random sampling, so as to provide better measure of the generalization ability and robustness of the classification model.

\section{Experiments}
In this section, we first introduce several commonly used HS image datasets and the details for datasets partion methods. Then, we compare our proposed SaaFormer with other state-of-the-art methods under the various dataset partition methods to validate the generalization ablility of the models. Extensive experiments are finally conducted, including ablation analysis, to validate the efficacy of each component in the proposed model.

\subsection{Data Description}
Experiments are conducted on six hyperspectral datasets\footnote{All these datasets can be obtained at https://www.ehu.eus/ccwintco/index.php/\par Hyperspectral\_Remote\_Sensing\_Scenes}. The detailed descriptions of these datasets are listed as follows.

\begin{itemize}
	\item{The Indian Pines dataset consists of 145 × 145 pixels and 220 spectral bands. The dataset has 20 m per pixel spatial resolutions and 10 nm spectral resolutions covering a spectrum range of 200–2400 nm. In the experiment, after removing 20 noisy and water absorption bands, 200 spectral bands are retained. The ground truth is composed of 16 vegetation classes.}

	\item{The Pavia Centre dataset consists of 102 spectral bands, while the Pavia University dataset has 103 spectral bands. The Pavia Centre image has a size of 1096 × 1096 pixels, and Pavia University is 610 × 610 pixels. Some samples within both images do not contain any information and need to be discarded for analysis purposes.  Both datasets have ground truths that classify into 9 different classes each.}

	\item{The Salinas scene dataset was obtained using the 224-band AVIRIS sensor over Salinas Valley, California. It covers an area of 512 lines by 217 samples. The dataset also removed 20 bands affected by water absorption. And the image was provided solely as at-sensor radiance data.}

    \item{The Kennedy Space Center (KSC) dataset was collected from an altitude of around 20 km, providing a spatial resolution of 18 m. After eliminating bands affected by water absorption and low signal-to-noise ratio, the analysis was conducted using 176 bands. The dataset defined 13 classes that represent different land cover types}
    
    \item{The Botswana dataset was obtained at a pixel resolution of 30 m, covering a strip of 7.7 km in 242 bands. To enhance data quality, uncalibrated and noisy bands that corresponded to water absorption features were removed. The remaining 145 bands, comprising bands were included as candidate features for analysis. The dataset consists of observations from 14 identified classes representing land cover types found in seasonal swamps, occasional swamps, and drier woodlands located in the distal portion of the Delta.}
    
    \item {
    	The Houston2013 dataset was collected using the ITRES CASI-1500 sensor, covering the University of Houston campus and its surrounding rural areas in Texas, USA which is widely recognized as a benchmark for evaluating land cover classification performance \cite{refh17}. The hyperspectral (HS) cube consists of 349×1905 pixels with 144 wavelength bands ranging from 364 nm to 1046 nm at 10 nm intervals. 
    }
	
\end{itemize}


\begin{table}[htbp]
	\centering
	\caption{The sample numbers for the training and test sets for the three data partition methods.}
	\label{tab:sample_num}
	\resizebox{.45\textwidth}{!}{
	\begin{tabular}{cc|ccccc}
		\hline \hline
		\multicolumn{2}{c|}{}                          & Botswana & KSC  & PaviaU & PaviaC & Salinas \\ \hline
		\multirow{2}{*}{Checkerboard} & train & 1403     & 2185 & 17153  & 57070  & 24993   \\
		& test     & 1845     & 3008 & 24395  & 89963  & 28353   \\
		\multirow{2}{*}{Block-wise}   & train & 1585     & 2043 & 14264  & 65076  & 25964   \\
		& test     & 1663     & 3168 & 28512  & 83076  & 28165   \\
		\multirow{2}{*}{K-means}      &train & 1170     & 1914 & 19982  & 67948  & 25517   \\
		& test     & 2078     & 3279 & 21566  & 79085  & 27829   \\ \hline \hline
	\end{tabular}}
\end{table}

%

\begin{table*}[htbp]
	\centering
	\caption{Comparison of classification accuracy of different sampling methods on Botswana dataset}
	\label{tab:botswana}
	\resizebox{.75\textwidth}{!}{
		\begin{tabular}{cc|cccccccccc}
			\hline \hline
			&       & 1DCNN  & 2DCNN  & 3DCNN  & MS3DCNN & RNN    & ViT    & SpectralFormer & MorphFormer & SSFTTnet & SaaFormer       \\ \hline
			\multirow{3}{*}{Random}     & OA    & 0.8869 & 0.9818 & 0.9465 & 0.9345  & 0.8700 & 0.9102 & 0.9782         & 0.9559      & 0.9780   & $\bold{0.9925}$ \\
			& AA    & 0.8977 & 0.9757 & 0.9475 & 0.9306  & 0.8779 & 0.9136 & 0.9808         & 0.9440      & 0.9755   & $\bold{0.9933}$ \\
			& Kappa & 0.8775 & 0.9803 & 0.9420 & 0.9290  & 0.8592 & 0.9028 & 0.9764         & 0.9522      & 0.9761   & $\bold{0.9919}$ \\ \hline
			\multirow{3}{*}{Block}      & OA    & 0.8151 & 0.5968 & 0.7810 & 0.6529  & 0.7406 & 0.8296 & 0.6700         & 0.7811      & 0.8196   & $\bold{0.9117}$ \\
			& AA    & 0.8611 & 0.6169 & 0.7585 & 0.6786  & 0.8094 & 0.7870 & 0.6770         & 0.8266      & 0.8633   & $\bold{0.9202}$ \\
			& Kappa & 0.7973 & 0.5645 & 0.7609 & 0.6233  & 0.7193 & 0.8141 & 0.6431         & 0.7619      & 0.8022   & $\bold{0.9027}$ \\ \hline
			\multirow{3}{*}{Checkerboard} & OA    & 0.8738 & 0.9505 & 0.8788 & 0.8392  & 0.7631 & 0.9181 & 0.9043         & 0.9766      & 0.8268   & $\bold{0.9786}$ \\
			& AA    & 0.9114 & 0.9569 & 0.9182 & 0.8879  & 0.8366 & 0.9301 & 0.9208         & 0.9713      & 0.8997   & $\bold{0.9793}$ \\
			& Kappa & 0.8620 & 0.9455 & 0.8675 & 0.8252  & 0.7424 & 0.9101 & 0.8947         & 0.9742      & 0.8125   & $\bold{0.9765}$ \\ \hline
			\multirow{3}{*}{K-means}     & OA    & 0.8639 & 0.8851 & 0.8905 & 0.8277  & 0.7158 & 0.8371 & 0.8921         & 0.8581      & 0.6350   & $\bold{0.9122}$ \\
			& AA    & 0.8365 & 0.8328 & 0.8611 & 0.7496  & 0.6672 & 0.8070 & 0.8405         & 0.8199      & 0.7533   & $\bold{0.8661}$ \\
			& Kappa & 0.8510 & 0.8738 & 0.8799 & 0.8112  & 0.6879 & 0.8167 & 0.8817         & 0.8435      & 0.6088   & $\bold{0.9037}$ \\ \hline \hline
		\end{tabular}
	}
\end{table*}

\begin{table*}[htbp]
	\centering
	\caption{Comparison of classification accuracy of different sampling methods on PaviaU dataset}
	\label{tab:paviau}
	\resizebox{.75\textwidth}{!}{
		\begin{tabular}{cc|cccccccccc}
			\hline \hline
			&       & 1DCNN  & 2DCNN  & 3DCNN  & MS3DCNN & RNN    & ViT    & SpectralFormer & MorphFormer & SSFTTnet        & SaaFormer        \\ \hline
			\multirow{3}{*}{Random}     & OA    & 0.8820 & 0.9935 & 0.9121 & 0.9736  & 0.9057 & 0.9480 & 0.9625         & 0.9932      & 0.9947          & $\bold{0.9980}$  \\
			& AA    & 0.8619 & 0.9847 & 0.9130 & 0.9580  & 0.8833 & 0.9331 & 0.9499         & 0.9869      & 0.9914          & $\bold{0.9958}$  \\
			& Kappa & 0.8416 & 0.9814 & 0.8825 & 0.9651  & 0.8743 & 0.9312 & 0.9503         & 0.9909      & 0.9930          & $\bold{0.9973} $ \\ \hline
			\multirow{3}{*}{Block}      & OA    & 0.7868 & 0.7307 & 0.9391 & 0.9673  & 0.9223 & 0.9462 & 0.8992         & 0.8703      & 0.8309          & $\bold{0.9747}$  \\
			& AA    & 0.8405 & 0.7774 & 0.9086 & 0.9261  & 0.8774 & 0.8993 & 0.8188         & 0.8935      & 0.8163          & $\bold{0.9695}$  \\
			& Kappa & 0.6934 & 0.6216 & 0.9026 & 0.9477  & 0.8767 & 0.9160 & 0.8405         & 0.8182      & 0.7671          & $\bold{0.9634}$  \\ \hline
			\multirow{3}{*}{Checkerboard} & OA    & 0.8908 & 0.9211 & 0.9011 & 0.9499  & 0.8102 & 0.8837 & 0.8775         & 0.9419      & 0.9483          & $\bold{0.9666}$  \\
			& AA    & 0.8970 & 0.8894 & 0.9055 & 0.9558  & 0.7894 & 0.8835 & 0.8992         & 0.9305      & $\bold{0.9658}$ & 0.9641           \\
			& Kappa & 0.8510 & 0.8937 & 0.8665 & 0.9322  & 0.7412 & 0.8427 & 0.8361         & 0.9218      & 0.9309          & $\bold{0.9543}$  \\ \hline
			\multirow{3}{*}{K-means}     & OA    & 0.8163 & 0.9071 & 0.8386 & 0.8403  & 0.7992 & 0.8498 & 0.8220         & 0.8498      & 0.8275          & $\bold{0.9559}$  \\
			& AA    & 0.8664 & 0.8892 & 0.8737 & 0.9259  & 0.8445 & 0.8841 & 0.8723         & 0.8810      & 0.8507          & $\bold{0.9518}$  \\
			& Kappa & 0.7784 & 0.8861 & 0.8068 & 0.7884  & 0.7490 & 0.8192 & 0.7594         & 0.8200      & 0.7941          & $\bold{0.9420}$  \\ \hline \hline
		\end{tabular}
	}
\end{table*}

\begin{figure*}[htbp]
	\centering
	\subcaptionbox{\label{oa}}{\includegraphics[width=0.8\textwidth]{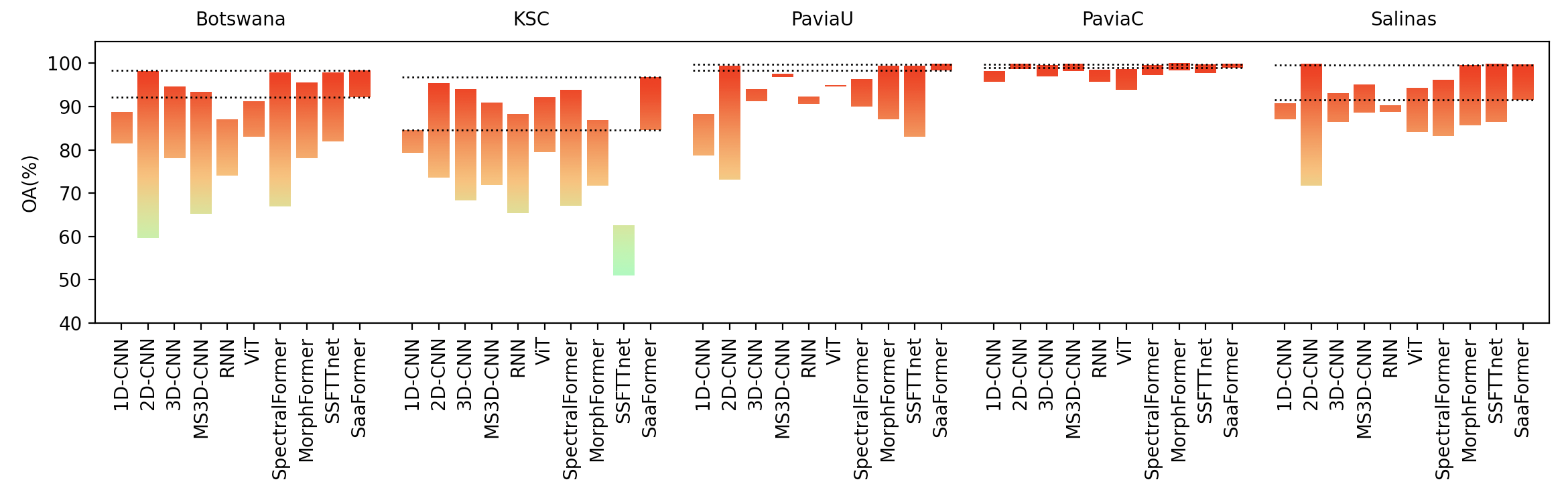}}
	\hfill
	\subcaptionbox{\label{aa}}{\includegraphics[width=0.8\textwidth]{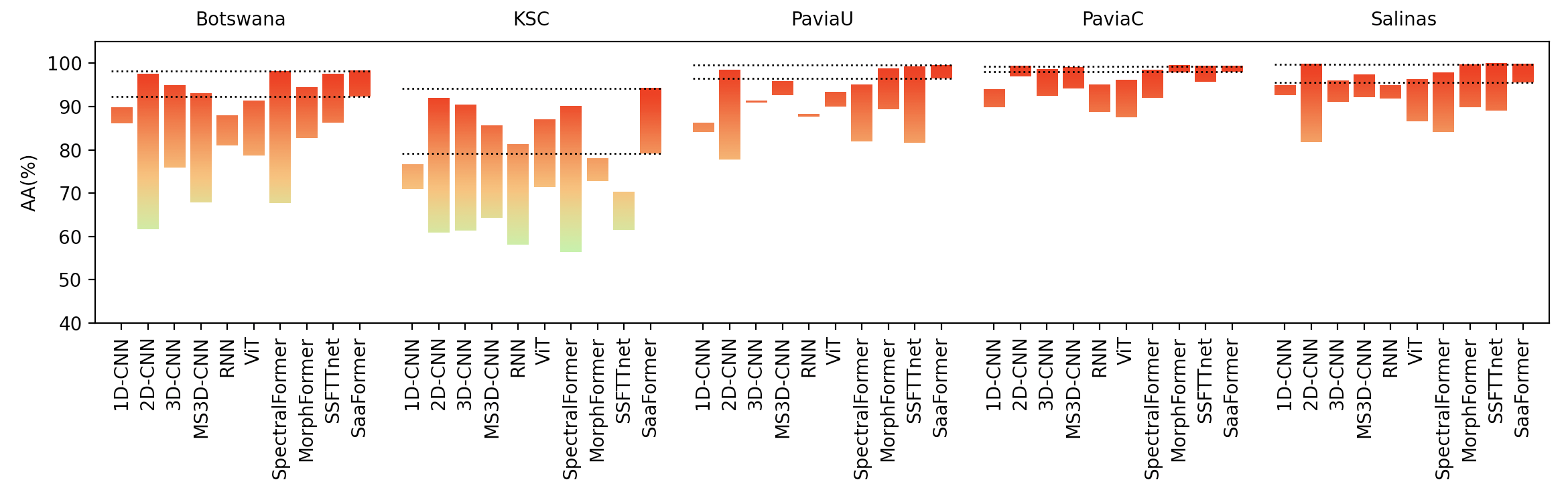}}
	\hfill
	\subcaptionbox{\label{ka}}{\includegraphics[width=0.8\textwidth]{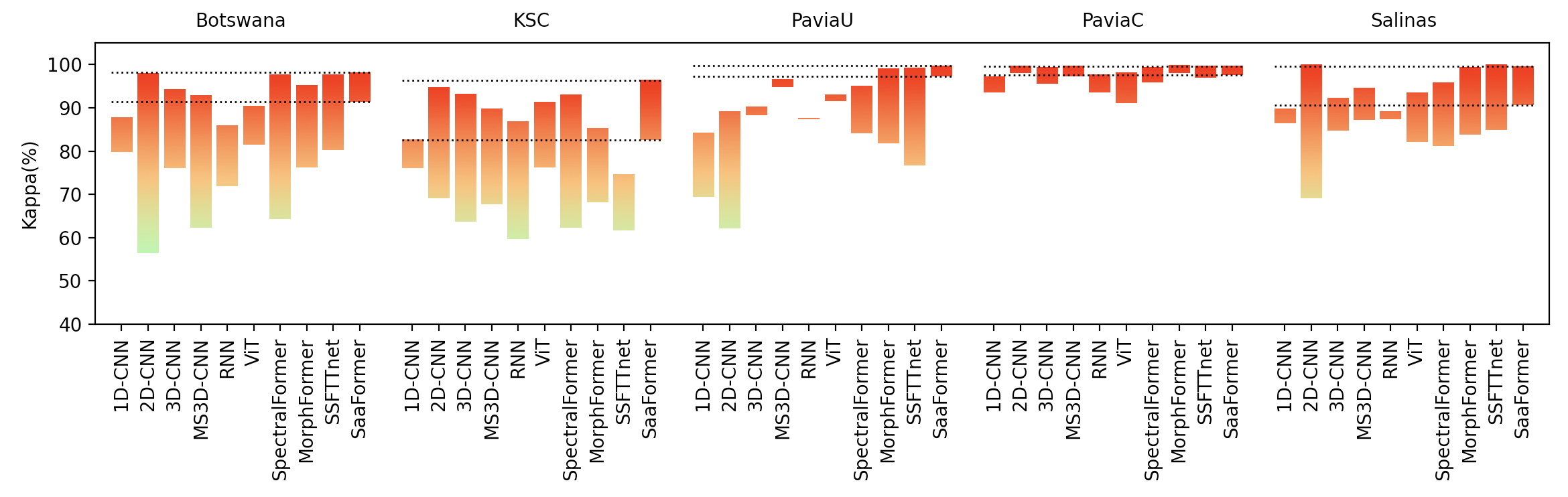}}
	\caption{Comparison of the results of different models on the three indicators of OA, AA and Kappa on the five datasets. The top and bottom of the bars correspond to the model’s performance metrics when employing random sampling and block-wise sampling.}
	\label{Fig:re}
\end{figure*}

\begin{figure*}[htbp]
	\centering
	\subcaptionbox{\label{oa1}}{\includegraphics[width=0.8\textwidth]{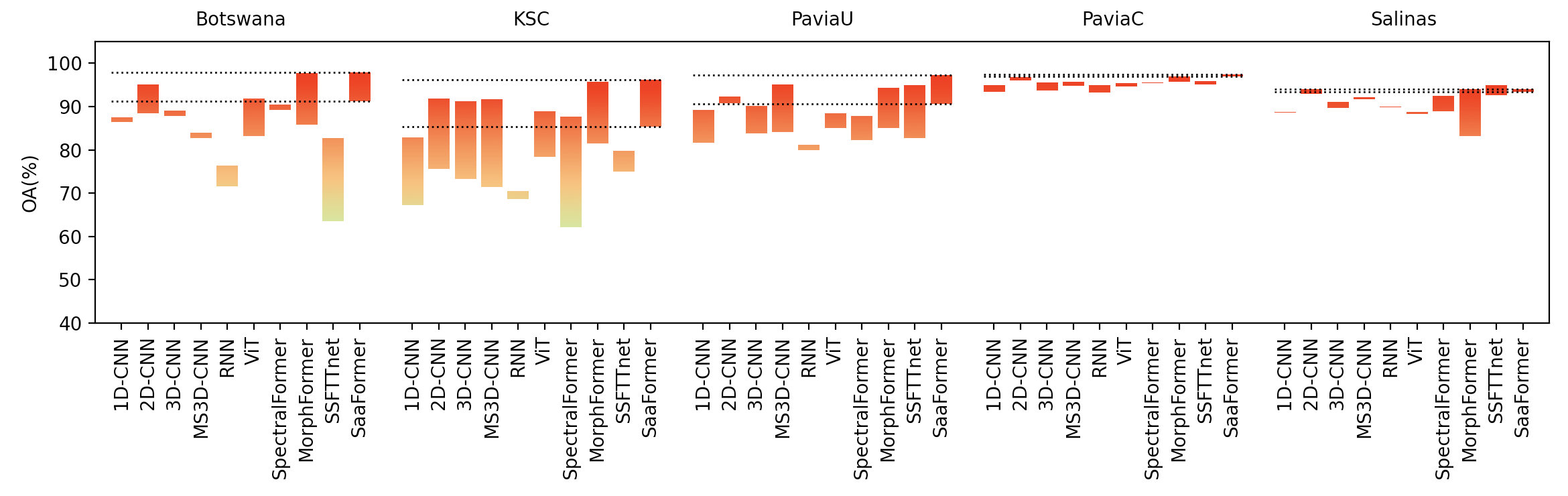}}
	\quad
	\subcaptionbox{\label{aa1}}{\includegraphics[width=0.8\textwidth]{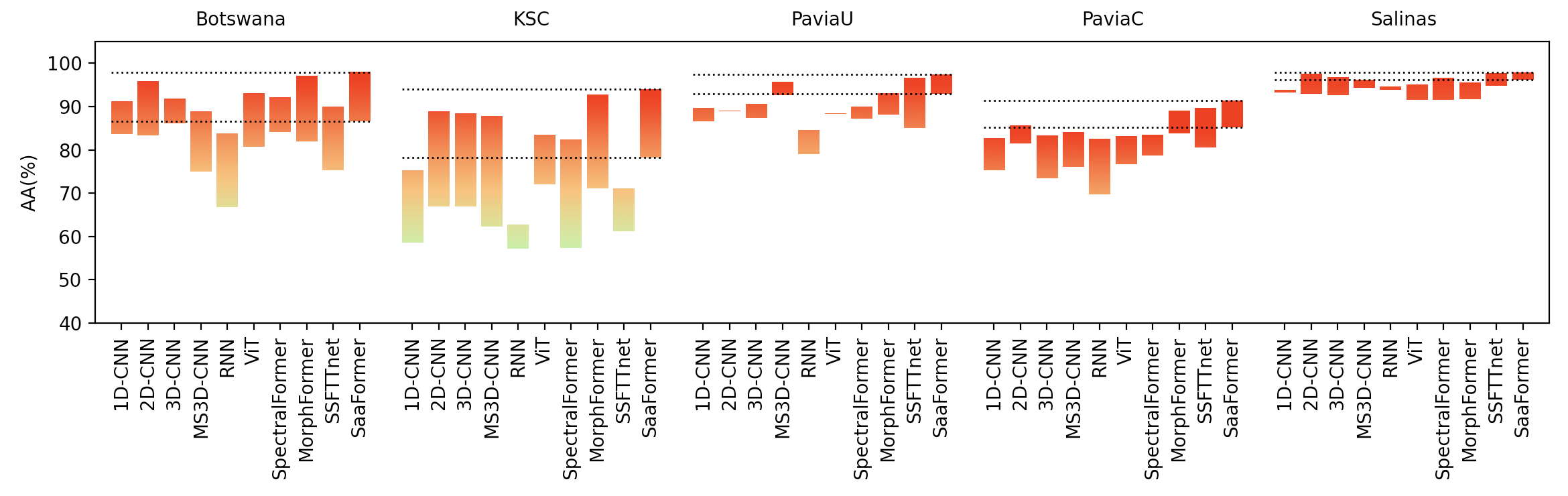}}
	\quad
	\subcaptionbox{\label{ka1}}{\includegraphics[width=0.8\textwidth]{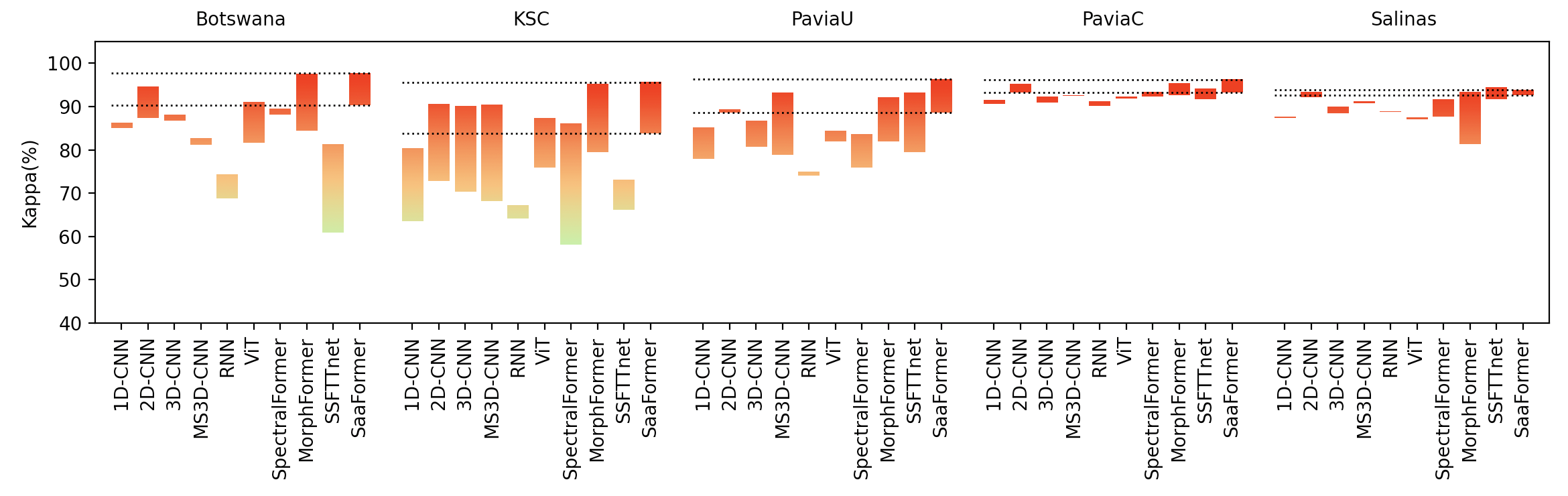}}
	\caption{Comparison of the results of different models on the three indicators of OA, AA and Kappa on the five datasets. The top and bottom of the bars represent the model’s metrics when applying checkerboard sampling and k-means sampling.}
	\label{Fig:re1}
\end{figure*}

\subsection{Data Partitions}

\subsubsection{Random Sampling}
$5\%$ of the samples from each class are randomly selected as the training set, and the remaining samples are used for test. Each sample is $s\times s$ pixels patch.

\subsubsection{Checkerboard  Sampling}

We divide the entire dataset into an equal number of blocks based on the height and width, for example, if we divide the whole dataset ($H\times W$) into $c\times c$ blocks i.e. [1,2,...., $c\times c$], each small block would have a size of $\frac{H}{c} \times \frac{W}{c}$. Then, using the interleaving method, we allocate 50\% of the blocks as the training set and the remaining blocks as the test set. Specifically, let \( N_1 \) denote the total number of sample points in the odd sequence block and \( N_2 \) denote the total number of sample points in the even sequence block. If \( N_1 < N_2 \), then \( N_1 \) will be used as the training set, otherwise, \( N_2 \) will be used as the training set.


\subsubsection{Block-wise Sampling}

Taking a different approach from the checkerboard sampling, we opt to divide the dataset based on the shorter dimension. The dataset ($H>W$) is split into $b$ portions i.e. [1,2,...,b], each portion would have a size of $H\times W/b$. We set \( b \) is the minimum value that ensures all classes are represented in both the training and testing sets. And the selection of the training set is the same as before.


\subsubsection{K-means Sampling}
Predefining the number of cluster centers $k$, we allocate all samples from different categories in the dataset to these $k$ clusters using the k-means clustering method. Afterwards, we select $k/2$ clusters as training set, and the remaining clusters as the test set. And we also select a smaller sample area as the training set.


It is noteworthy that chessboard sampling techniques have already been applied in neural network sampling methods, as discussed in \cite{ref23}. However, the main objective and application context of the sampling method is different from those of ours. Specifically, Arada \textit{et al.}\cite{ref23} introduced the checkerboard sampling aimed at addressing database memory issues. In contrast, our method focuses on addressing data leakage issues between training and test samples. Unlike the method in \cite{ref23}, which reduces the number of fingerprint templates in the database by performing checkerboard sampling on each sample image, our strategy involves directly applying checkerboard sampling across the whole dataset to effectively reduce the overlap between training and test samples.

	
Moreover, our proposed chessboard sampling differs significantly from block sampling in its handling of boundary samples. Checkerboard sampling discards boundary samples to prevent overlap between training and test samples, while block-wise sampling deal with the overlapping boundary regions with zero padding. By employing this differentiated sampling approach, we can more effectively validate the generalization capability and robustness of the proposed model. Due to significant differences in data distribution across various datasets, Table \ref{tab:sample_num} shows the sample numbers for the training and test sets for the three data partition methods.


\begin{figure*}[htbp]
	\centering
	\begin{subfigure}{0.7\textwidth}
		\centering
		\includegraphics[width=\textwidth,height=0.22\textheight]{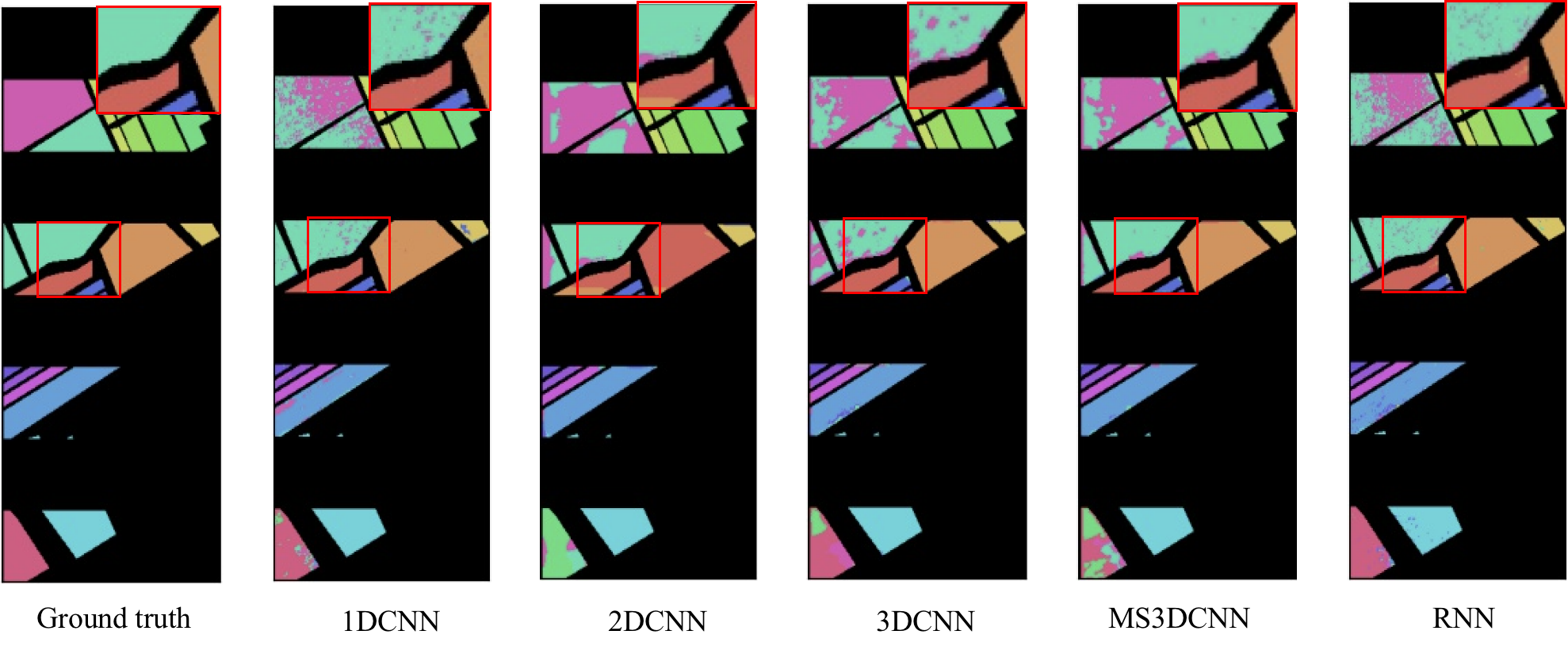}
	\end{subfigure}
	\begin{subfigure}{0.7\textwidth}
		\centering
		\includegraphics[width=\textwidth,height=0.22\textheight]{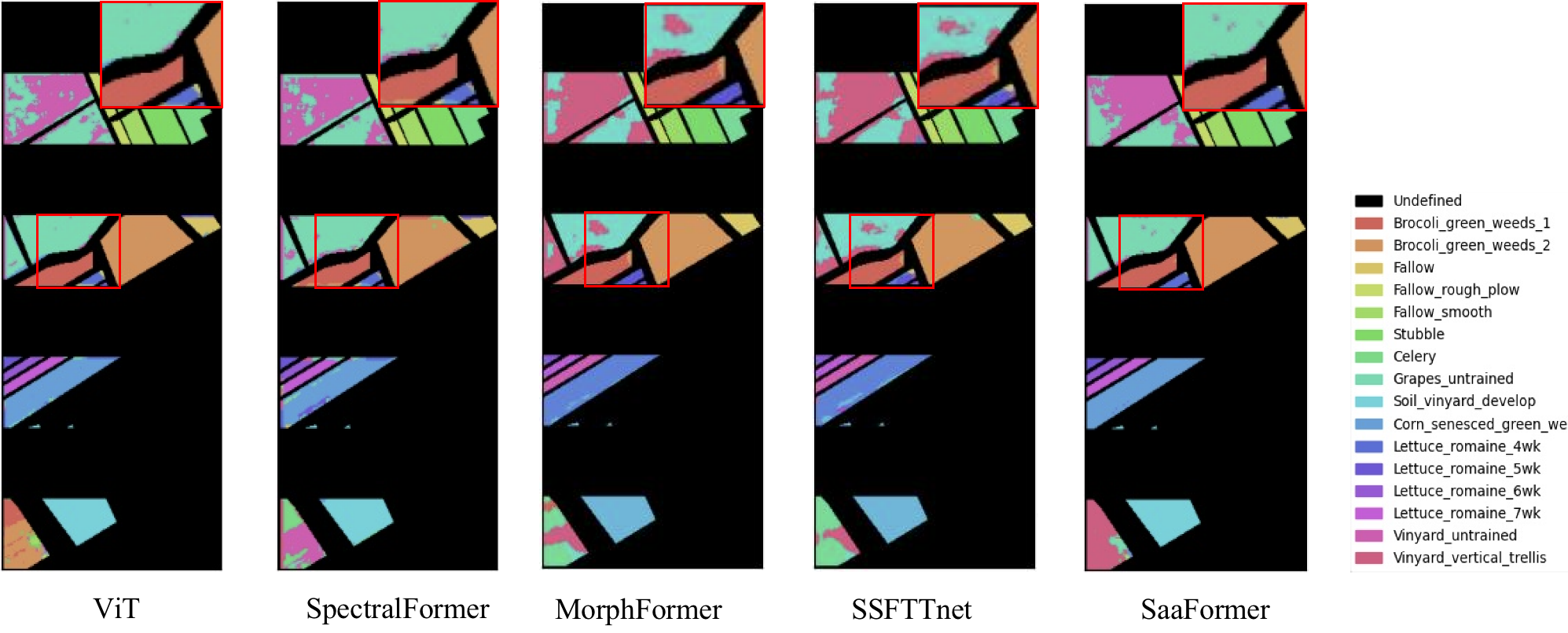}
	\end{subfigure}
	\caption{Spatial distribution of testing sets using block-wise sampling, and the classification maps obtained by different models on the Salinas dataset}
	\label{fig:salinas}
\end{figure*}


\begin{figure*}[htbp]
	\centering
	\begin{subfigure}{0.7\textwidth}
		\centering
		\includegraphics[width=\textwidth]{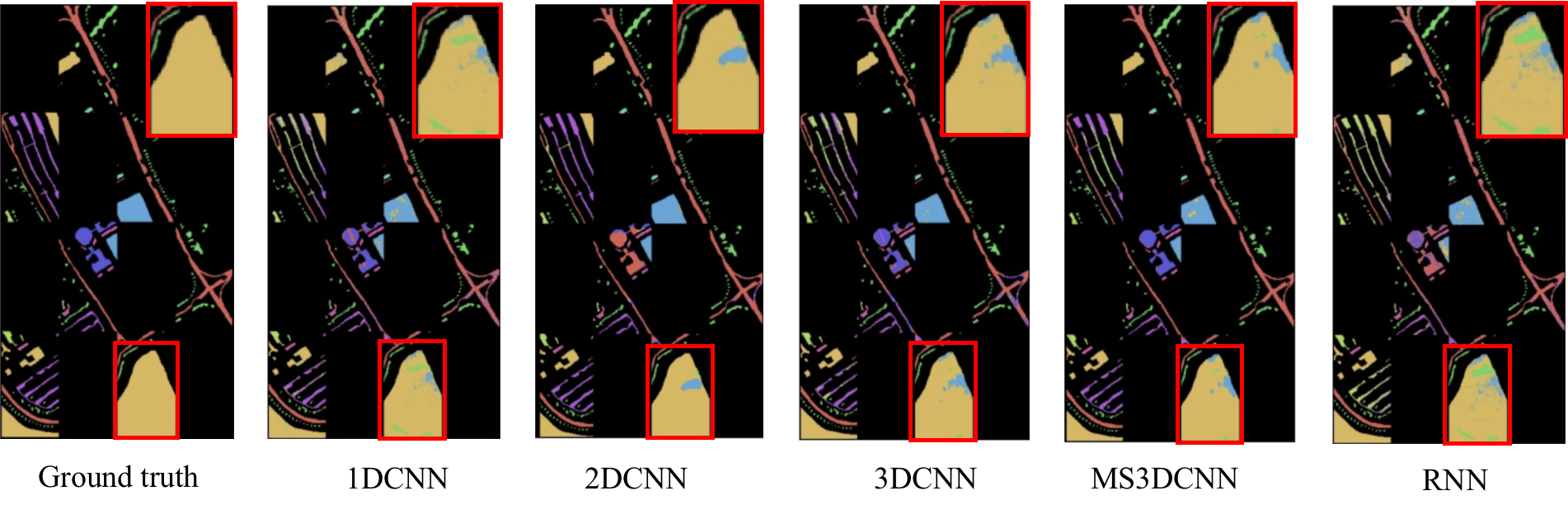}
	\end{subfigure}\\ \vspace{0.1cm} 
	\begin{subfigure}{0.7\textwidth}
		\centering
		\includegraphics[width=\textwidth]{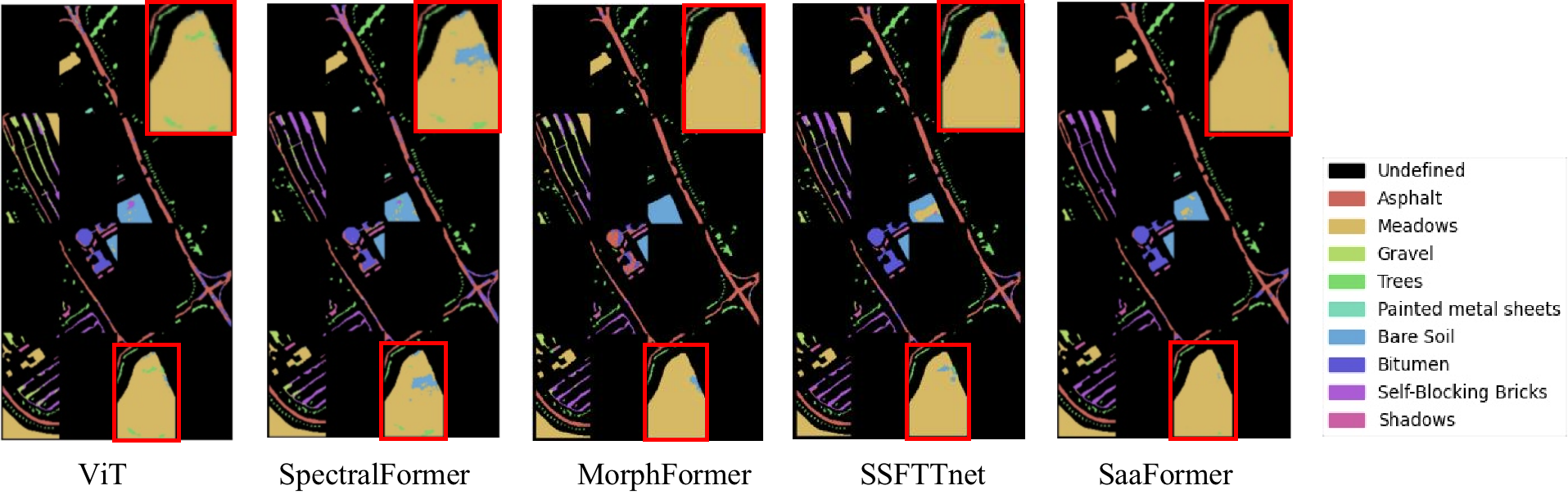}
	\end{subfigure}
	\caption{Spatial distribution of testing sets using random sampling, and the classification maps obtained by different models on the PaviaU dataset}
	\label{fig:paviau_c}
\end{figure*}
\subsection{Experimental Setting}

The experiments will utilize various classification measures as outlined in \cite{ref24}. A summary of these measures is provided below.
\begin{itemize}
\item $C_m$ represents the set of data samples in the $m$th class as determined by the ground truth.
\item $n_{mm}$ represents the number of samples in the $m$-th class that are correctly classified into the $m$-th class $\hat C_m$.
\item $n_{jm}$ represents the number of data samples in the $m$-th class $C_m$ that are actually classified into the $j$-th class $\hat C_{j}$.
\item $\hat n_{mj}$ represents the number of samples from the $j$-th class that are classified as the $m$-th class
\item $M$ represents the number of classes.
\item $n_m = \sum_{j=1}^{M} n_{jm}$ denotes the number of data samples in the $m$-th class $C_m$.
\item $\hat{n}_m=\sum_{m=1}^M \hat n_{mj}$ denotes the total number of data samples that be classified into $m$-th class
\item $N= \sum_{m=1}^M n_{m}$ represents the total number of data samples.
\item $\hat N=\sum_{m=1}^M \hat n_{m}$ denotes the total number of data samples that is  classified.
\end{itemize}

\subsubsection{Evaluation Metrics}
In HSIC, commonly used performance measurements include class accuracy (CA) and overall accuracy (OA), Average Accuracy (AA), and $\kappa$ (Kappa), which can be expressed as
\begin{align}
P_{OA}=\frac{1}{N}\sum_{m=1}^M n_{mm}=\sum_{m=1}^M\frac{n_m}{N} p_{A}(C_m)
\end{align}
where $p_{A}(C_m)=\frac{n_{mm}}{\sum_{j=1}^{M} n_{jm}}=\frac{n_{mm}}{n_m}$ is the accuracy of the $m$-th class.  $P_{OA}$ employs sample ratios as weights to calculate the average accuracy of each class. Additionally, we further consider all classes to have equal likelihood i.e.  $P_{AA}$ as the average accuracy.
\begin{align}
P_{AA}=\frac{1}{M}\sum_{m=1}^M \frac{n_{mm}}{n_m}=\frac{1}{M}\sum_{m=1}^Mp_{A}(C_m)
\end{align}
Moreover, the kappa coefficient is a measurement indicator used to assess the consistency between model predictions and true labels in deep learning classification tasks. It takes into account the possibility of random agreement in the predicted results and compares the model's predictions with random chance agreement. 
\begin{align}
	\kappa =\frac{P_{OA}-p_e}{1-p_e} = \frac{N\sum_{m=1}^{M} n_{mm}-\sum_{j=1}^{M}n_m\sum_{j=1}^{M}n_{jm}}{N^{2}-\sum_{m=1}^{M}n_m\sum_{j=1}^{M}n_{jm}}
\end{align}
where  $\kappa \in [-1,1]$, 1 represents perfect agreement, 0 indicates that the predictions are as consistent as would be expected by chance, and -1 represents complete disagreement. A higher kappa coefficient indicates better classification performance of the model.

\subsubsection{Comparison with State-of-the-Art Backbone Networks} 
To assess the classification performance of the proposed model, we compared it with some established methods for hyperspectral image classification. These methods include 1DCNN Classifier\cite{ref8}, 2DCNN-LR\cite{ref15}, 3-DCNN-based FE model\cite{ref16},  M3D-DCNN\cite{ref17}, RNN\cite{ref9}, as well as four Transformer-based models, the traditional ViT \cite{ref18}, SpectralFormer \cite{ref19}, morphFormer \cite{ref19_1} and SSFTTnet\cite{ref19_4}. The parameter configurations for these network structures and the training sample size is set according to the default settings specified in their respective references.

As for  SaaFormer architecture, it employs an embedded spectrum with $d=128$ channels. The embedding is further segmented into $K=2$, $c_1=128$, $c_2=32$ spectral clips, which is then processed through two consecutive transformer encoder blocks for classification. Each encoder block includes a four-head axial aggregation attention layer, a MLP layer and a GELU nonlinear activation layer. To enhance model robustness, a dropout layer is applied after encoding positional embeddings and within the MLPs, with a 40\% dropout rate. And we use the Adam optimizer \cite{ref22} with a batch size of 128 to train the model with a weight decay of 5e-3 and learning rate of 5e-4. In addition, these methods used a step scheduler with a gamma of 0.9, steps of size tenth of the total epochs, and trained during about 500 epochs.

\subsection{Generalization Validation}

To assess the generalization capabilities of the proposed model and several popular models, experiments are conducted on the mentioned datasets. The evaluation is based on classification results, using metrics OA, AA, and $\kappa$.


The visualization illustrates the changes in accuracy of the classification model across various sampling strategies using bar charts. The length of each bar indicates the magnitude of accuracy variation, with longer bars indicating more significant fluctuations and shorter bars suggesting more stable performance. Specifically, in Fig. \ref{Fig:re}, the top and bottom of the bars correspond to the model’s performance metrics when employing random sampling and block-wise sampling, respectively. On the other hand, in Fig. \ref{Fig:re1}, the top and bottom of the bars represent the model’s metrics when applying k-means sampling and checkerboard sampling.


The results from Fig. \ref{Fig:re} and Fig. \ref{Fig:re1} underscore the high accuracy of classification results obtained via random sampling for partitioning. This further corroborates the presence of data leakage in such partitioning schemes, leading to overfitting of the models.There exist significant disparities among different datasets, resulting in varied performance across models. Botswana and KSC datasets, characterized by their limited data volume and sparse distribution, exhibit considerable fluctuations in model accuracies. Notably, the 2DCNN models are more susceptible to spatial distribution effects, whereas 1DCNN models emphasize spectral features, thus showing relatively smoother fluctuations. Additionally, the poorer performance of SSFTTnet on the KSC dataset may be attributed to the adverse impact of principal component analysis on spectral information.In contrast, the latter three datasets, with larger volumes and denser distributions, demonstrate overall less variability. Under random sampling, MorphFormer, SSFTTNet, and SaaFormer achieve remarkably high accuracies, albeit with significant fluctuations when partitioning methods are altered. In comparison, our model exhibits greater stability, maintaining optimal performance even under alternative partitioning strategies.In summary, these visualizations eloquently highlight the superiority of our model, which not only excels in accuracy but also demonstrates exceptional generalization capabilities and robustness.

\begin{figure}[htbp]
	\centering
	\includegraphics[width=0.5\textwidth]{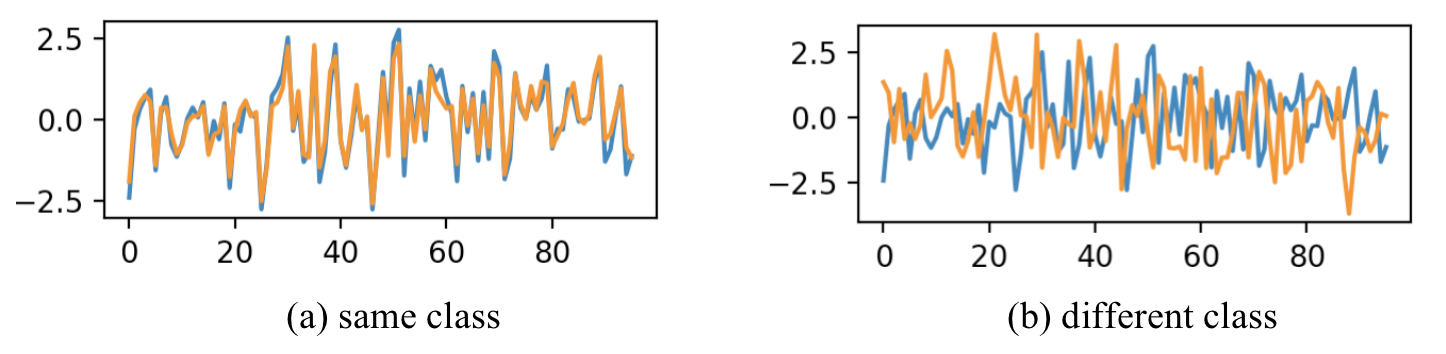}
	\caption{Visualization the spectral curves of samples feature maps after passing through the Axial Transformer Block on PaviaU dataset. The orange and blue curves in the figure represent the spectral characteristic plots of different sample points in the test set. The blue curves in both (a) and (b) correspond to the same test sample. In (a), the orange curve represents the spectral characteristic plot of a sample belonging to the same category as the blue curve, while in (b), the orange curve represents the spectral characteristic plot of a sample from a different category than the blue curve.}
	\label{Fig:keshihua}
\end{figure}




Tables \ref{tab:botswana} and \ref{tab:paviau} present detailed result regarding the three evaluation metrics on the Botswana and PaviaU datasets. The best accuracy is highlighted in bold. Our model consistently achieves the highest accuracy in nearly all cases. Moreover, it's noteworthy that the variations in the three evaluation metrics among different partitioning methods are minimal, typically within a 1\% range. This highlights that our model consistently delivers high accuracy across diverse datasets and remains stable under various partitioning methods, underscoring its exceptional generalization and robustness.

\subsection{Model Analysis}

\subsubsection{Ablation Study}


We first conducted experiments to determine the optimal length for spectral clips. With $c_1=128$ and $K=2$ fixed, we evaluated spectral clips of lengths 8, 16, 32, and 64 for $c_2$. Experiments were performed on four datasets, and the specific experimental results are shown in Table \ref{tab:ablation0}.


\begin{table*}[htbp]
	\centering
	\caption{Classification performance analysis of different spectral clip lengths across various datasets. With \( c_1 = 128 \) fixed, spectral clips of lengths 8, 16, 32, and 64 for \( c_2 \) were evaluated.}
	\label{tab:ablation0}
	\resizebox{\textwidth}{!}{
		\begin{tabular}{cc|cccc|cccc|cccc|cccc}
			\hline \hline
			\multicolumn{2}{c|}{\multirow{2}{*}{}} & \multicolumn{4}{c|}{random}                         & \multicolumn{4}{c|}{block-wise}                           & \multicolumn{4}{c|}{k-means}                         & \multicolumn{4}{c}{chessboard}                      \\
			\multicolumn{2}{c|}{}        & $c_2=8$    & $16$            & $32$            & $64$   & $c_2=8$     & $16$            & $32$            & $64$   & $c_2=8$             & $16$   & $32$            & $64$   & $c_2=8$    & $16$   & $32$            & $64$            \\ \hline
			\multirow{3}{*}{Botswana}    & OA      & 0.983 & \pmb{0.993} & 0.989          & 0.987 & 0.909  & \pmb{0.912} & 0.890          & 0.892 & 0.838          & 0.874 & \pmb{0.888} & 0.855 & 0.921 & 0.933 & 0.939          & \pmb{0.963} \\
			& AA      & 0.985 & \pmb{0.993} & 0.991          & 0.989 & 0.913  & \pmb{0.920} & 0.900          & 0.892 & 0.876          & 0.884 & \pmb{0.893} & 0.825 & 0.944 & 0.954 & 0.938          & \pmb{0.940} \\
			& Kappa   & 0.981 & \pmb{0.992} & 0.988         & 0.986 & 0.900  & \pmb{0.903} & 0.879          & 0.882 & 0.819          & 0.860 & \pmb{0.875} & 0.837 & 0.913 & 0.926 & 0.933          & \pmb{0.959} \\ \hline
			\multirow{3}{*}{KSC}         & OA      & 0.904 & 0.918          & \pmb{0.930} & 0.929 & 0.842  & 0.846          & \pmb{0.867} & 0.838 & \pmb{0.811} & 0.808 & 0.794          & 0.781 & 0.960 & 0.956 & \pmb{0.964} & 0.954          \\
			& AA      & 0.834 & 0.848          & \pmb{0.873} & 0.883 & 0.814  & 0.815          & \pmb{0.818} & 0.805 & \pmb{0.738} & 0.737 & 0.711          & 0.691 & 0.933 & \pmb{0.939} & 0.936 & 0.927          \\
			& Kappa   & 0.893 & 0.909          & \pmb{0.922} & 0.920 & 0.825  & 0.825          & \pmb{0.848} & 0.816 & \pmb{0.791} & 0.787 & 0.771          & 0.757 & 0.955 & 0.950 & \pmb{0.959} & 0.95          \\ \hline
			\multirow{3}{*}{PaviaU}      & OA      & 0.991 & 0.992         & \pmb{0.992} & 0.992 & 0.972  & 0.972          & \pmb{0.982} & 0.970 & 0.923          & 0.918 & \pmb{0.956} & 0.926 & 0.958 & 0.959 & \pmb{0.967} & 0.965          \\
			& AA      & 0.987 & 0.989          & \pmb{0.988} & 0.988 & 0.968  & \pmb{0.968}          & 0.967 & 0.969 & 0.942          & 0.940 & \pmb{0.952} & 0.934 & 0.964 & 0.963 & \pmb{0.964} & 0.971          \\
			& Kappa   & 0.989 & 0.989          & \pmb{0.990} & 0.989 & 0.959 & 0.959         & \pmb{0.972} & 0.956 & 0.901          & 0.894 & \pmb{0.942} & 0.903 & 0.943 & 0.944 & \pmb{0.954} & 0.953          \\ \hline
			\multirow{3}{*}{Salinas}     & OA      & 0.965 & 0.966          & \pmb{0.975} & 0.971 & 0.908  & 0.908          & \pmb{0.925} & 0.900 & 0.888          & 0.887 & \pmb{0.895} & 0.892 & 0.917 & 0.918 & \pmb{0.922} & 0.917          \\
			& AA      & 0.985 & 0.981          & \pmb{0.988} & 0.986 & 0.954  & 0.953          & \pmb{0.964} & 0.954 & 0.949          & 0.949 & \pmb{0.950} & 0.948 & 0.947 & 0.947 & \pmb{0.949} & 0.946          \\
			& Kappa   & 0.961 & 0.962          & \pmb{0.972} & 0.967 & 0.896  & 0.897          & \pmb{0.915} & 0.888 & 0.875          & 0.874 & \pmb{0.883} & 0.878 & 0.908 & 0.909 & \pmb{0.914} & 0.908          \\ \hline \hline
		\end{tabular}
    }

\end{table*}

We further investigate the optimal number $K$ of the Multi-level Spectral Extraction. For each level, we preset $c_1=128$, $c_2=64$, $c_3=32$, $c_4=16$ to extract different granular spectral features. For different datasets, we set $K$ according to the optimal spectral clips in the Table \ref{tab:ablation1}. The results indicates that our model exhibits higher accuracy when incorporating Multi-level Spectral Extraction components ($K=2, 3$), compared to relying solely on the complete spectral information ($K=1$). It indicates the effectiveness of the Multi-level Spectral Extraction structure in capturing underlying patterns and correlations within the spectral data.

\begin{table*}[htbp]
	\centering
	\caption{Ablation analysis of the proposed Saaformer with a combination of different structures on different datasets.}
	\label{tab:ablation1}
	\resizebox{.85\textwidth}{!}{
		\begin{tabular}{cc|ccc|ccc|ccc|ccc}
			\hline \hline
			& \multirow{2}{*}{\begin{tabular}[c]{@{}c@{}}\\ $K$\end{tabular}} & \multicolumn{3}{c|}{random}                         & \multicolumn{3}{c|}{block-wise}                     & \multicolumn{3}{c|}{k-means}                         & \multicolumn{3}{c}{chressborad}                     \\
			&                                                                          & OA              & AA              & kappa           & OA              & AA              & kappa           & OA              & AA              & kappa           & OA              & AA              & kappa           \\ \hline
			\multirow{4}{*}{\begin{tabular}[c]{@{}c@{}}Botswana\end{tabular}} & 1                                                                        & 0.9708          & 0.9744          & 0.9684          & 0.9041          & 0.9184          & 0.8947          & 0.8154          & 0.8330          & 0.7944          & 0.8925          & 0.9398          & 0.8829          \\
			& 2                                                                        & \pmb{0.9925} & \pmb{0.9933} & \pmb{0.9919} & 0.9117          & 0.9202          & 0.9027          & \pmb{0.9122} & \pmb{0.8661} & \pmb{0.9037} & \pmb{0.9631} & \pmb{0.9400} & \pmb{0.9594} \\
			& 3                                                                        & 0.9752          & 0.9747          & 0.9724          & 0.9098          & 0.9168          & 0.9009          & 0.8726          & 0.8722          & 0.8584        & 0.9012          & 0.9386          & 0.8924          \\
			& 4                                                                        & 0.9646          & 0.9644          & 0.9617          & \pmb{0.9135} & \pmb{0.9204} & \pmb{0.9050} & 0.8667          & 0.8709          & 0.8508          & 0.9213          & 0.9346          & 0.9137          \\ \hline
			\multirow{4}{*}{\begin{tabular}[c]{@{}c@{}}KSC\end{tabular}}      & 1                                                                        & 0.9048          & 0.8424          & 0.8939          & 0.8350          & 0.8123          & 0.8128          & 0.7748          & 0.7049          & 0.7443          & 0.9432          & 0.9205          & 0.9355          \\
			& 2                                                                        & \pmb{0.9303} & \pmb{0.8730} & \pmb{0.9223} & \pmb{0.8668} & \pmb{0.8176} & \pmb{0.8475} & \pmb{0.8114} & \pmb{0.7383} & \pmb{0.7907} & \pmb{0.9638} & \pmb{0.9357} & \pmb{0.9588} \\
			& 3                                                                        & 0.9125          & 0.8629          & 0.9026          & 0.8556          & 0.8153          & 0.8354          & 0.7973          & 0.7154          & 0.7747          & 0.9579          & 0.9301          & 0.9521          \\
			& 4                                                                        & 0.9107          & 0.8561          & 0.9006          & 0.8218          & 0.7344          & 0.7968          & 0.7591          & 0.6582          & 0.7323          & 0.9510          & 0.9316          & 0.9433          \\ \hline
			\multirow{4}{*}{\begin{tabular}[c]{@{}c@{}}PaviaU\end{tabular}}   & 1                                                                        & 0.9898          & 0.9843          & 0.9865          & 0.9750          & 0.9654          & 0.9637          & 0.8898          & 0.9359          & 0.8594          & 0.9063          & 0.9259          & 0.8722          \\
			& 2                                                                        & \pmb{0.9921} & \pmb{0.9881} & \pmb{0.9895} & \pmb{0.9824} & 0.9673          & \pmb{0.9720} & \pmb{0.9559}          & \pmb{0.9518}          & \pmb{0.9420}          & \pmb{0.9666} & \pmb{0.9641} & \pmb{0.9543} \\
			& 3                                                                        & 0.9877          & 0.9791          & 0.9837          & 0.9764          & \pmb{0.9684} & 0.9659          & 0.9133 			& 0.9262		 & 0.8870 			& 0.9144          & 0.8910          & 0.8831          \\
			& 4                                                                        & 0.9910          & 0.9858          & 0.9881          & 0.9682          & 0.9663          & 0.9536          & 0.8924          & 0.9181          & 0.8611          & 0.9066          & 0.9272          & 0.8738          \\ \hline
			\multirow{4}{*}{\begin{tabular}[c]{@{}c@{}}PaviaC\end{tabular}}   & 1                                                                        & 0.9948          & 0.9815          & 0.9927          & 0.9878          & 0.9767          & 0.9837          & 0.9695          & 0.9052          & 0.9557          & 0.9599          & 0.8511          & 0.9324          \\
			& 2                                                                        & \pmb{0.9976} & \pmb{0.9915} & \pmb{0.9966} & \pmb{0.9908} & \pmb{0.9795} & \pmb{0.9877} & \pmb{0.9713} & \pmb{0.9180} & \pmb{0.9583} & 0.9559          & 0.8404          & 0.9257          \\
			& 3                                                                        & 0.9956          & 0.9862          & 0.9938          & 0.9889          & 0.9768          & 0.9851          & 0.9695          & 0.9091          & 0.9556          & \pmb{0.9610} & \pmb{0.8593} & \pmb{0.9344} \\
			& 4                                                                        & 0.9959          & 0.9845          & 0.9942          & 0.9834          & 0.9757          & 0.9770          & 0.9685          & 0.9055          & 0.9542          & 0.9586          & 0.8535          & 0.9303          \\ \hline
			\multirow{4}{*}{\begin{tabular}[c]{@{}c@{}}Salinas\end{tabular}}  & 1                                                                        & 0.9664          & 0.9863          & 0.9626          & 0.8847          & 0.9419          & 0.8704          & 0.8893          & 0.9481          & 0.8763         & 0.9116          & 0.9402          & 0.9019          \\
			& 2                                                                        & \pmb{0.9749} & \pmb{0.9878} & \pmb{0.9721} & \pmb{0.9246} & \pmb{0.9637} & \pmb{0.9154} & 0.8951 & 0.9503 & 0.8827 & \pmb{0.9222} & \pmb{0.9490} & \pmb{0.9136} \\
			& 3                                                                        & 0.9734          & 0.9874          & 0.9703          & 0.9088          & 0.9525          & 0.8975           & \pmb{0.8963}          & \pmb{0.9517}          & \pmb{0.8837}          & 0.9155          & 0.9395          & 0.9060          \\
			& 4                                                                        & 0.9714          & 0.9866          & 0.9681          & 0.9030          & 0.9501          & 0.8911          & 0.8871          & 0.9428          & 0.8739         & 0.9147          & 0.9419          & 0.9050          \\ \hline \hline
		\end{tabular}
	}
\end{table*}

\subsubsection{Computational Complexity}

For a given HS image, let $X\in \mathbb{R}^{h\times w\times c}$ be the input sample.  The per-layer times complexity of the proposed SaaFormer is dominated by axial self-attention and the spatial-enhanced modules. When traditional attention module is applied on a feature map of $x$, the time complexity can be $\mathcal O(3h^2w^2d)$. However, for the axial aggregation attention, time complexity is $\mathcal{O}(3(h+w)d+2(h^2+w^2)d)$. Thus, our axial aggregation attention can successfully reduce time complexity. And the spatial-enhanced modules takes $\mathcal{O}(k^2hwd^2)$, where $k$ is the size of convolution kernel and $d$ is the size of the hidden features. Furthermore, we compared the computational complexities of models with different Spectral Extraction levels $K$ in Table \ref{tab:compute}.

\begin{table}[htbp]
	\centering
	\caption{Computational Complexity studies on components in Multi-level Spectral Extraction on four datasets (PaviaU and PaviaC datasets have the same computational complexity).}
	\label{tab:compute}
	\resizebox{.48\textwidth}{!}{
	\begin{tabular}{ll|rrrr}
		\hline \hline
		&        & $K=1$       & $K=2$        & $K=3$        & $K=4$        \\ \hline
		\multirow{2}{*}{Botswana} & FLOPs  & 25.87M & 828.22M & 1218.81M & 1442.13M \\
		& Params & 1.57M  & 4.56M   & 6.09M    & 7.83M    \\ \hline
		\multirow{2}{*}{KSC}      & FLOPs  & 25.97M  & 828.32M  & 1218.90M & 1442.23M \\
		& Params & 1.58M   & 4.56M    & 6.09M    & 7.83M    \\ \hline
		\multirow{2}{*}{PaviaU}   & FLOPs  & 25.74M  & 828.09M  & 1218.67M & 1442.00M \\
		& Params & 1.57M   & 4.55M    & 6.08M    & 7.82M    \\ \hline
		\multirow{2}{*}{Salinas}  & FLOPs  & 26.06M  & 828.41M  & 1218.99M & 1442.32M \\
		& Params & 1.58M   & 4.56M    & 6.10M    & 7.83M    \\ \hline \hline
	\end{tabular}}
\end{table}

\subsubsection{Spectral Variability}

Hyperspectral imagery collected from airborne or satellite sources inevitably suffers from spectral variability, making it difficult for spectral classification \cite{ref26}. Spectral variability refers to a variation of a spectral signature for a given material, due to illumination conditions and topography, atmospheric effects, or even the intrinsic variability of the material. We conducted a comparative analysis to evaluate the performance of different models when faced with spectral variability on the Botswana dataset. According to \cite{ref26}, a 25dB white Gaussian noise was added to the dataset and subsequently re-evaluated the performance of different models using various partitioning methods. And the detailed information regarding the experimental outcomes can be found in Table \ref{tab:ablation2}.

\begin{table*}[htbp]
	\centering
	\caption{Comparative performance analysis of models under spectral variability on the Botswana dataset.}
	\label{tab:ablation2}
		\resizebox{.8\textwidth}{!}{
	\begin{tabular}{cc|cccccccccc}
		\hline
		&       & 1DCNN  & 2DCNN  & 3DCNN  & MS3DCNN & RNN    & ViT    & SpectralFormer & MorphFormer & SSFTTnet & SaaFormer                  \\ \hline
		\multirow{3}{*}{Random}     & OA    & 0.7547 & 0.9588 & 0.8931 & 0.6056  & 0.5629 & 0.7926 & 0.8837         & 0.9543      & 0.9064   & \pmb{ 0.9802 } \\
		& AA    & 0.7622 & 0.9584 & 0.8947 & 0.5805  & 0.5425 & 0.7893 & 0.8893         & 0.9449      & 0.8889   & \pmb{ 0.9794 } \\
		& Kappa & 0.7341 & 0.9554 & 0.8841 & 0.5718  & 0.5258 & 0.7752 & 0.8740         & 0.9505      & 0.8984   & \pmb{ 0.9786 } \\ \hline
		\multirow{3}{*}{Block}      & OA    & 0.5584 & 0.5905 & 0.7129 & 0.6719  & 0.5445 & 0.6890 & 0.6833         & 0.7476      & 0.7451   & \pmb{ 0.8972 } \\
		& AA    & 0.6898 & 0.5922 & 0.7071 & 0.6686  & 0.6685 & 0.7436 & 0.6776         & 0.7281      & 0.7090   & \pmb{ 0.9054 } \\
		& Kappa & 0.5276 & 0.5558 & 0.6890 & 0.6447  & 0.5122 & 0.6630 & 0.6580         & 0.7265      & 0.7213   & \pmb{ 0.8871 } \\ \hline
		\multirow{3}{*}{Checkerboard} & OA    & 0.6125 & 0.9453 & 0.8187 & 0.8079  & 0.6381 & 0.8031 & 0.9074         & 0.9406      & 0.8382   & \pmb{ 0.9731 } \\
		& AA    & 0.7079 & 0.9544 & 0.8778 & 0.8698  & 0.7355 & 0.8686 & 0.9325         & 0.9500      & 0.8817   & \pmb{ 0.9788 } \\
		& Kappa & 0.5811 & 0.9399 & 0.8029 & 0.7914  & 0.6090 & 0.7867 & 0.8984         & 0.9347      & 0.8233   & \pmb{ 0.9704 } \\ \hline
		\multirow{3}{*}{K-means}     & OA    & 0.7120 & 0.8901 & 0.8863 & 0.8545  & 0.6479 & 0.7906 & 0.8853         & 0.9184      & 0.5816   & \pmb{ 0.9436 } \\
		& AA    & 0.7164 & 0.8424 & 0.8438 & 0.8407  & 0.6518 & 0.8347 & 0.8163         & 0.8694      & 0.6740   & \pmb{0.9331} \\
		& Kappa & 0.6839 & 0.8781 & 0.8740 & 0.8404  & 0.6115 & 0.7698 & 0.8728         & 0.9098      & 0.5486   & \pmb{0.9378} \\ \hline
	\end{tabular}
	}
\end{table*}
\begin{figure*}[htbp]
	\centering
	\includegraphics[width=.8\textwidth]{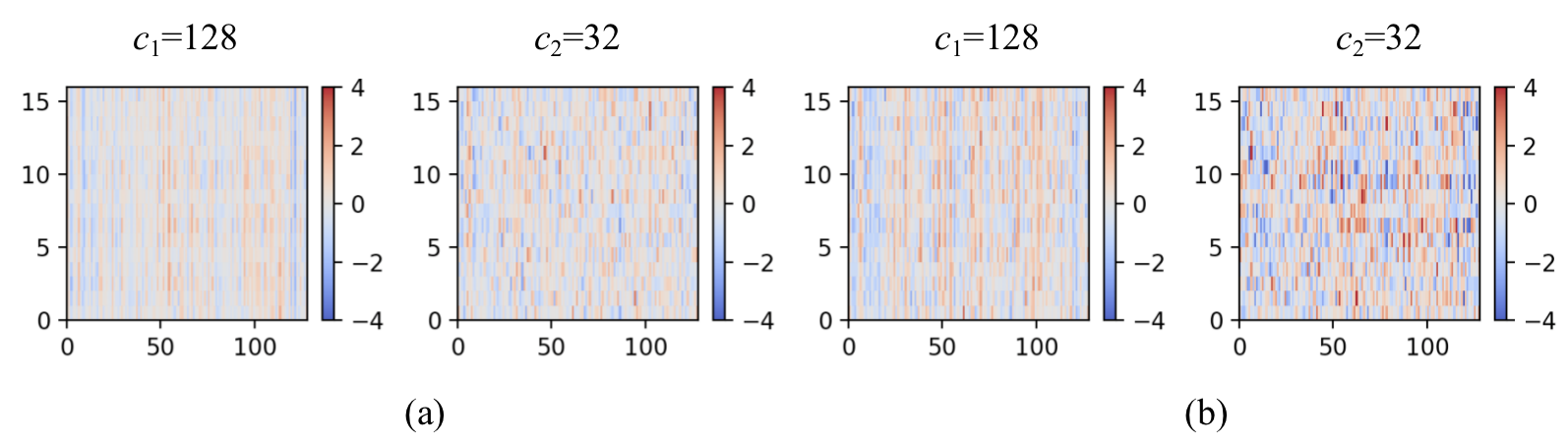}
	\caption{Visualization of selected spectral encoder output feature maps obtained with different length of spectral channels ($K=2, c_1=128, c_2=32$). We randomly selected a sample from the test set and set the number of axial aggregation attention layers in our model to $n=2$ and visualized the feature maps $\mathbf{x}\in \mathbb{R}^{hw\times c_i}, (i=1,2)$ obtained from the two layers, where the color intensity represents the relative magnitude of the spectral values. (a) shows the visualization of the spectral feature maps with spectral lengths of $c_1=128$ and $c_2=32$ from the first layer feature outputs, while (b) displays the visualization of the second layer feature outputs.}
	\label{Fig:spectral}
\end{figure*}

\begin{table*}[htbp!]
	\centering
	\caption{Number of training samples allocated into each class for Indiana Pines with number of data samples in each class.}
	\label{tab:ip_class}
	\resizebox{.9\textwidth}{!}{
	\begin{tabular}{c|ccccccccccccccccc}
		\hline \hline
		Class                       & 1  & 2    & 3   & 4   & 5   & 6   & 7  & 8   & 9  & 10  & 11   & 12  & 13  & 14   & 15  & 16 & Total \\ \hline
		number of training sample   & 6  & 6    & 6   & 6   & 7   & 7   & 6  & 6   & 6  & 6   & 7    & 7   & 6   & 7    & 7   & 6  & 102   \\
		total number of each class & 46 & 1428 & 830 & 237 & 483 & 730 & 28 & 478 & 20 & 972 & 2455 & 593 & 205 & 1265 & 386 & 93 & 10249 \\ \hline \hline
	\end{tabular}}
\end{table*}
\begin{table*}[htbp!]
	\centering
	\caption{Number of training samples allocated into each class for Salinas with number of data samples in each class.}
	\label{tab:salinas_class}
		\resizebox{\textwidth}{!}{
		\begin{tabular}{c|ccccccccccccccccc}
			\hline \hline
			Class                      & 1    & 2    & 3    & 4    & 5    & 6    & 7    & 8     & 9    & 10   & 11   & 12   & 13  & 14   & 15   & 16   & Total \\ \hline
			number of training sample  & 3    & 3    & 3    & 3    & 4    & 4    & 3    & 4     & 4    & 3    & 3    & 3    & 3   & 3    & 4    & 4    & 54    \\
			total number of each class & 2009 & 3726 & 1976 & 1394 & 2678 & 3959 & 3579 & 11271 & 6203 & 3278 & 1068 & 1927 & 916 & 1070 & 7268 & 1807 & 54129 \\ \hline \hline
		\end{tabular}}
\end{table*}
\begin{table*}[htbp!]
	\centering
	\caption{Number of training samples allocated into each class for PaviaU with number of data samples in each class.}
	\label{tab:paviau_class}
			\resizebox{.7\textwidth}{!}{
		\begin{tabular}{c|cccccccccc}
			\hline \hline
			Class                      & 1    & 2     & 3    & 4    & 5    & 6    & 7    & 8    & 9   & Total \\ \hline
			number of training sample  & 4    & 6     & 5    & 7    & 4    & 4    & 4    & 5    & 4   & 43    \\
			total number of each class & 6631 & 18649 & 2099 & 3064 & 1345 & 5029 & 1330 & 3682 & 947 & 42776 \\ \hline \hline
		\end{tabular}}
\end{table*}

Based on the findings presented in Table \ref{tab:ablation2}, it is evident that the introduction of noise interference leads to a varying degree of degradation in the classification performance of each model. Nonetheless, our proposed model consistently attains the highest level of accuracy.

\begin{figure}[htbp]
	\centering
	\begin{subfigure}{0.45\textwidth}
		\centering
		{\includegraphics[width=\textwidth]{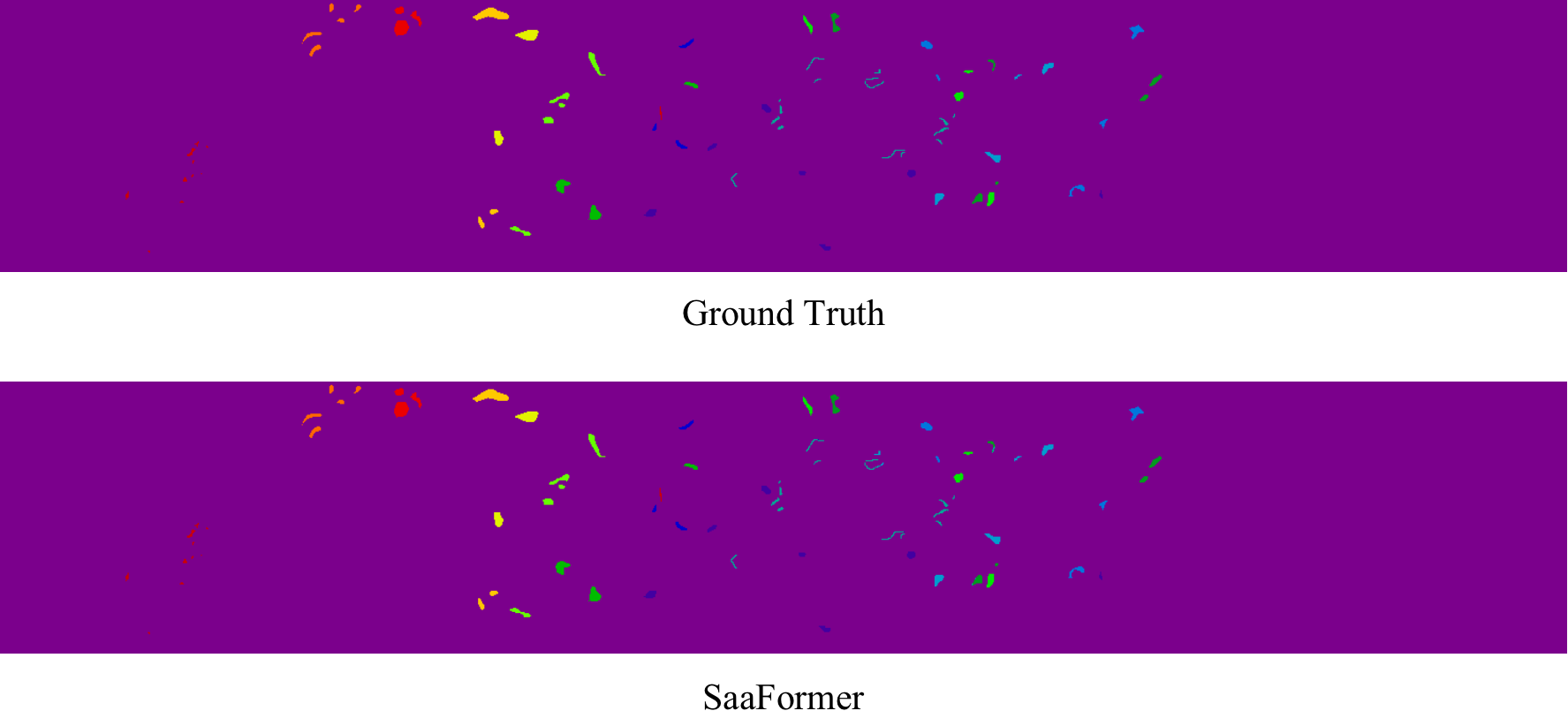}}
	\end{subfigure}\\ \vspace{2 mm} 
	\begin{subfigure}{0.3\textwidth}
		\includegraphics[width=\textwidth]{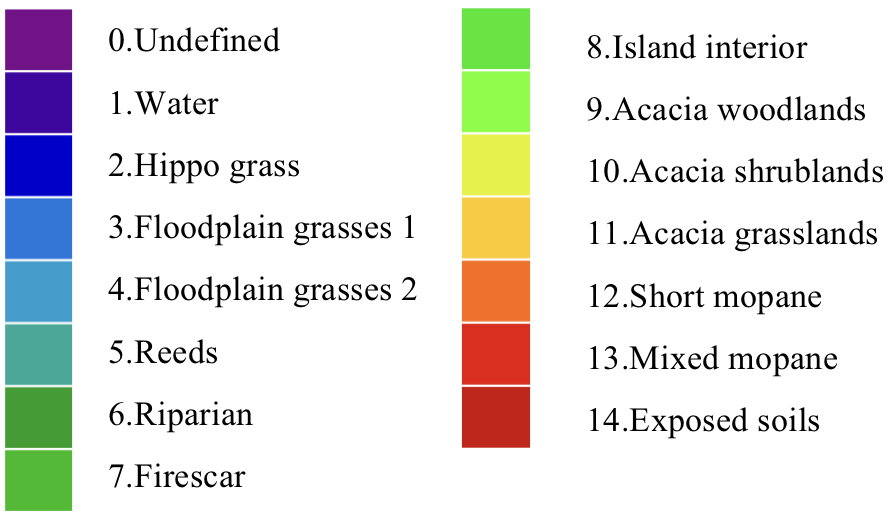}
	\end{subfigure}
	\caption{Classification maps of SaaFormer for the Botswana dataset using 5\% of training samples.}
	\label{fig:bot}
\end{figure}

\begin{figure}[htbp]
	\centering
	\begin{subfigure}{0.45\textwidth}
		\centering
		{\includegraphics[width=\textwidth]{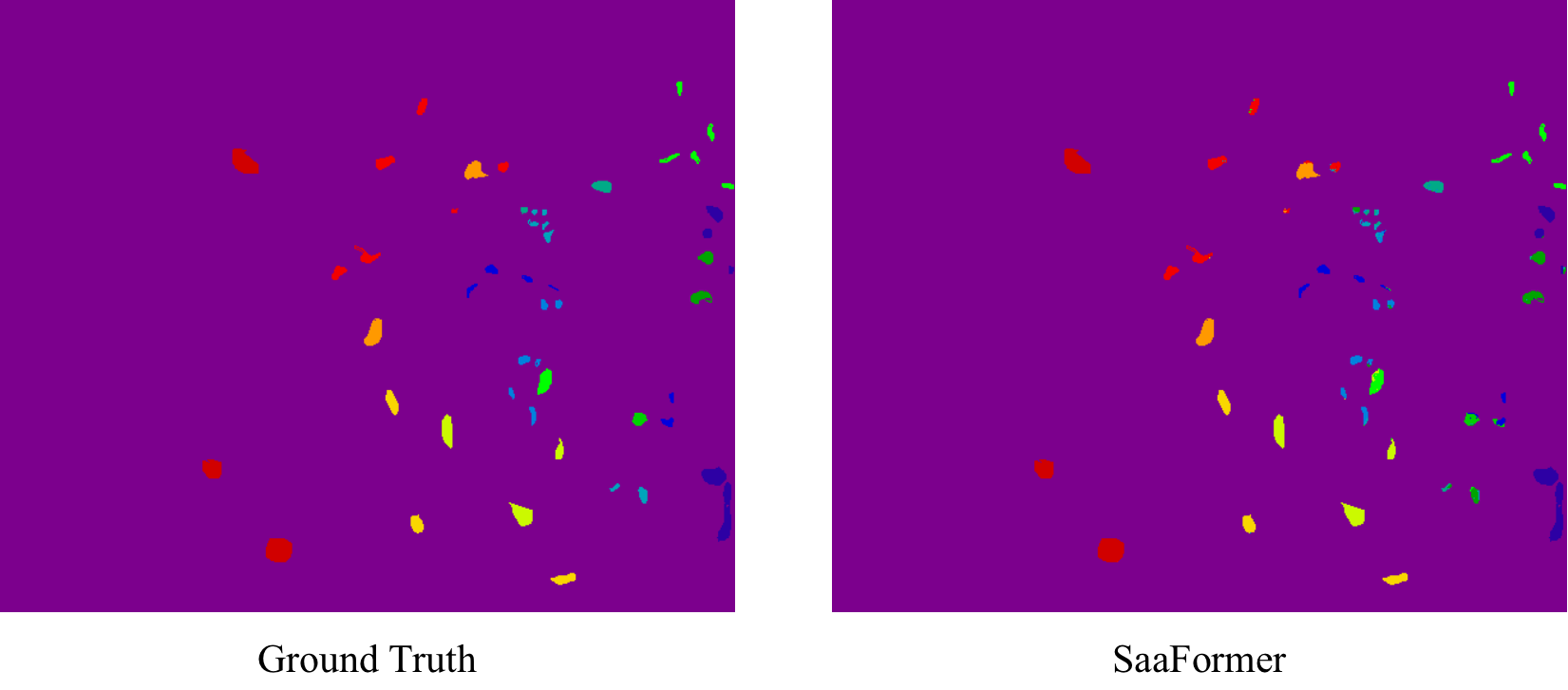}}
	\end{subfigure}\\ \vspace{2 mm} 
    \begin{subfigure}{0.3\textwidth}
    	\centering
    	\includegraphics[width=\textwidth]{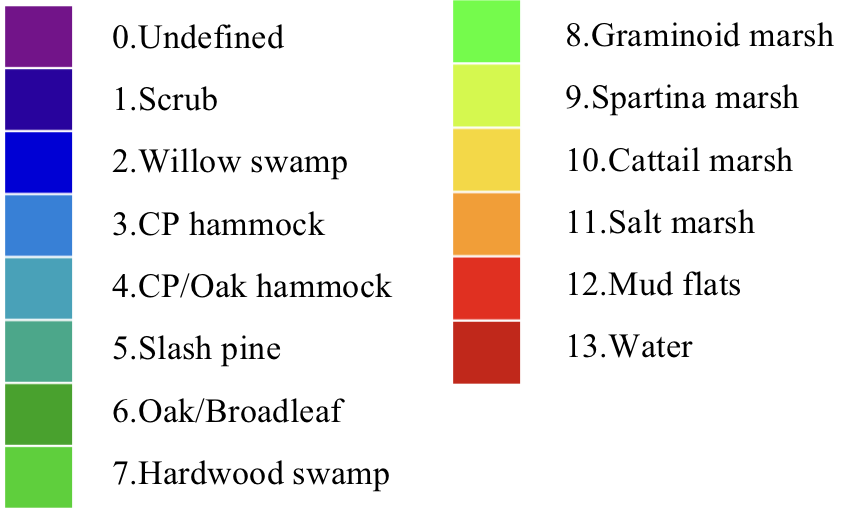}
    \end{subfigure}
	\caption{Classification maps of SaaFormer for the KSC dataset using 5\% of training samples.}
	\label{fig:ksc}
\end{figure}

\begin{figure}[htbp]
	\centering
	\begin{subfigure}{0.40\textwidth}
		\centering
		{\includegraphics[width=\textwidth]{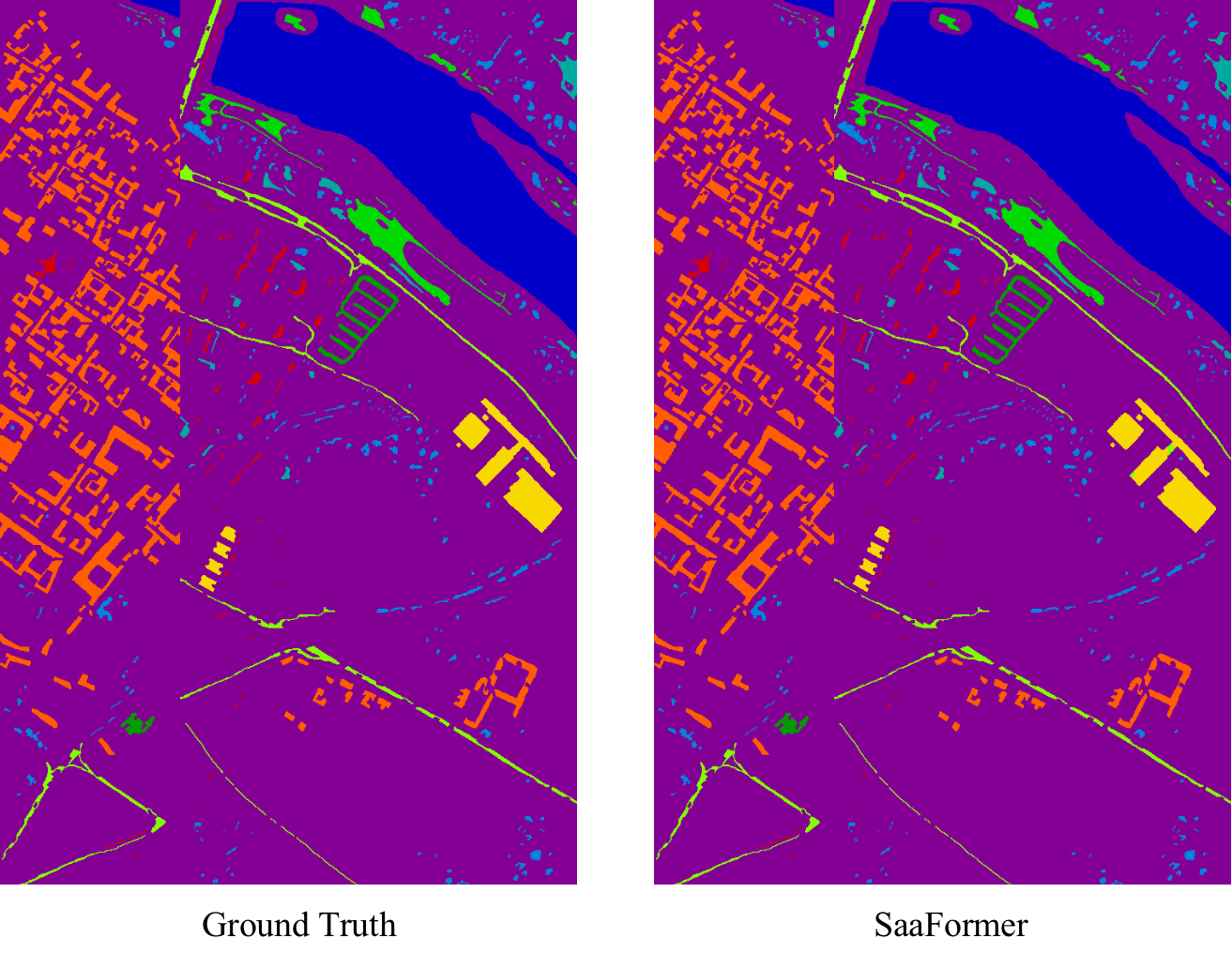}}
	\end{subfigure}\\ \vspace{1 mm} 
    \begin{subfigure}{0.25\textwidth}
    	\includegraphics[width=\textwidth]{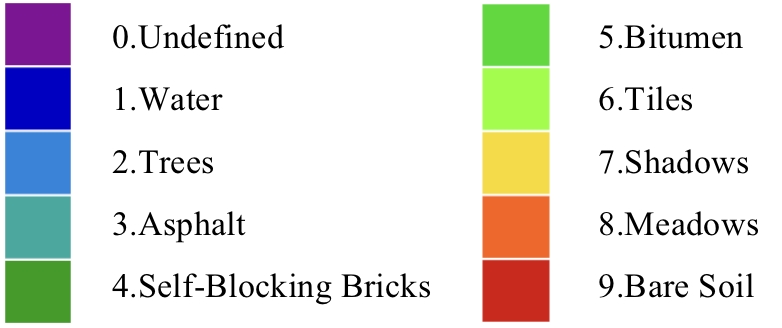}
    \end{subfigure}
	\caption{Classification maps of SaaFormer for the PaviaC dataset using 5\% of training samples.}
	\label{fig:pc}
\end{figure}

\begin{table}[htbp]
	\centering
	\caption{$P_{PR}, P_{OA}$ AND $P_{AA}$ of IRTS-3D-CNN and Saaformer by using 1\% training samples and $N_{bk}=0$ for Indian Pines dataset with the highest accuracy boldfaced.}
	\label{tab:ip_pr}
		\resizebox{.42\textwidth}{!}{
			\begin{tabular}{l|rr|rr}
				\hline \hline
				& \multicolumn{2}{c|}{IRTS-3D-CNN}    & \multicolumn{2}{c}{SaaFormer}      \\ \hline
				Class              & $P_A(C_m)$        & $P_{PR}(C_m)$   & $P_A(C_m)$       & $P_{PR}(C_m)$   \\ \hline
				1(alfalfa)         & 99.13             & $\bold{34.64}$           & $\bold{100}$     & 30.08  \\
				2(Corn\_N)         & 78.60              & $\bold{60.34}$           & $\bold{89.31}$   & 54.60  \\
				3(Corn\_M)         & $\bold{91.64}$    & 59.85           & 76.82            & $\bold{69.27}$  \\
				4(Corn)            & 97.81    & $\bold{56.55}$           & $\bold{98.70}$            & 53.60  \\
				5(Grass\_P)        & 92.34    & $\bold{35.26}$           & $\bold{95.59}$            & 23.62    \\
				6(Grass\_T)        & 98.67    & 34.50            & $\bold{99.86}$            & $\bold{46.60}$  \\
				7(Grass\_M)        & 100               & $\bold{26.13}$           & $\bold{100}$     & 7.86  \\
				8(Hay\_W)          & 99.64             & $\bold{71.88}$           & $\bold{100}$   & 67.24    \\
				9(Oats)            & $\bold{100}$               & 21.43           & $\bold{100}$              & $\bold{25.00}$  \\
				10(Soybean\_N)     & $\bold{87.70}$     & 56.40            & 83.64            & $\bold{57.73}$  \\
				11(Soybean\_M)     & $\bold{77.84}$             & 76.95           & 77.28   & $\bold{84.85}$  \\
				12(Soybean\_C)     & 90.35    & 49.53           & $\bold{96.08}$            & $\bold{53.45}$  \\
				13(Wheats)         & $\bold{100}$               & 65.13           & $\bold{100}$              & $\bold{65.25}$  \\
				14(Woods)          & 97.91    & $\bold{26.76}$           & $\bold{98.41}$            & 26.38  \\
				15(Building)       & 98.08             & 16.37           & $\bold{100}$   & $\bold{20.07}$  \\
				16(Tower)          & 99.79             & $\bold{42.23}$           & $\bold{100}$     & 21.75  \\ \hline
				$P_{OA}$(\%)       & \multicolumn{2}{c|}{88.45}          & \multicolumn{2}{c}{$\bold{88.91}$} \\
				$P_{AA}$(\%)       & \multicolumn{2}{c|}{94.34} & \multicolumn{2}{c}{$\bold{94.73}$}          \\
				$P_{OPR}(BKG)$(\%) & \multicolumn{2}{c|}{43.12}          & \multicolumn{2}{c}{$\bold{45.21}$} \\
				$P_{APR}(BKG)$(\%) & \multicolumn{2}{c|}{$\bold{45.87}$}          & \multicolumn{2}{c}{44.21} \\
				Time(s) & \multicolumn{2}{c|}{2448.4}          & \multicolumn{2}{c}{285.2} \\ \hline \hline
		\end{tabular}
}
\end{table}

\begin{table}[htbp]
	\centering
	\caption{$P_{PR}$ of IRTS-3D-CNN and Saaformer using different $N_{bk}$ for Indian Pines dataset.}
	\label{tab:ip_prbg}
	\resizebox{.5\textwidth}{!}{
		\begin{tabular}{l|c|ccccc}
			\hline \hline
			& \multicolumn{1}{r|}{IRTS-3D-CNN}   & \multicolumn{5}{c}{SaaFormer}                                                                                                                                                     \\
			& \multicolumn{1}{l|}{}              & $N_{bk}=7$                        & $N_{bk}=27$                       & $N_{bk}=67$                       & \multicolumn{1}{l}{$N_{bk}=107$}  & \multicolumn{1}{l}{$N_{bk}=207$}  \\ \hline
			Class              & \multicolumn{1}{r|}{$P_{PR}(C_m)$} & \multicolumn{1}{r}{$P_{PR}(C_m)$} & \multicolumn{1}{r}{$P_{PR}(C_m)$} & \multicolumn{1}{r}{$P_{PR}(C_m)$} & \multicolumn{1}{r}{$P_{PR}(C_m)$} & \multicolumn{1}{r}{$P_{PR}(C_m)$} \\ \hline
			1(alfalfa)         & 34.64                              & 43.33                             & 56.34                             & 40.40                             & 79.07                             & 61.02                             \\
			2(Corn\_N)         & 60.34                              & 61.16                             & 74.59                             & 66.13                             & 73.82                             & 79.02                             \\
			3(Corn\_M)         & 59.85                              & 59.67                             & 82.40                             & 70.89                             & 68.23                             & 84.80                             \\
			4(Corn)            & 56.55                              & 54.85                             & 70.37                             & 72.52                             & 68.62                             & 76.87                             \\
			5(Grass\_P)        & 35.26                              & 57.41                             & 75.93                             & 80.91                             & 78.07                             & 90.89                             \\
			6(Grass\_T)        & 34.50                              & 50.96                             & 88.91                             & 71.10                             & 77.69                             & 96.06                             \\
			7(Grass\_M)        & 26.13                              & 18.03                             & 48.89                             & 33.85                             & 72.00                             & 52.78                             \\
			8(Hay\_W)          & 71.88                              & 92.62                             & 94.80                             & 83.54                             & 95.00                             & 100                               \\
			9(Oats)            & 21.43                              & 51.85                             & 34.15                             & 38.89                             & 33.33                             & 50.00                             \\
			10(Soybean\_N)     & 56.40                              & 59.47                             & 83.84                             & 79.02                             & 81.58                             & 89.15                             \\
			11(Soybean\_M)     & 76.95                              & 71.61                             & 84.19                             & 79.82                             & 86.16                             & 86.51                             \\
			12(Soybean\_C)     & 49.53                              & 61.99                             & 75.22                             & 68.05                             & 61.68                             & 65.75                             \\
			13(Wheats)         & 65.13                              & 74.53                             & 74.53                             & 75.97                             & 93.69                             & 88.08                             \\
			14(Woods)          & 26.76                              & 75.26                             & 91.73                             & 61.01                             & 87.02                             & 98.38                             \\
			15(Building)       & 16.37                              & 24.30                             & 74.38                             & 62.92                             & 68.01                             & 89.36                             \\
			16(Tower)          & 42.23                              & 50.60                             & 52.10                             & 85.57                             & 75.24                             & 67.74                             \\
			0(Background)      &                                    & 89.51                             & 92.18                             & 87.47                             & 86.59                             & 86.76                             \\ \hline
			$P_{OPR}(BKG)$(\%) & 43.12                              & 70.38                             & 84.42                             & 77.96                             & 82.84                             & 86.22                             \\
			$P_{APR}(BKG)$(\%) & 45.87                              & 58.35                             & 73.23                             & 66.94                             & 74.83                             & 80.19                             \\
			Time(s)            & 2448.4                             & 287.3                            & 291.6                            & 286.5                            & 288.4                            & 293.7                            \\ \hline \hline
		\end{tabular}
}
\end{table}

\subsection{Visualization}
\subsubsection{Classification Result Visualization}
We perform a qualitative assessment by visually representing the classification maps produced by various methods. Fig. \ref{fig:salinas} is the results on Salinas dataset while, respectively, Fig. \ref{fig:paviau_c} present the results on PaviaU dataset using different dataset division methods.

By visualizing the datasets, it is evident that our model produces highly satisfactory classification maps, particularly with regard to texture and edge details. Within  the region of interests (ROIs), our model displays notably more realistic and intricate details. Consequently, the visual representation of the classification results unequivocally showcases that our model excels in generating highly favorable classification maps, especially in terms of texture and edge details. Futhermore, In order to provide a more comprehensive visual inspection of test methods, Figs. \ref{fig:bot}-\ref{fig:pc} show their classification maps for the other three datasets i.e. Botswanan, KSC and PaviaC datasets using 5\% of training samples.


\subsubsection{Feature Extraction Visualization}
Furthermore, when employing block-wise sampling, we randomly select a set of samples from the PaviaU dataset, both from the same class and different classes. We visualize the spectral pattern of their feature maps after passing through the Axial Transformer Block. Fig. \ref{Fig:keshihua} (a) represents the spectral pattern of samples from the same class, while Fig. \ref{Fig:keshihua} (b) represents the spectral pattern of samples from different classes. Samples of the same class exhibit highly similar spectral patterns, while those from different classes show significant differences. This further demonstrates that our model excels at distinguishing between different types of samples which yields more accurate classification performance, displaying enhanced generalization and robustness capabilities.

\begin{figure}[htbp]
	\centering
	\begin{subfigure}{0.48\textwidth}
		{\includegraphics[width=\textwidth]{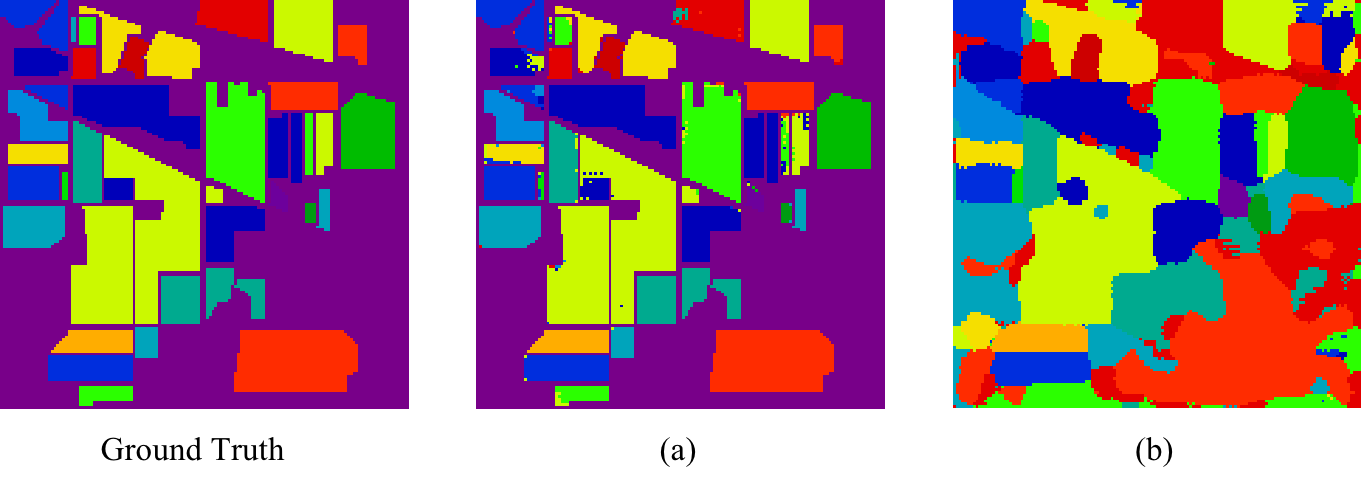}}
	\end{subfigure}\\ \vspace{2 mm} 
	\begin{subfigure}{0.3\textwidth}
		\centering
		\includegraphics[width=\textwidth]{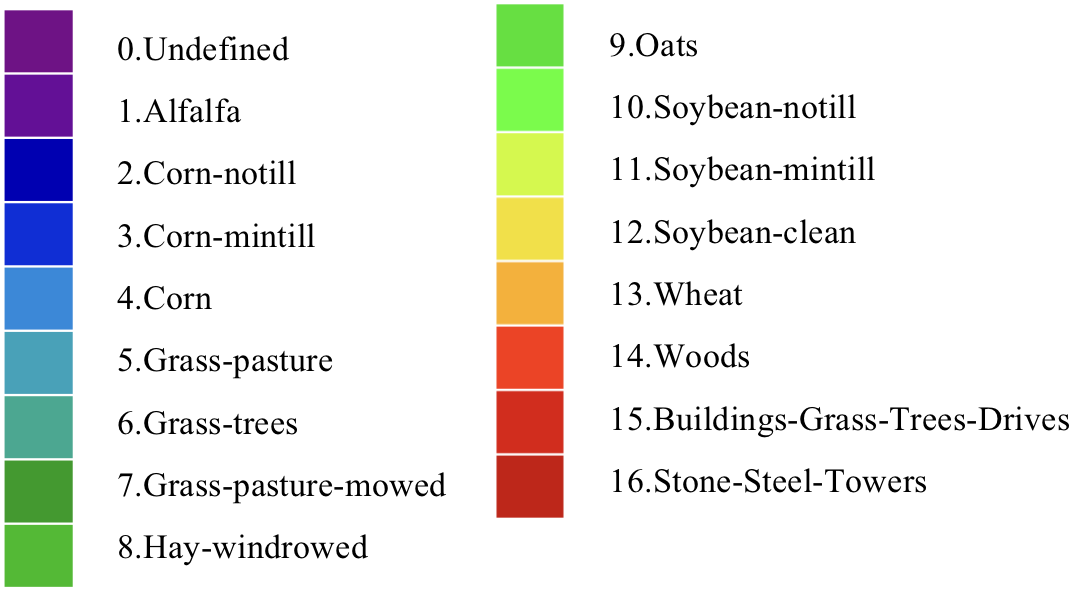}
	\end{subfigure}
	\caption{Classification maps of SaaFormer for the Indian Pines dataset using 1\% of training samples (a) without BKG (b) with BKG.}
	\label{fig:ip_bkg}
\end{figure}

\subsubsection{Spectral Feature Visualization}

\begin{table}[t]
	\centering
	\caption{{$P_A, P_{PR}, P_{OA}$AND $P_{AA}$ of IRTS-3D-CNN and SaaFormer using 0.1\% of training samples and $N_{bk}=0$ for Salinas dataset with the highest accuracy boldfaced.}}
	\label{tab:sa_pr}
		\resizebox{.42\textwidth}{!}{
			\begin{tabular}{l|rr|rr}
				\hline \hline
				& \multicolumn{2}{c|}{IRTS-3D-CNN} & \multicolumn{2}{c}{SaaFormer}      \\ \hline
				Class              & $P_A(C_m)$      & $P_{PR}(C_m)$  & $P_A(C_m)$       & $P_{PR}(C_m)$   \\ \hline
				1(Weeds\_1)        & 99.35  & 68.62          & ${99.95}$            & $\bold{71.57}$  \\
				2(Weeds\_2)        & $\bold{99.91}$           & 63.77          & 99.73     & $\bold{77.25}$  \\
				3(Fallow)          & 96.58           & $\bold{14.45}$         & $\bold{98.32}$   & 11.08  \\
				4(Fallow\_P)       & $\bold{99.89}$  & $\bold{26.02}$          & 98.92           & 15.02  \\
				5(Fallow\_S)       & 98.82  & $\bold {57.48}$          & $\bold{100}$            & 49.10  \\
				6(Stubble)         & 99.42           & $\bold{82.43}$          & $\bold{100}$     & 73.11  \\
				7(Celery)          & $\bold{99.94}$  & 78.33          & 99.18            & $\bold{82.17}$  \\
				8(Grapes)          & 69.16           & 75.18 & $\bold{80.40}$   & $\bold{91.02}$           \\
				9(Soil)            & 99.60            & 28.60           & $\bold{99.79}$   & $\bold{36.86}$  \\
				10(Corn)           & 84.49           & $\bold{64.65}$          & $\bold{93.80}$   & 56.22  \\
				11(Lettuce\_4)     & $\bold{98.55}$  & 19.41          & 98.22            & $\bold{24.05}$  \\
				12(Lettuce\_5)     & ${99.98}$           & $\bold{32.37}$          & 94.49     & 27.63  \\
				13(Lettuce\_6)     & 98.47           & 60.09 & $\bold{100}$   & $\bold{62.02}$          \\
				14(Lettuce\_7)     & 96.45           & 60.49          & $\bold{98.78}$   & $\bold{76.17}$  \\
				15(Vinyard\_U)     & $\bold{81.05}$  & 57.44 & 73.71            & $\bold{83.34}$          \\
				16(Vinyard\_T)     & 98.24           & $\bold{55.08}$          & $\bold{99.11}$   & 31.24  \\ \hline
				$P_{OA}$(\%)       & \multicolumn{2}{c|}{89.60}        & \multicolumn{2}{c}{$\bold{92.83}$} \\
				$P_{AA}$(\%)       & \multicolumn{2}{c|}{94.99}       & \multicolumn{2}{c}{$\bold{96.50}$} \\
				$P_{OPR}(BKG)$(\%) & \multicolumn{2}{c|}{43.65}       & \multicolumn{2}{c}{$\bold{45.60}$} \\
				$P_{APR}(BKG)$(\%) & \multicolumn{2}{c|}{52.77}       & \multicolumn{2}{c}{$\bold{53.40}$} \\ 
				Time(s) & \multicolumn{2}{c|}{5531.7}          & \multicolumn{2}{c}{882.8} \\
				\hline \hline
			\end{tabular}
	}
\end{table}

We make a qualitative evaluation by visualizing the spectral feature map obtained by the proposed SaaFormer framework on Botswana dataset. According to the results of the above ablation experiment, Fig. \ref{Fig:spectral} visualizes selected spectral encoder output feature maps with $K=2$. We randomly selected a sample from the test set and set the number of axial aggregation attention layers in our model to $n=2$. We visualized the feature maps obtained from the last two layers. Specifically, for a spectral feature map $\mathbf{x}\in \mathbb{R}^{h\times w\times c_i}$, we reshaped it to $\mathbf{x}\in \mathbb{R}^{hw\times c_i}$,(i=1,2), where the color intensity represents the relative magnitude of the spectral values. The results demonstrate that a longer spectral clip ($c_1=128$) leads to lesser variation in the amplitude of the spectral feature map, highlighting a greater emphasis on extracting overall features. Conversely, a shorter spectral clip ($c_2=32$) results in deeper colors and larger amplitudes, signifying a heightened focus on extracting detailed features. This also demonstrates the effectiveness and superiority of the designed multi-level spectral extraction module from the visual perspective.

\begin{figure}[htbp]
	\centering
	\begin{subfigure}{0.5\textwidth}
		{\includegraphics[width=\textwidth]{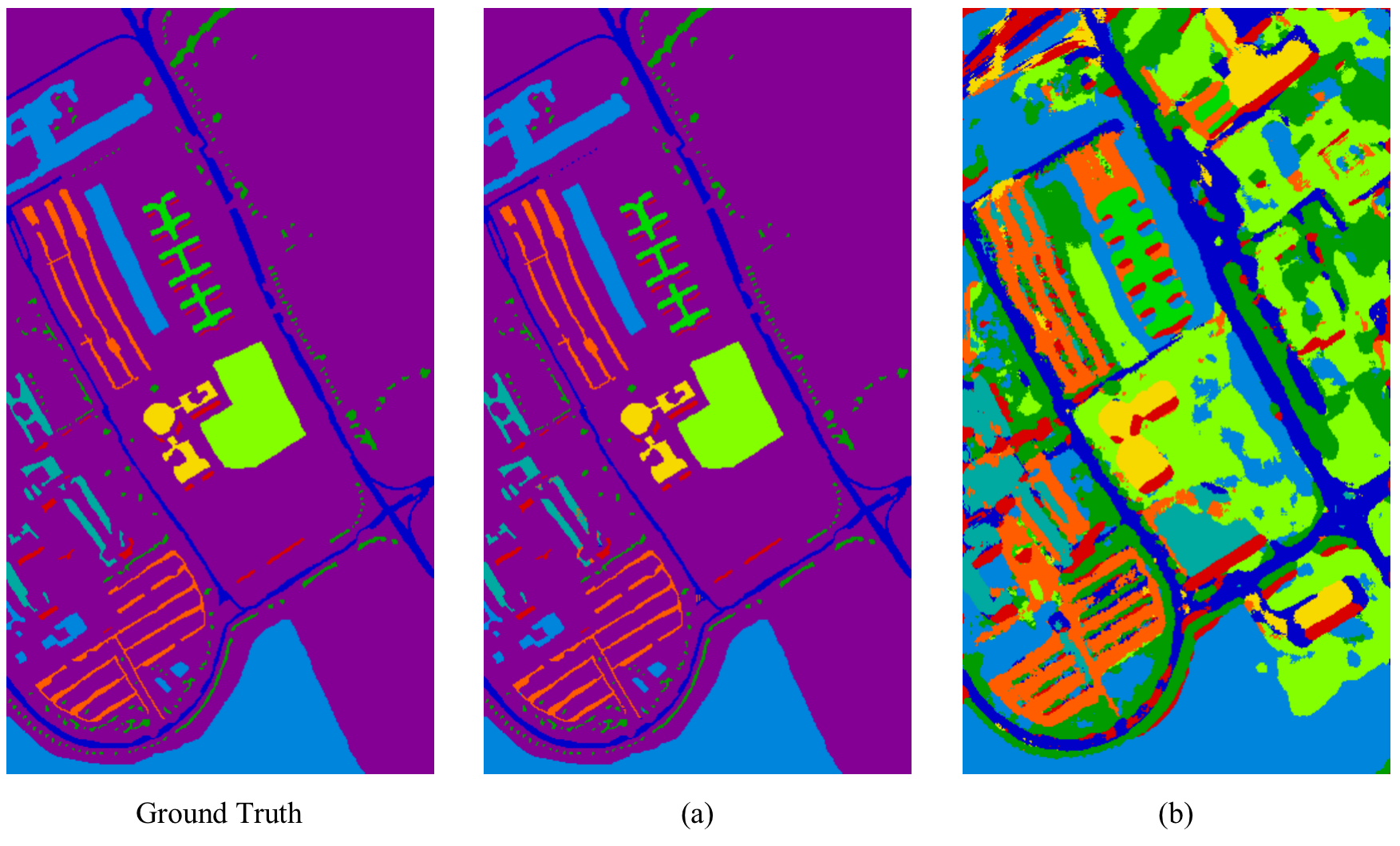}}
	\end{subfigure}\\ \vspace{2 mm} 
	\begin{subfigure}{0.23\textwidth}
		\centering
		\includegraphics[width=\textwidth]{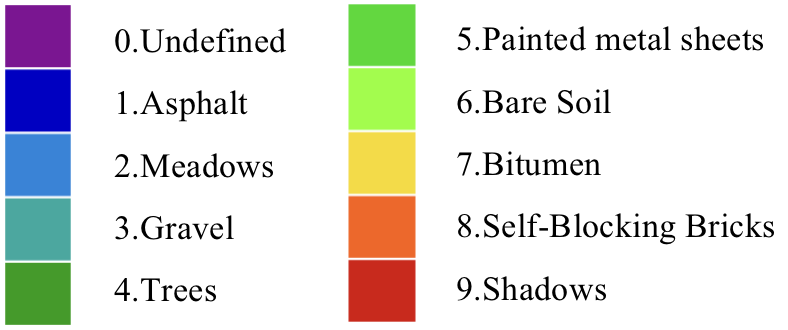}
	\end{subfigure}
	
	\caption{Classification maps of SaaFormer for the PaviaU dataset using 0.1\% of training samples (a) without BKG (b) with BKG.}
	\label{fig:pu_bkg}
\end{figure}

\begin{figure}[htbp]
	\centering
	\begin{subfigure}{0.48\textwidth}
		{\includegraphics[width=\textwidth]{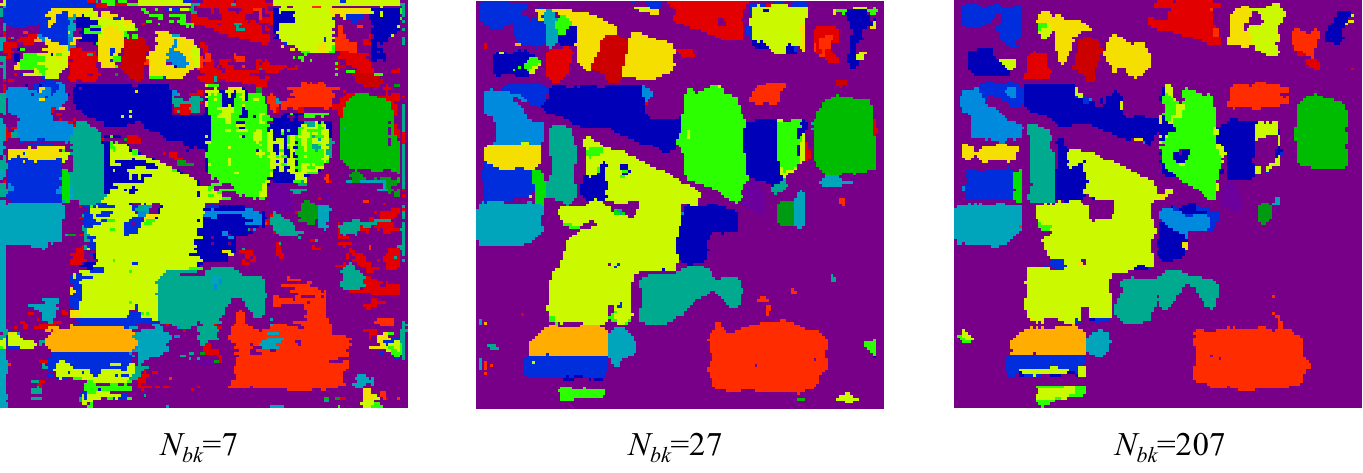}}
	\end{subfigure}\\ \vspace{2 mm} 
	\caption{Classification maps of SaaFormer for the Indian Pines dataset using different numbers $N_{bk}$ of background samples and 1\% training samples.}
	\label{fig:ip_bkg11}
\end{figure}

\subsection{Background Classification}

As \cite{ref24, ref28} mentioned, HSIC analysis has been conducted based on the provided ground truth (GT), with the exclusion of background (BKG) samples from the classification process. A data sample is classified as a background (BKG) sample simply since it cannot be labeled. Two main reasons lead to the presence of BKG samples. Firstly, the complexity of the data being analyzed may make it prohibitively expensive to assign complete ground truth labels. Secondly, hyperspectral imaging sensors have the capability to detect numerous subtle material substances that may not be identifiable through existing knowledge or visual inspection methods. Following \cite{ref24}, we choose $P_{PR}$, $ P_{OPR}$ and $P_{APR}$ as metrics of BKG.

\begin{align}
	p_{PR}(C_m)=\frac{\hat{n}_{mm}}{\hat{n}_m}
\end{align}
\begin{align}\label{eq:opr}
	P_{OPR}=\sum_{m=1}^{M}p(\hat{C}_m)p_{PR}(\hat C_m)=\sum_{m=1}^{M}\frac{\hat{n}_m}{\hat{N}}p_{PR}(C_m)
\end{align}
\begin{align}\label{eq:apr}
	P_{APR}=\frac{1}{M}\sum_{m=1}^M p_{PR}(\hat C_m)
\end{align}

We conducted experimental comparisons on three datasets \cite{ref24}. Instead of being treated as a separate class, background samples are considered to potentially belong to a known class that cannot be labeled. During training, only samples with ground truth labels are used, while during testing, the background samples are included.  For the Indiana Pines data, the number of  training samples is 102 (SR=1\%) with the number of selected training sample numbers for each class tabulated in Table \ref{tab:ip_class}. Unlike the Indian Pines data, Salinas data has a large number of data samples. We selected small training sizes with 0.1\% of total data samples as training samples. This ensured that the training sample sizes for each class were roughly equal, ranging from 3 to 4, respectively, according to the training class sizes allocated in Table \ref{tab:salinas_class}. For the University of Pavia data, training sample size for each class is roughly set ranging from 4 to 7 as listed in Table \ref{tab:paviau_class}  with total number of 428 training samples with approximate SR=0.1\% which was the same as that used for Salinas. The above SR selections follow \cite{ref24} exactly and train in an iterative way. Then we evaluate the performance metrics of our model in comparison to IRTS-3D-CNN \cite{ref24}.


\begin{figure}[t]
	\centering
	\begin{subfigure}{0.4\textwidth}
		\centering
		{\includegraphics[width=\textwidth]{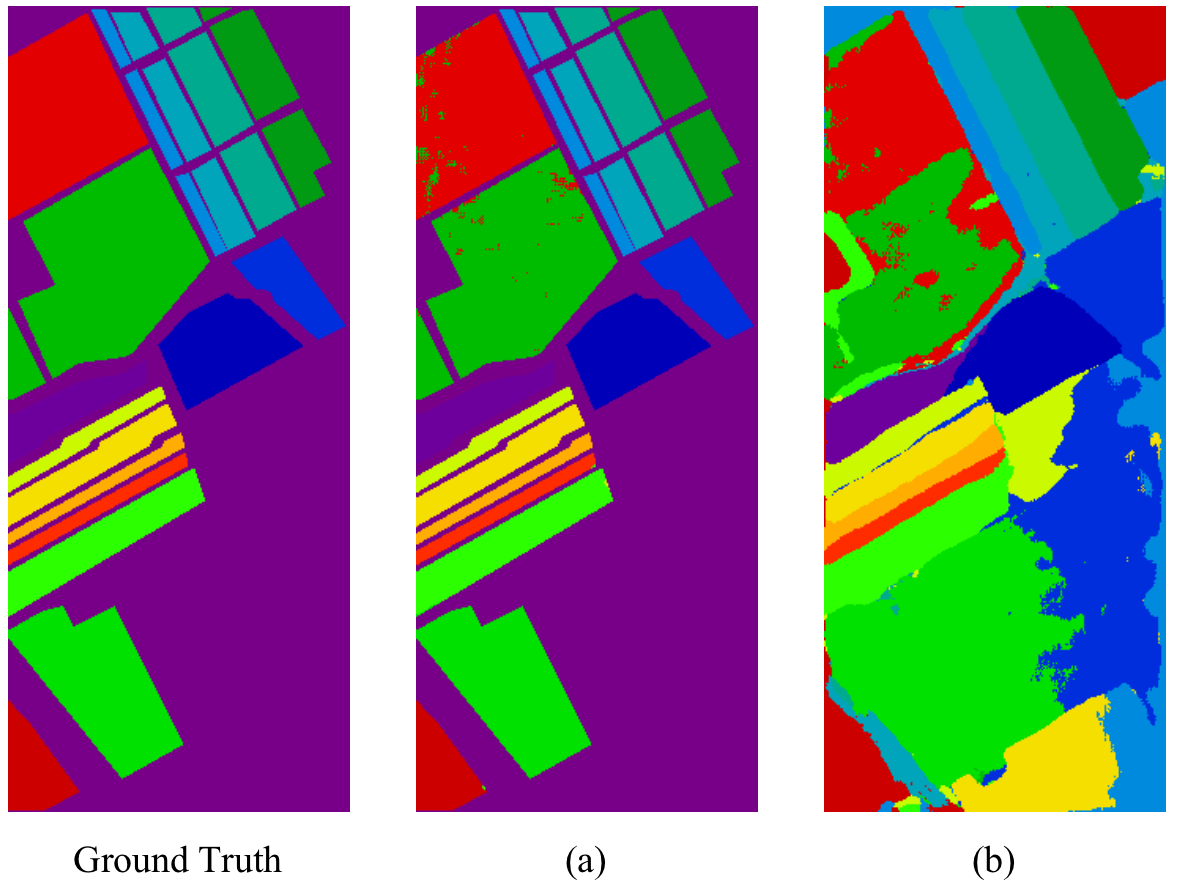}}
	\end{subfigure}\\ \vspace{2 mm} 
	\begin{subfigure}{0.3\textwidth}
		\includegraphics[width=\textwidth]{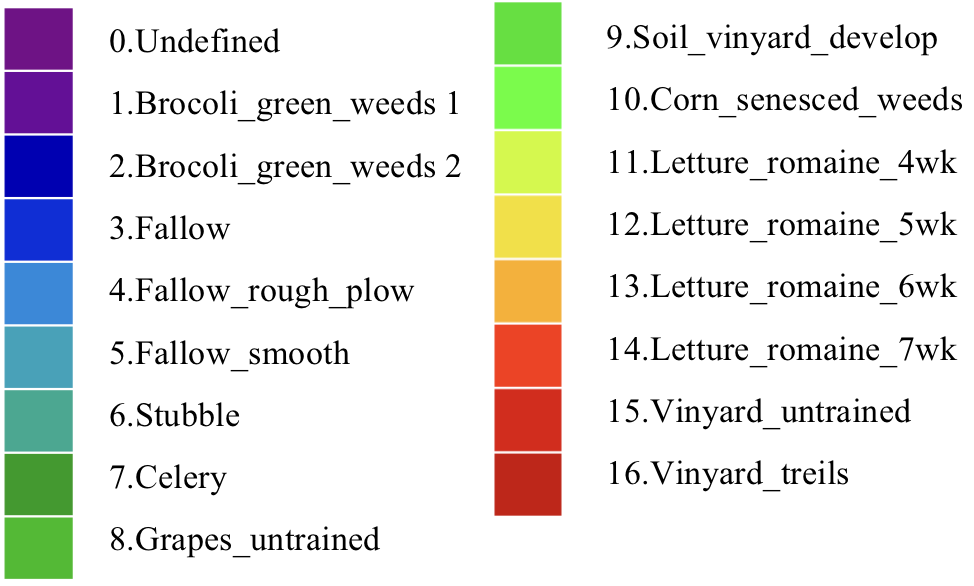}
	\end{subfigure}
	\caption{Classification maps of SaaFormer for the Salinas dataset using 0.1\% of training samples (a) without BKG (b) with BKG.}
	\label{fig:sa_bkg}
\end{figure}

\begin{table}[t]
	\centering
	\caption{$P_A, P_{PR}, P_{OA}$AND $P_{AA}$ of IRTS-3D-CNN and SaaFormer using 0.1\% of training samples and $N_{bk}=0$ for PaviaU dataset with the highest accuracy boldfaced.}
	\label{tab:paviau_pr}
	\resizebox{.48\textwidth}{!}{
		\begin{tabular}{l|rr|rr}
			\hline \hline
			& \multicolumn{2}{c|}{IRTS-3D-CNN} & \multicolumn{2}{c}{SaaFormer}      \\ \hline
			Class                   & $P_A(C_m)$      & $P_{PR}(C_m)$  & $P_A(C_m)$       & $P_{PR}(C_m)$   \\ \hline
			1(Asphalt)              & $\bold{86.77}$  & $\bold{23.44}$          & 77.38            & 23.37  \\
			2(Meadows)              & 84.70            & 32.75          & $\bold{88.42}$   & $\bold{43.19}$  \\
			3(Gravel)               & $\bold{93.67}$  & 15.57          & 92.31            & $\bold{16.27}$  \\
			4(Tree)                 & $\bold{96.50}$   & $\bold{8.87}$           & 96.34            & 6.56  \\
			5(Painted metal)        & 99.03           & 34.16          & $\bold{100}$     & $\bold{44.22}$  \\
			6(Bare Soil)            & 71.17           & 10.04          & $\bold{91.60}$   & $\bold{16.34}$  \\
			7(Bitumen)              & $\bold{97.43}$  & $\bold{24.02}$          & 97.36            & 17.73  \\
			8(Self-Blocking Bricks) & $\bold{95.38}$  & $\bold{15.41}$          & 86.65            & 8.25  \\
			9(Shadows)              & 99.84           & 8.17           & $\bold{100}$     & $\bold{12.38}$  \\ \hline
			$P_{OA}$(\%)            & \multicolumn{2}{c|}{86.82}       & \multicolumn{2}{c}{$\bold{88.58}$} \\
			$P_{AA}$(\%)            & \multicolumn{2}{c|}{91.61}       & \multicolumn{2}{c}{$\bold{92.23}$} \\
			$P_{OPR}(BKG)$(\%)      & \multicolumn{2}{c|}{17.91}       & \multicolumn{2}{c}{$\bold{18.99}$} \\
			$P_{APR}(BKG)$(\%)      & \multicolumn{2}{c|}{19.16}       & \multicolumn{2}{c}{$\bold{20.92}$} \\
			Time(s) & \multicolumn{2}{c|}{4234.9}          & \multicolumn{2}{c}{691.4} \\ \hline \hline
		\end{tabular}
	}
\end{table}

We conducted research on two experimental conditions concerning the Indian Pines dataset, primarily focusing on the inclusion of background samples during the training process. $N_{bk}$ represents the number of background samples in the training dataset. To comprehensively evaluate model performance, we used different numbers of background samples. Initially, we set $N_{bk}=0$, indicating the absence of any background samples in the training dataset. Subsequently, we set $N_{bk}=7$ to ensure an equal number of background samples compared to other classes in the training dataset. Additionally, we set $N_{bk}=107$ (with a total of 10,776 background samples) to maintain a sampling rate of 1\% for background samples. Furthermore, we explored cases where $N_{bk}$ assumes values between and above these specified values to further observe fluctuations in model accuracy.

The performance comparisons are summarized in Table \ref{tab:ip_pr}, \ref{tab:sa_pr} and \ref{tab:paviau_pr}. It is evident that SaaFormer demonstrates superior classification performance compared to IRTS-3D-CNN. When considering the inclusion of background (BKG) in the classification process to compute $P_{OPR}$(BKG) and $P_{APR}$(BKG), SaaFormer still outperforms RTS-3D-CNN. Furthermore, Figs. \ref{fig:ip_bkg}-\ref{fig:sa_bkg} show their classification maps without/with BKG for the three datasets, respectively.

In response to the background issue, we conducted further analysis regarding the impact of varying quantities of background samples on precision $P_{PR}$. It is discerned from the data presented in Table \ref{tab:ip_prbg} that augmenting the number of background samples $N_{bk}$ in the training set significantly influences $P_{PR}$. As the number of background samples increases, both $P_{OPR}$(BKG) and $P_{APR}$(BKG) of SaaFormer notably surpass those of IRTS-3D-CNN. Particularly noteworthy is that at a threshold of 27 background samples, SaaFormer demonstrates markedly elevated precision, with subsequent increments in background sample count yielding marginal changes in model precision. If we looked at classes 2 and 15, our model was significantly improved after it was introducted the additional background samples results in pronounced enhancements in precision compared to IRTS-3D-CNN. In order to provide a visual inspection of test methods, Figs. \ref{fig:ip_bkg11} shows their classification maps using different numbers of background samples for the Indian Pines dataset, respectively. This underscores the potency of incorporating background samples in bolstering model discriminative capacity and enhancing its generalization capabilities.

\begin{table}[htbp]
	\centering
	\caption{Number of training samples and test samples allocated into each class for four benchmark datasets}
	\label{tab:sample_size}
	\resizebox{.48\textwidth}{!}{
		\begin{tabular}{c|cc|cc|cc|cc}
			\hline \hline
			\multirow{2}{*}{Class} & \multicolumn{2}{c|}{Indian Pines} & \multicolumn{2}{c|}{Salinas} & \multicolumn{2}{c|}{PaviaU} & \multicolumn{2}{c}{Houston} \\
			& $n_{train}$    & $n_{test}$    & $n_{train}$  & $n_{test}$ & $n_{train}$ & $n_{test}$ & $n_{training}$ & $n_{test}$ \\ \hline
			1                      & 23                & 23            & 31              & 1978       & 56             & 6575       & 50             & 1201       \\
			2                      & 36                & 1392          & 31              & 3695       & 56             & 18593      & 50             & 1204       \\
			3                      & 35                & 795           & 31              & 1945       & 55             & 2044       & 50             & 647        \\
			4                      & 32                & 205           & 31              & 1363       & 56             & 3008       & 50             & 1194       \\
			5                      & 32                & 451           & 31              & 2647       & 56             & 1289       & 50             & 1192       \\
			6                      & 35                & 695           & 31              & 3928       & 55             & 4974       & 50             & 275        \\
			7                      & 14                & 14            & 31              & 3548       & 56             & 1274       & 50             & 1218       \\
			8                      & 32                & 446           & 33              & 11238      & 55             & 3627       & 50             & 1194       \\
			9                      & 10                & 10            & 32              & 6171       & 55             & 892        & 50             & 1202       \\
			10                     & 36                & 936           & 32              & 3246       &                &            & 50             & 1177       \\
			11                     & 43                & 2412          & 31              & 1037       &                &            & 50             & 1185       \\
			12                     & 34                & 559           & 31              & 1896       &                &            & 50             & 1183       \\
			13                     & 33                & 172           & 31              & 885        &                &            & 50             & 419        \\
			14                     & 40                & 1225          & 31              & 1039       &                &            & 50             & 378        \\
			15                     & 34                & 352           & 31              & 7237       &                &            & 50             & 610        \\
			16                     & 31                & 62            & 31              & 1776       &                &            &                &            \\ \hline
			BKG                    & -                 & 10776         & -               & 56975      & -              & 164624     & -              & 649816     \\
			Total                  & 500               & 9749          & 500             & 53629      & 500            & 42276      & 750            & 14279      \\ \hline \hline
		\end{tabular}
	}
\end{table}

Python and Pytorch libraries were used to run the experiments. All experimental results are generated on a desktop with 16 vCPU Intel(R) Xeon(R) Platinum 8352V CPU and NVIDIA GeForce RTX 4090. In terms of parameter setting, we used a patch size of 7 and trained the models for 1000 epoches with a batch size of 128, and the embedding is further segmented into $K=2$, $c_1=128$  and $c_2=32$ spectral clips which is then processed through two consecutive transformer encoder blocks for classification. Each encoder block includes a four-head axial aggregation attention layer, a MLP layer and a GELU nonlinear activation layer. The specific parameters of our model have been detailed in the experimental setting which employs an embedded spectrum with $d=128$ channels.And the time presented in Table \ref{tab:ip_prbg}  indicates that our model exhibits relatively short execution times, with the size of $N_{bk}$ exerting minimal influence on the runtime.

Furthermore, we compared and analyzed the experimental results with reference \cite{ref28}, which delved deeply into the background issue. For fairness, we adopt the same size of training samples for each class as \cite{ref28}, and the numbers of training samples and test samples for all image scenes are summarized in Table \ref{tab:sample_size}.

\begin{table}[htbp]
	\centering
	\caption{Averaged $P_A, P_{PR}, P_{OA}, P_{AA} , P_{APR}$ and $P_{OPR}(BKG)$ resulting from IRTS-NKCEM, IRTS-NKLCMV and SaaFormer for Indian Pines dataset.}
	\label{tab:ip}
		\resizebox{.45\textwidth}{!}{
			\begin{tabular}{c|cc|cc|cc}
				\hline \hline
				\multirow{2}{*}{Class} & \multicolumn{2}{c|}{IRTS-NKCEM} & \multicolumn{2}{c|}{IRTS-NKLCMV}    & \multicolumn{2}{c}{SaaFormer}      \\
				& $P_A(C_i)$     & $P_{PR}(C_i)$  & $P_A(C_i)$       & $P_{PR}(C_i)$    & $P_A(C_i)$       & $P_{PR}(C_i)$   \\ \hline
				1                      & $\bold{100}$   & $\bold{100}$   & $\bold{100}$     & $\bold{100}$     & $\bold{100}$     & $\bold{100}$    \\
				2                      & $\bold{98.96}$ & 99.48          & 98.40            & 99.37            & 98.28            & $\bold{99.71}$  \\
				3                      & $\bold{99.81}$ & 99.24          & 99.71            & $\bold{99.43}$   & 99.75            & 99.00           \\
				4                      & 99.92          & 99.92          & $\bold{100}$     & 99.58            & $\bold{100}$     & $\bold{100}$    \\
				5                      & $\bold{98.72}$ & 99.84          & 99.42            & 99.10            & 99.56            & $\bold{100}$    \\
				6                      & 99.94          & 99.43          & 99.97            & $\bold{99.81}$   & $\bold{100}$     & 99.71           \\
				7                      & $\bold{100}$   & 97.93          & $\bold{100}$     & 97.93            & $\bold{100}$     & $\bold{100}$    \\
				8                      & $\bold{100}$   & $\bold{100}$   & $\bold{100}$     & $\bold{100}$     & $\bold{100}$     & $\bold{100}$    \\
				9                      & $\bold{100}$   & $\bold{100}$   & $\bold{100}$     & 99.05            & $\bold{100}$     & $\bold{100}$    \\
				10                     & 99.49          & 98.72          & 99.40            & 99.15            & $\bold{99.57}$   & $\bold{99.36}$  \\
				11                     & 99.12          & $\bold{99.67}$ & 99.49            & 99.47            & $\bold{99.79}$   & 99.50           \\
				12                     & 99.39          & 99.53          & 99.33            & 99.86            & $\bold{99.64}$   & $\bold{100}$    \\
				13                     & $\bold{100}$   & 99.71          & 99.90            & $\bold{100}$     & $\bold{100}$     & $\bold{100}$    \\
				14                     & $\bold{100}$   & 99.56          & $\bold{100}$     & 99.86            & 99.27            & $\bold{99.84}$  \\
				15                     & $\bold{100}$   & 99.79          & $\bold{100}$     & 99.95            & $\bold{100}$     & 96.70           \\
				16                     & $\bold{100}$   & 96.71          & 99.78            & 96.72            & $\bold{100}$     & 96.88           \\ \hline
				$P_{OA}$               & \multicolumn{2}{c|}{99.48}      & \multicolumn{2}{c|}{99.50}          & \multicolumn{2}{c}{$\bold{99.51}$} \\
				$P_{AA}$               & \multicolumn{2}{c|}{99.71}      & \multicolumn{2}{c|}{99.71}          & \multicolumn{2}{c}{$\bold{99.74}$} \\
				$P_{APR}$              & \multicolumn{2}{c|}{99.34}      & \multicolumn{2}{c|}{99.32}          & \multicolumn{2}{c}{$\bold{99.42}$} \\
				$P_{OPR}(BKG)$         & \multicolumn{2}{c|}{48.49}      & \multicolumn{2}{c|}{$\bold{48.50}$} & \multicolumn{2}{c}{47.31}           \\ \hline \hline
			\end{tabular}
}
\end{table}

\begin{table}[htbp]
	\centering
	\caption{Averaged $P_A, P_{PR}, P_{OA}, P_{AA} , P_{APR}$ and $P_{OPR}(BKG)$ resulting from IRTS-NKCEM, IRTS-NKLCMV and SaaFormer for Salinas dataset.}
	\label{tab:sa}
		\resizebox{.45\textwidth}{!}{
			\begin{tabular}{c|cc|cc|cc}
				\hline \hline
				\multirow{2}{*}{Class} & \multicolumn{2}{c|}{IRTS-NKCEM}     & \multicolumn{2}{c|}{IRTS-NKLCMV}    & \multicolumn{2}{c}{SaaFormer}  \\
				& $P_A(C_i)$       & $P_{PR}(C_i)$    & $P_A(C_i)$        & $P_{PR}(C_i)$   & $P_A(C_i)$     & $P_{PR}(C_i)$ \\ \hline
				1                      & $\bold{100}$     & $\bold{100}$     & $\bold{100}$      & $\bold{100}$    & $\bold{100}$   & $\bold{100}$  \\
				2                      & $\bold{100}$     & $\bold{100}$     & $\bold{100}$      & $\bold{100}$    & $\bold{100}$   & $\bold{100}$  \\
				3                      & $\bold{100}$     & $\bold{100}$     & $\bold{100}$      & $\bold{100}$    & $\bold{100}$   & $\bold{100}$  \\
				4                      & 99.80            & $\bold{99.39}$   & 99.94             & 98.94           & $\bold{100}$   & 99.27         \\
				5                      & $\bold{99.75}$   & 99.90            & 99.49             & 99.97           & 99.62          & $\bold{100}$  \\
				6                      & 99.99            & $\bold{100}$     & 99.98             & 99.99           & $\bold{100}$   & $\bold{100}$  \\
				7                      & $\bold{100}$     & 99.99            & $\bold{100}$      & 99.98           & $\bold{100}$   & $\bold{100}$  \\
				8                      & 97.89            & 98.17            & $\bold{98.37}$    & $\bold{98.71}$  & 94.36          & 97.92         \\
				9                      & $\bold{100}$     & 99.96            & $\bold{100}$      & $\bold{99.98}$  & $\bold{100}$   & 99.87         \\
				10                     & 99.91            & $\bold{99.79}$   & $\bold{99.92}$    & 99.75           & 99.75          & 99.54         \\
				11                     & $\bold{100}$     & $\bold{100}$     & $\bold{100}$      & $\bold{100}$    & $\bold{100}$   & $\bold{100}$  \\
				12                     & $\bold{100}$     & $\bold{100}$     & 99.99             & $\bold{100}$    & 99.95          & $\bold{100}$  \\
				13                     & 99.98            & $\bold{100}$     & 99.96             & 99.98           & $\bold{100}$   & 99.89         \\
				14                     & 99.87            & $\bold{99.96}$   & 99.85             & 99.81           & $\bold{99.90}$ & 99.90         \\
				15                     & 97.13            & 96.81            & $\bold{97.98}$    & $\bold{97.60}$  & 96.89          & 91.89         \\
				16                     & $\bold{99.99}$   & $\bold{100}$     & 99.98             & $\bold{100}$    & 99.94          & $\bold{100}$  \\ \hline
				$P_{OA}$               & \multicolumn{2}{c|}{99.15}          & \multicolumn{2}{c|}{$\bold{99.35}$} & \multicolumn{2}{c}{99.14}      \\
				$P_{AA}$               & \multicolumn{2}{c|}{99.65}          & \multicolumn{2}{c|}{$\bold{99.72}$} & \multicolumn{2}{c}{99.68}      \\
				$P_{APR}$              & \multicolumn{2}{c|}{99.62} & \multicolumn{2}{c|}{$\bold{99.67}$}          & \multicolumn{2}{c}{99.59}      \\
				$P_{OPR}(BKG)$         & \multicolumn{2}{c|}{48.31}          & \multicolumn{2}{c|}{$\bold{48.40}$} & \multicolumn{2}{c}{47.97}      \\ \hline \hline
			\end{tabular}
}
\end{table}

\begin{figure}[htbp]
	\centering
	\begin{subfigure}{0.4\textwidth}
		{\includegraphics[width=\textwidth]{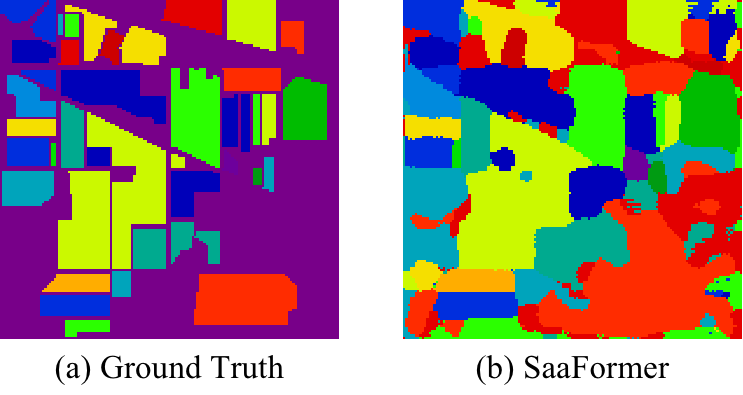}}
	\end{subfigure}\\ \vspace{2 mm} 
	\caption{Final binary classification maps for Indian Pines dataset produced by SaaFormer.}
	\label{fig:ip_bkg1}
\end{figure}

\begin{figure}[htbp]
	\centering
	\begin{subfigure}{0.4\textwidth}
		{\includegraphics[width=\textwidth]{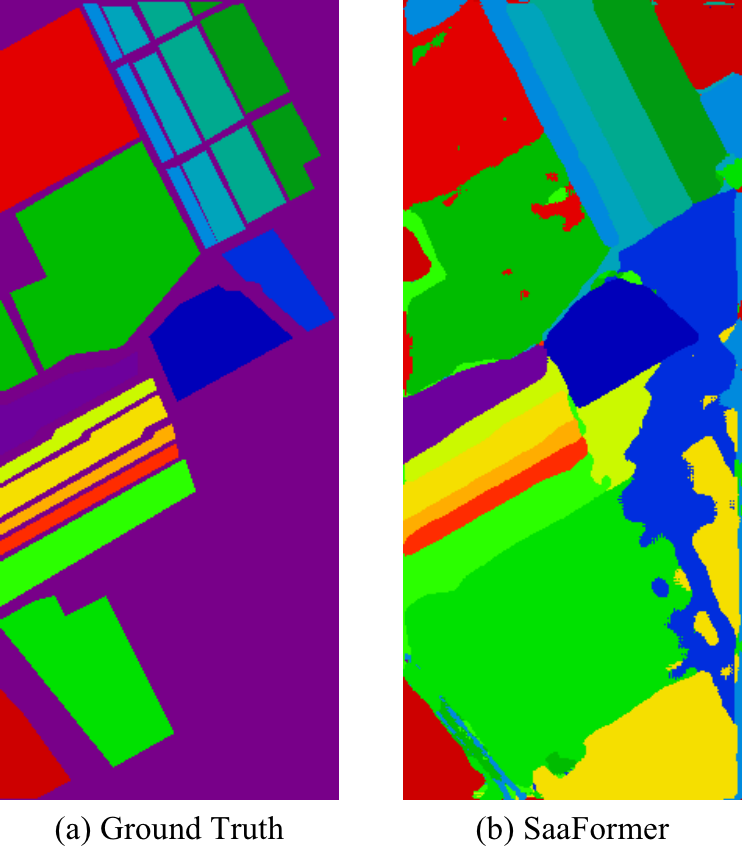}}
	\end{subfigure}\\ \vspace{2 mm} 
	\caption{Final binary classification maps for Salinas dataset produced by SaaFormer.}
	\label{fig:sa_bkg1}
\end{figure}

\begin{table}[htbp]
	\centering
	\caption{Averaged $P_A, P_{PR}, P_{OA}, P_{AA} , P_{APR}$ and $P_{OPR}$(BKG) resulting from IRTS-NKCEM, IRTS-NKLCMV and SaaFormer for PaviaU dataset.}
	\label{tab:pavia}
		\resizebox{.45\textwidth}{!}{
			\begin{tabular}{c|cc|cc|cc}
				\hline \hline
				\multirow{2}{*}{Class} & \multicolumn{2}{c|}{IRTS-NKCEM} & \multicolumn{2}{c|}{IRTS-NKLCMV}    & \multicolumn{2}{c}{SaaFormer}      \\
				& $P_A(C_i)$     & $P_{PR}(C_i)$  & $P_A(C_i)$        & $P_{PR}(C_i)$   & $P_A(C_i)$       & $P_{PR}(C_i)$   \\ \hline
				1                      & 98.86          & 99.51          & 98.73             & 99.50           & $\mathbf{98.87}$   & $\mathbf{99.63}$  \\
				2                      & 99.37          & 99.67          & 99.52             & 99.77           & $\mathbf{99.78}$   & $\mathbf{99.93}$  \\
				3                      & 94.75          & $\mathbf{98.88}$ & 95.93             & 98.43           & $\mathbf{96.18}$   & 96.75           \\
				4                      & 98.53          & 98.49          & $\mathbf{98.90}$    & 98.22           & 98.53            & $\mathbf{99.60}$  \\
				5                      & $\mathbf{100}$   & 99.99          & $\mathbf{100}$      & 99.99           & $\mathbf{100}$     & $\mathbf{100}$    \\
				6                      & 99.68          & 98.52          & 99.73             & $\mathbf{99.31}$  & $\mathbf{99.80}$   & 99.08           \\
				7                      & $\mathbf{99.79}$ & $\mathbf{100}$   & $\mathbf{99.79}$    & $\mathbf{100}$    & $\mathbf{100}$     & 97.70           \\
				8                      & $\mathbf{98.63}$ & 95.31          & 98.42             & 95.57           & 98.59            & $\mathbf{97.15}$  \\
				9                      & 99.89          & $\mathbf{100}$   & 99.91             & 99.96           & $\mathbf{100}$     & 99.89           \\ \hline
				$P_{OA}$               & \multicolumn{2}{c|}{99.02}      & \multicolumn{2}{c|}{99.14}          & \multicolumn{2}{c}{$\bold{99.29}$} \\
				$P_{AA}$               & \multicolumn{2}{c|}{98.83}      & \multicolumn{2}{c|}{98.99}          & \multicolumn{2}{c}{$\bold{99.40}$} \\
				$P_{APR}$              & \multicolumn{2}{c|}{98.93}      & \multicolumn{2}{c|}{98.97} & \multicolumn{2}{c}{$\bold{99.07}$}          \\
				$P_{OPR}(BKG)$         & \multicolumn{2}{c|}{20.44}      & \multicolumn{2}{c|}{$\mathbf{20.45}$}          & \multicolumn{2}{c}{20.37}          \\ \hline \hline
			\end{tabular}
	}
\end{table}

\begin{table}[htbp]
	\centering
	\caption{Averaged $P_A, P_{PR}, P_{OA}, P_{AA} , P_{APR}$ and $P_{OPR}(BKG)$ resulting from IRTS-NKCEM, IRTS-NKLCMV and SaaFormer for Houston dataset.}
	\label{tab:hou}
		\resizebox{.45\textwidth}{!}{
			\begin{tabular}{c|cc|cc|cc}
				\hline \hline
				\multirow{2}{*}{Class} & \multicolumn{2}{c|}{IRTS-NKCEM}     & \multicolumn{2}{c|}{IRTS-NKLCMV} & \multicolumn{2}{c}{SaaFormer}      \\
				& $P_A(C_i)$       & $P_{PR}(C_i)$    & $P_A(C_i)$      & $P_{PR}(C_i)$  & $P_A(C_i)$       & $P_{PR}(C_i)$   \\ \hline
				1                      & 99.45            & 98.85            & 99.18           & $\bold{99.24}$ & $\bold{99.83}$   & 98.12           \\
				2                      & 99.31            & $\bold{99.88}$   & $\bold{99.49}$  & 99.81          & 98.33            & 99.66           \\
				3                      & 99.37            & $\bold{100}$     & $\bold{99.92}$  & 99.97          & 99.84            & $\bold{100}$    \\
				4                      & $\bold{99.92}$   & 99.46            & 99.90           & 99.24          & 99.66            & $\bold{99.92}$  \\
				5                      & 99.95            & 99.92            & 99.44           & $\bold{100}$   & $\bold{100}$     & $\bold{100}$    \\
				6                      & 93.51            & $\bold{100}$     & 92.10           & $\bold{100}$   & $\bold{100}$     & $\bold{100}$    \\
				7                      & 98.78            & 97.22            & $\bold{99.10}$  & 97.50          & $\bold{99.10}$   & $\bold{98.29}$  \\
				8                      & $\bold{98.45}$   & 98.27            & 97.46           & 99.31          & 96.73            & $\bold{99.91}$  \\
				9                      & $\bold{97.79}$   & 98.09            & 97.73           & 96.19          & 97.08            & $\bold{98.81}$  \\
				10                     & $\bold{99.91}$   & $\bold{98.85}$   & 99.65           & 98.12          & 98.89            & 95.88           \\
				11                     & 98.21            & $\bold{98.72}$   & 97.61           & 97.61          & $\bold{98.31}$   & 97.90           \\
				12                     & $\bold{98.82}$   & 98.03            & 97.27           & 98.06          & 98.73            & $\bold{99.57}$  \\
				13                     & $\bold{96.39}$   & $\bold{96.98}$   & 96.04           & 96.89          & 95.47            & 92.81           \\
				14                     & 98.87            & $\bold{100}$     & $\bold{100}$    & $\bold{100}$   & $\bold{100}$     & $\bold{100}$    \\
				15                     & 99.65            & 99.62            & 99.97           & $\bold{99.92}$ & $\bold{100}$     & 99.35           \\ \hline
				$P_{OA}$               & \multicolumn{2}{c|}{98.79} & \multicolumn{2}{c|}{98.59}       & \multicolumn{2}{c}{$\bold{99.04}$}          \\
				$P_{AA}$               & \multicolumn{2}{c|}{98.50}          & \multicolumn{2}{c|}{98.33}       & \multicolumn{2}{c}{$\bold{99.19}$} \\
				$P_{APR}$              & \multicolumn{2}{c|}{98.92} & \multicolumn{2}{c|}{98.79}       & \multicolumn{2}{c}{$\bold{99.09}$}          \\
				$P_{OPR}(BKG)$         & \multicolumn{2}{c|}{$\bold{2.57}$}  & \multicolumn{2}{c|}{2.56}        & \multicolumn{2}{c}{2.13}           \\ \hline \hline
			\end{tabular}
	}
\end{table}

\begin{figure}[htbp]
	\centering
	\begin{subfigure}{0.4\textwidth}
		{\includegraphics[width=\textwidth]{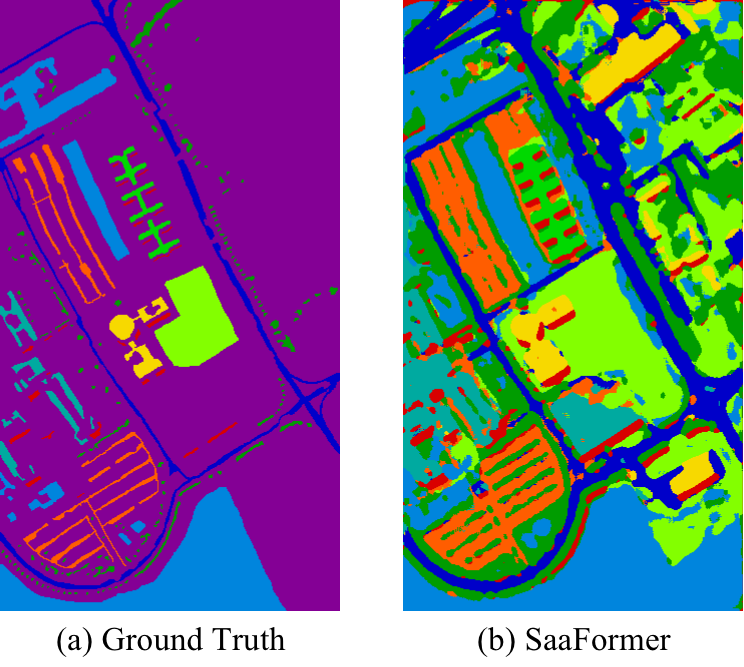}}
	\end{subfigure}\\ \vspace{2 mm} 
	\caption{Final binary classification maps for PaviaU dataset produced by SaaFormer.}
	\label{fig:pu_bkg1}
\end{figure}

\begin{figure}[htbp]
	\centering
	\begin{subfigure}{0.48\textwidth}
		{\includegraphics[width=\textwidth]{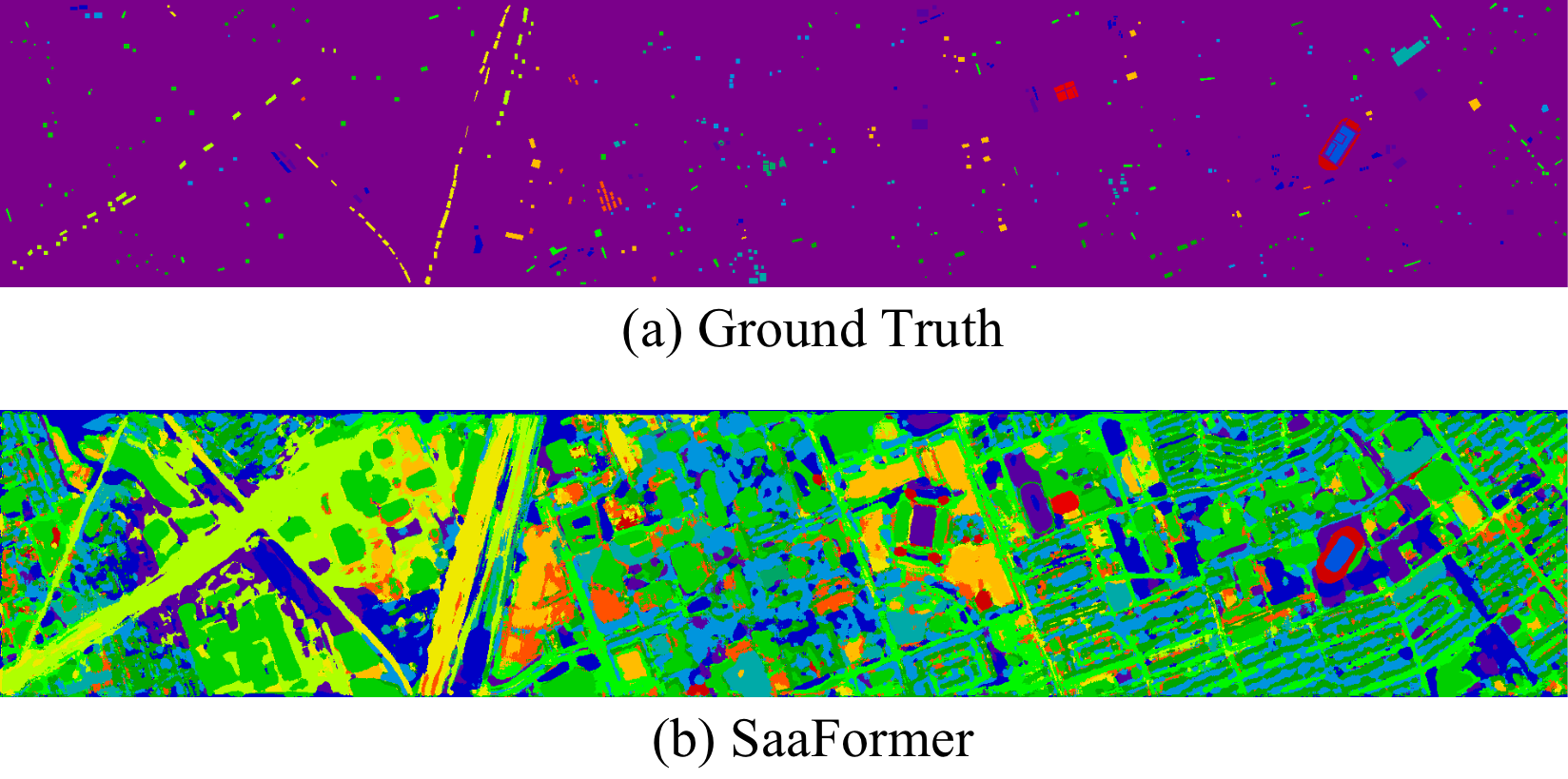}}
	\end{subfigure}\\ \vspace{2 mm} 
	\caption{Final binary classification maps for Houston dataset produced by SaaFormer.}
	\label{fig:hou_bkg1}
\end{figure}

Fig. \ref{fig:ip_bkg1}-\ref{fig:hou_bkg1} show the classification maps produced by SaaFormer on the four datasets \cite{ref28}. Meanwhile, Table \ref{tab:ip}-\ref{tab:hou} summarize their averaged $P_A, P_{PR}, P_{OA}, P_{AA}, P_{APR}$ and $P_{OPR}$(BKG). At the same sampling rate, our model outperforms IRTS-NKCEM and IRTS-NKLCMV on Indian Pines, PaviaU, and Houston datasets for $P_{OA}, P_{AA}$,  and $P_{APR}$. On the Salinas dataset, SaaFormer has completable performance with IRTS-NKCEM but slightly worse than IRTS-NKLCMV.



When computing $P_{OPR}$(BKG) values, it is essential to consider that the training set comprises labeled samples of known classes but excludes background samples. However, the test set includes background samples. Consequently, the model misclassifies all background samples as known labels, leading to the $P_{(OPR)}$(BKG)  value decreases when the size of the background samples increases. Theoretically, by Eq. \ref{eq:opr} and Eq. \ref{eq:apr}, $P_{OPR}$(BKG) is proportional to $P_{APR}$. It is feasible to calculate the theoretical upper bound of the $P_{OPR}$(BKG) for these four datasets, i.e. $P_{OPR}(BKG)_{max}=\frac{{n}_{test}}{{n}_{test}+n_{BKG}}$, where $n_{BKG}$ represents the total number of background samples. Tab \ref{tab:pr-rate} displays the theoretical upper bounds of the $P_{OPR}$(BKG) values for these datasets alongside the results obtained by the model. Examination of Tab \ref{tab:pr-rate} reveals that although the $P_{OPR}$(BKG) values are relatively low, they are remarkably close to their theoretical upper bounds.

\begin{table}[htbp]
	\centering
	\caption{$P_{OPR}(BKG)_{max}$ and $P_{OPR}(BKG)$ for the four datasets produced by SaaFormer.}
	\label{tab:pr-rate}
		\resizebox{.45\textwidth}{!}{
	\begin{tabular}{c|cccc}
		\hline \hline
		& IndianPines & Salinas & PaviaU & Houston \\ \hline
		$P_{PR}(BKG)_{max}$(\%)                     & 47.50       & 48.49   & 20.43  & 2.15    \\
		$P_{PR}(BKG)$(\%)                           & 47.31       & 47.97   & 20.37  & 2.13    \\
		$\frac{P_{PR}(BKG)}{P_{PR}(BKG)_{max}}$(\%) & 99.60       & 98.93   & 99.61  & 99.07   \\ \hline \hline
	\end{tabular}
}
\end{table}

\section{Conclusion}

HSIs are often represented as data cubes with both spatial and spectral dimensions, typically viewed sequentially along the spectral axis. Deep neural networks have achieved high accuracy on these images even with limited training data, primarily under the commonly used random sampling evaluation method. However, we found significant overlap between the training and validation datasets, with classification accuracy directly related to this overlap. This suggests that random sampling overestimates the model's generalization ability. To address this, we propose non-overlapping data partitioning methods for more accurate generalization assessment. Our experiments show that popular models, including 2D-CNN, 3D-CNN, RNN, and transformer-based models, experience notable performance drops with non-overlapping partitions. In response, we introduce SaaFormer, a transformer-based model that emphasizes spectral characteristics through axial aggregation attention and multi-level spectral extraction. SaaFormer maintains similar performance under random sampling but significantly outperforms other methods with alternative sampling strategies, as demonstrated on five public datasets.


Our future work will focus on developing more generalizable and interpretable models that integrate deep learning with traditional methodologies. The primary emphasis will be on enhancing model interpretability and accurately and efficiently determining the optimal number of layers in multi-level spectral extraction. To achieve this, we will incorporate the physical characteristics of spectral bands and prior knowledge of hyperspectral images to construct models with stronger interpretative capabilities. Additionally, we will explore the embedding of interpretative knowledge as a means to develop and evaluate more advanced models.
	


\section*{Acknowledgements}
The authors would like to thank the support by National Natural Science Foundation of China (U21B2075, 12401557, 11971131, 11871133), Basic Science Center Project of the National Natural Science Foundation of China (62388101), Natural Science Foundation of Heilongjiang Province of China (ZD2022A001) , Fundamental Research Funds for the Central Universities (2022FRFK060020, 2022FRFK060031) and the Hyperspectral Image Analysis group at the University of Houston for providing the CASI University of Houston datasets and the IEEE GRSS DFC2013.

\vspace{11pt}
\begin{IEEEbiography}[{\includegraphics[width=1in,height=1.25in,clip,keepaspectratio]{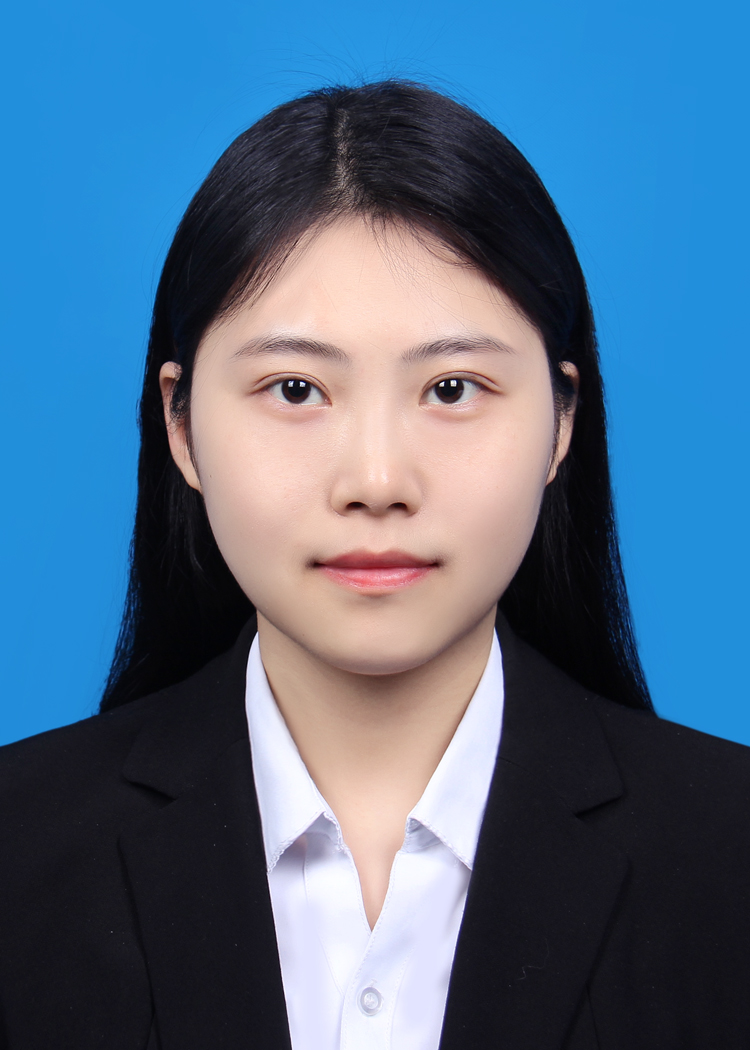}}]{Enzhe Zhao}
	(Student Member IEEE) received the Master degree  from  Harbin Institute of Technology, Harbin, China, in 2022. She is currently working toward the Ph.D. degree with the School of Mathematics, Harbin Institute of Technology, Harbin, China. 
	
	Her research interests include image processing and analysis, hyperspectral remote sensing and deep learning.
\end{IEEEbiography}

\begin{IEEEbiography}[{\includegraphics[width=1in,height=1.25in,clip,keepaspectratio]{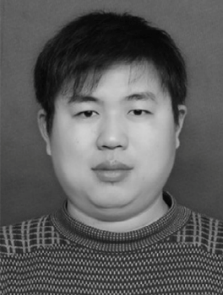}}]{Zhichang Guo}
	received his Ph.D. degree from Jilin University, Changchun, China, in 2010. He is currently a professor at School of Mathematics, Harbin Institute of Technology, Harbin, China. 
	
	His research interests include partial differential equations, nonlinear analysis, mathematical methods in image analysis, and deep learning.
\end{IEEEbiography}

\begin{IEEEbiography}[{\includegraphics[width=1in,height=1.25in,clip,keepaspectratio]{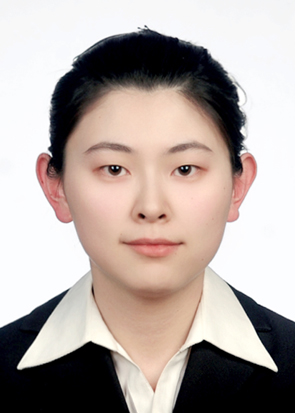}}]{Shengzhu Shi}
	received the Ph.D. degree from Harbin Institute of Technology, Harbin, China, in 2022.
	
	She is currently a lecturer in the School of Mathematics, Harbin Institute of Technology, Harbin, China. Her research interests include numerical methods for partial differential equations, image processing, deep learning and uncertainty quantification.
\end{IEEEbiography}

\begin{IEEEbiography}[{\includegraphics[width=1in,height=1.25in,clip,keepaspectratio]{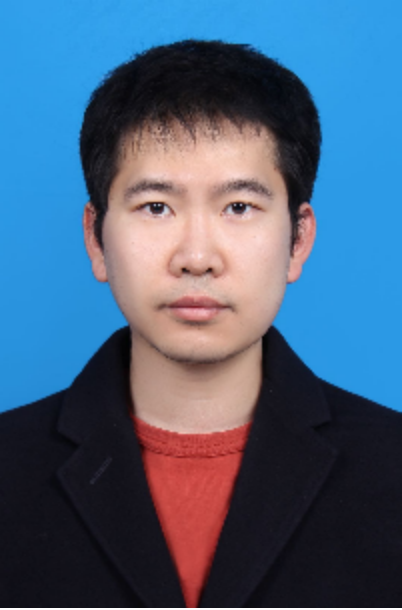}}]{Yao Li}
	received the Ph.D. degree in system engineering from the University of Illinois Urbana-Champaign, Champaign, IL, USA, in 2020.
	
	He is currently an Assistant Professor with the School of Mathematics, Harbin Institute of Technology, Harbin, China. His research interests include adversarial attacks, image processing, and electroencephalogram (EEG) processing.
\end{IEEEbiography}

\vspace{-110mm}

\begin{IEEEbiography}[{\includegraphics[width=1in,height=1.25in,clip,keepaspectratio]{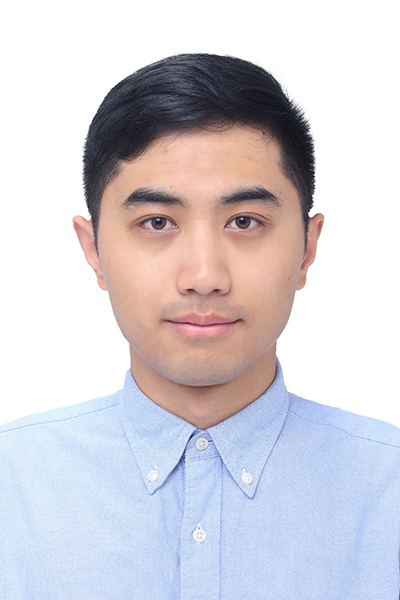}}]{\textbf{Jia Li}}
	received the Ph.D. degree from Harbin Institute of Technology, Harbin, China, in 2020. 
	
	He is currently working as assistant professor with the School of Mathematics, Harbin Institute of Technology, Harbin. His research interests include numerical methods for partial differential equation and deep learning.
\end{IEEEbiography}

\vspace{-110mm}

\begin{IEEEbiography}[{\includegraphics[width=1in,height=1.25in,clip,keepaspectratio]{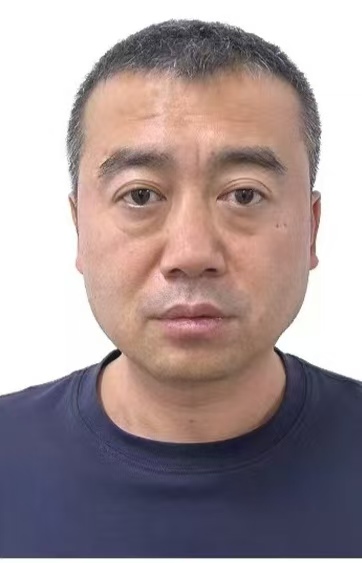}}]{\textbf{Dazhi Zhang}}
	received the Ph.D. degree from Jilin University, Changchun, China, in 2006. He is currently a Professor with the School of Mathematics, Harbin Institute of Technology, Harbin, China. 
	
	His research interests include partial Numerical solution of partial differential equations, mathematical methods in image analysis, deep learning and remote sensing.
\end{IEEEbiography}


\begin{thebibliography}{}
\providecommand{\url}[1]{#1}
\csname url@samestyle\endcsname
\providecommand{\newblock}{\relax}
\providecommand{\bibinfo}[2]{#2}
\providecommand{\BIBentrySTDinterwordspacing}{\spaceskip=0pt\relax}
\providecommand{\BIBentryALTinterwordstretchfactor}{4}
\providecommand{\BIBentryALTinterwordspacing}{\spaceskip=\fontdimen2\font plus
\BIBentryALTinterwordstretchfactor\fontdimen3\font minus
  \fontdimen4\font\relax}
\providecommand{\BIBforeignlanguage}[2]{{%
\expandafter\ifx\csname l@#1\endcsname\relax
\typeout{** WARNING: IEEEtran.bst: No hyphenation pattern has been}%
\typeout{** loaded for the language `#1'. Using the pattern for}%
\typeout{** the default language instead.}%
\else
\language=\csname l@#1\endcsname
\fi
#2}}
\providecommand{\BIBdecl}{\relax}
\BIBdecl

\end{thebibliography}


\begin{thebibliography}{1}
\bibliographystyle{IEEEtran}


\bibitem{refh1} M. Ahmad \textit{et al.}, "Spatial prior fuzziness pool-based interactive classification of hyperspectral images," \textit{Remote Sens.}, vol. 11, no. 9, pp. 1-19, May. 2019.


\bibitem{refs1} D. Hong \textit{et al.}, "Interpretable hyperspectral artificial intelligence: When nonconvex modeling meets hyperspectral remote sensing," \textit{IEEE Geosci.Remote Sens. Mag.}, vol. 9, no. 2, pp. 52–87, Jun. 2021.


\bibitem{refh2} M. Ahmad \textit{et al.}, "Hyperspectral image classification-traditional to deep models: A survey for future prospects," \textit{IEEE J. Sel. Topics Appl. Earth Observ. Remote Sens.}, vol. 15, pp. 968-999, 2022.

\bibitem{ref25} S. Zhong, C.-I Chang and Y. Zhang, "Spectral-spatial feedback close
network system for hyperspectral image classification," \textit{IEEE Trans. Geosci. Remote Sens.}, vol. 57, no. 10, pp. 8131-8143, Oct. 2019.

\bibitem{refz1} Richards J A, Jia X, "Using suitable neighbors to augment the training set in hyperspectral maximum likelihood classification," \textit{IEEE Trans. Geosci. Remote Sens.}, vol. 5, no. 4, pp. 774-777, 2008.

\bibitem{refz2} Peng J, Li L, Tang Y Y, "Maximum likelihood estimation-based joint sparse representation for the classification of hyperspectral remote sensing images," \textit{IEEE Trans. Neural Netw. Learn. Syst.}, vol. 30, no. 6, pp. 1790-1802, 2018.

\bibitem{refz3} Kuching S, "The performance of maximum likelihood, spectral angle mapper, neural network and decision tree classifiers in hyperspectral image analysis," \textit{Journal of Computer Science}, vol. 3, no. 6, pp. 419-423, 2007.


\bibitem{ref1} Qingwang Wang, Yanfeng Gu, and Devis Tuia, "Discriminative multiple kernel learning for hyperspectral image classification," \textit{IEEE Trans. Geosci. Remote Sens.}, vol. 54, no. 7, pp. 3912–3927, 2016.


\bibitem{reff1} C. Cariou and K. Chehdi, " A new k-nearest neighbor density-based clustering method and its application to hyperspectral images, " \textit{Proc. IEEE Int. Geosci. Remote Sens. Symp. (IGARSS)}, pp. 6161-6164, Jul. 2016.

\bibitem{reff2} Y. E. SahIn, S. Arisoy and K. Kayabol, "Anomaly detection with Bayesian Gauss background model in hyperspectral images," \textit{Proc. 26th Signal Process. Commun. Appl. Conf. (SIU)}, pp. 1-4, May 2018.

\bibitem{ref4} M. Khodadadzadeh, Jun Li, A. Plaza, and J.M. Bioucas-Dias, "A subspace-based multinomial logistic regression for hyperspectral image classification," \textit{IEEE Geosci. Remote Sens. Lett.}, vol. 11, no. 12, pp. 2105–2109, Dec 2014.

\bibitem{reff3} Y.-N. Chen, T. Thaipisutikul, C.-C. Han, T.-J. Liu and K.-C. Fan, "Feature line embedding based on support vector machine for hyperspectral image classification," \textit{Remote Sens.}, vol. 13, no. 1, pp. 130, Jan. 2021.


\bibitem{reff4} G. Licciardi, P. R. Marpu, J. Chanussot and J. A. Benediktsson, "Linear versus nonlinear PCA for the classification of hyperspectral data based on the extended morphological profiles," \textit{IEEE Trans. Geosci. Remote Sens.}, vol. 9, no. 3, pp. 447-451, May 2012.


\bibitem{reff5} A. Villa, J. A. Benediktsson, J. Chanussot and C. Jutten, "Hyperspectral image classification with independent component discriminant analysis," \textit{IEEE Trans. Geosci. Remote Sens.}, vol. 49, no. 12, pp. 4865-4876, Dec. 2011.

\bibitem{reff6} Q. Ye, J. Yang, F. Liu, C. Zhao, N. Ye and T. Yin, "$L_1$-norm distance linear discriminant analysis based on an effective iterative algorithm," \textit{IEEE Trans. Circuits Syst. Video Technol.}, vol. 28, no. 1, pp. 114-129, Jan. 2018.

\bibitem{reff7} J. A. Benediktsson, J. A. Palmason and J. R. Sveinsson, "Classification of hyperspectral data from urban areas based on extended morphological profiles," \textit{IEEE Trans. Geosci. Remote Sens.}, vol. 43, no. 3, pp. 480-491, Mar. 2005.


\bibitem{reff8} M. M. Dalla, A. Villa, J. A. Benediktsson, J. Chanussot and L. Bruzzone, "Classification of hyperspectral images by using extended morphological attribute profiles and independent component analysis," \textit{IEEE Trans. Geosci. Remote Sens.}, vol. 8, no. 3, pp. 542-546, Dec. 2011.


\bibitem{ref7} Yushi Chen, Zhouhan Lin, Xing Zhao, Gang Wang, and Yanfeng Gu, "Deep learning-based classification of hyperspectral data," \textit{IEEE J. Sel. Topics Appl. Earth Observ. Remote Sens.}, vol. 7, no .6, pp. 2094–2107, June 2014.

\bibitem{ref8} Wei Hu, Yangyu Huang, Li Wei, Fan Zhang, and Hengchao Li,  "Deep convolutional neural networks for hyperspectral image classification," \textit{Journal of Sensors}, 501:258619, 2015.

\bibitem{ref13} Jun Yue, Wenzhi Zhao, Shanjun Mao, and Hui Liu, "Spectral-spatial classification of hyperspectral images using deep convolutional neural networks," \textit{Remote Sens. Lett.}, vol. 6, no. 6, pp. 468–477, 2015.

\bibitem{ref14} Heming Liang and Qi Li, "Hyperspectral imagery classification using sparse representations of convolutional neural network features," \textit{Remote Sens.}, vol. 8, no. 2, 2016.

\bibitem{ref15} Y. Chen, H. Jiang, C. Li, X. Jia, and P. Ghamisi, "Deep feature extraction and classification of hyperspectral images based on convolutional neural networks," \textit{IEEE Trans. Geosci. Remote Sens.}, vol. 54, no. 10, pp. 6232–6251, Oct. 2016.

\bibitem{refh4} H. Gao, Y. Yang, C. Li, H. Zhou, and X. Qu, "Joint alternate small convolution and feature reuse for hyperspectral image classification," \textit{ISPRS Int. J. Geo Inf.}, vol. 7, no. 9, p. 349, 2018.

\bibitem{ref16} Li Y, Zhang H, Shen Q, "Spectral-spatial classification of hyperspectral imagery with 3D convolutional neural network," \textit{Remote Sens.}, vol. 9, no. 1, pp. 67, 2017.

\bibitem{ref17} M. He, B. Li and H. Chen, "Multi-scale 3D deep convolutional neural network for hyperspectral image classification," \textit{2017 IEEE International Conference on Image Processing. (ICIP)}, Beijing, China, pp. 3904-3908, 2017.


\bibitem{ref9} L. Mou, P. Ghamisi, and X. X. Zhu, "Deep recurrent neural networks for hyperspectral image classification," \textit{IEEE Trans. Geosci. Remote Sens.}, vol. 55, no. 7, pp. 3639–3655, July. 2017.


\bibitem{reff9} S. K. Roy, G. Krishna, S. R. Dubey and B. B. Chaudhuri, "HybridSN: Exploring 3-D–2-D CNN feature hierarchy for hyperspectral image classification," \textit{IEEE Trans. Geosci. Remote Sens.}, vol. 17, no. 2, pp. 277-281, Feb. 2020.



\bibitem{refq1} Z. Zhong, J. Li, Z. Luo and M. Chapman, "Spectral-spatial residual network for hyperspectral image classification: A 3D deep learning framework," \textit{IEEE Trans. Geosci. Remote Sens.}, vol. 56, no. 2, pp. 847-858, Aug. 2018.


\bibitem{refq2} M. E. Paoletti, J. M. Haut, R. Fernandez-Beltran, J. Plaza, J. Plaza and F. Pla, "Deep pyramidal residual networks for spectral-spatial hyperspectral image classification," \textit{IEEE Trans. Geosci. Remote Sens.}, vol. 57, no. 2, pp. 740-754, Aug. 2019.

\bibitem{refh8} S. K. Roy, S. Manna, T. Song, and L. Bruzzone, "Attention-based adaptive spectral-spatial kernel ResNet for hyperspectral image classification," \textit{IEEE Trans. Geosci. Remote Sens.}, vol. 
59, no. 9, pp. 7831–7843, Sep. 2021.



\bibitem{reff10}  S. K. Roy, S. R. Dubey, S. Chatterjee and B. B. Chaudhuri, "FuSENet: Fused squeeze- and-excitation network for spectral-spatial hyperspectral image classification," \textit{IET Image Process}., vol. 14, no. 8, pp. 1653-1661, 2020.


\bibitem{reff12} M. E. Paoletti, J. M. Haut, S. K. Roy, and E. M. Hendrix, "Rotation equivariant convolutional neural networks for hyperspectral image classification," \textit{IEEE Access}, vol. 8, pp. 179575–179591, 2020.

\bibitem{reff13} S. K. Roy, P. Kar, D. Hong, X. Wu, A. Plaza and J. Chanussot, "Revisiting deep hyperspectral feature extraction networks via gradient centralized convolution", \textit{IEEE Trans. Geosci. Remote Sens.}, vol. 60, pp. 1-19, 2021.

\bibitem{reff14} S. K. Roy, M. E. Paoletti, J. M. Haut, E. M. T. Hendrix, and A. Plaza, "A new max-min convolutional network for  hyperspectral image classification," in \textit{Proc. 11th Workshop  Hyperspectral Imag. Signal Process.:Evol. Remote Sens.}, pp. 1–5, 2021.


\bibitem{ref18} A. Dosovitskiy, L. Beyer, A. Kolesnikov, D. Weissenborn, X. Zhai, T. Unterthiner, M. Dehghani, M. Minderer, G. Heigold, S. Gelly, \textit{et al.}, "An image is worth 16x16 words: Transformers for image recognition at scale," in \textit{Proc. International Conference on Learning Representations. (ICLR)}, 2021.




\bibitem{ref19} D. Hong \textit{et al.}, "SpectralFormer: Rethinking Hyperspectral Image Classification With Transformers," \textit{IEEE Trans. Geosci. Remote Sens.}, vol. 60, pp. 1-15, 2022.

\bibitem{ref19_1} J. He, L. Zhao, H. Yang, M. Zhang and W. Li, "HSI-BERT: Hyperspectral image classification using the bidirectional encoder representation from transformers," \textit{IEEE Trans. Geosci. Remote Sens.}, vol. 58, no. 1, pp. 165-178, Sep. 2020.



\bibitem{ref19_3} S. K. Roy, A. Deria, C. Shah, J. M. Haut, Q. Du and A. Plaza, "Spectral-Spatial Morphological Attention Transformer for Hyperspectral Image Classification," \textit{IEEE Trans. Geosci. Remote Sens.}, vol. 61, pp. 1-15, 2023.

\bibitem{ref19_4} L. Sun, G. Zhao, Y. Zheng and Z. Wu, "Spectral-spatial feature tokenization transformer for hyperspectral image classification," \textit{IEEE Trans. Geosci. Remote Sens.}, vol. 60, pp. 1-14, 2022.


\bibitem{refh15} X. Wang, Y. Feng, R. Song, Z. Mu, and C. Song, "Multi-attentive hierarchical dense fusion net for fusion classification of hyperspectral and LiDAR data," \textit{Inf. Fusion}, vol. 82, pp. 1-18, Jun. 2022.

\bibitem{refh16} S. K. Roy, A. Deria, D. Hong, B. Rasti, A. Plaza and J. Chanussot, "Multimodal Fusion Transformer for Remote Sensing Image Classification," \textit{IEEE Trans. Geosci. Remote Sens.}, vol. 61, pp. 1-20, 2023.


\bibitem{refh18} J. Xia, N. Yokoya, and A. Iwasaki, "Fusion of hyperspectral and LiDAR data with a novel ensemble classifier," \textit{IEEE Trans. Geosci. Remote Sens.},vol. 15, no. 6, pp. 957–961, Jun. 2018.

\bibitem{refh19} D. Hong, L. Gao, R. Hang, B. Zhang, and J. Chanussot, “Deep encoder-decoder networks for classification of hyperspectral and LiDAR data,” \textit{IEEE Trans. Geosci. Remote Sens.}, vol. 19, pp. 1-5, 2022. 

\bibitem{refh17} D. Hong \textit{et al.}, "More diverse means better: Multimodal deep learning meets remote-sensing imagery classification,” \textit{IEEE Trans. Geosci. Remote Sens.}, vol. 59, no. 5, pp. 4340-4354, May 2021.

\bibitem{refh20} X. Wu, D. Hong and J. Chanussot, "Convolutional Neural Networks for Multimodal Remote Sensing Data Classification," \textit{IEEE Trans. Geosci. Remote Sens.}, vol. 60, pp. 1-10, 2022.



\bibitem{ref20_2} D. Hong, J. Yao, D. Meng, Z. Xu and J. Chanussot, "Multimodal GANs: Toward Crossmodal Hyperspectral-Multispectral Image Segmentation," \textit{IEEE Trans. Geosci. Remote Sens.}, vol. 59, no. 6, pp. 5103-5113, June 2021, doi: 10.1109/TGRS.2020.3020823

\bibitem{ref20_3} D. Hong, B. Zhang, H. Li, Y. Li, J. Yao, C. Li, M. Werner, J. Chanussot, A. Zipf, and X. X. Zhu, "Cross-city matters: A multimodal remote sensing benchmark dataset for cross-city semantic segmentation using high-resolution domain adaptation networks," \textit{Remote Sensing of Environment}, vol. 299, pp. 113856, 2023.

\bibitem{ref20_4} R. Bommasani, D. A. Hudson, E. Adeli, R. Altman, S. Arora, S. von Arx, M. S. Bernstein, J. Bohg, A. Bosselut, E. Brunskill, \textit{et al.}, "On the opportunities and risks of foundation models," 2021, \textit{arXiv:2108.07258}.


\bibitem{ref20_5}D. Hong \textit{et al.}, "SpectralGPT: Spectral Remote Sensing Foundation Model," in \textit{IEEE Transactions on Pattern Analysis and Machine Intelligence}, vol. 46, no. 8, pp. 5227-5244, Aug. 2024, doi: 10.1109/TPAMI.2024.3362475. 


\bibitem{reff15} Wang, Sheng, \textit{et al.}, "Trustworthy remote sensing interpretation: Concepts, technologies, and applications," \textit{ ISPRS J. Photogramm. Remote Sens.}, vol. 209, pp. 150-172, 2024.

\bibitem{reff16} Li, Jiaxin, \textit{et al.}, "Deep learning in multimodal remote sensing data fusion: A comprehensive review," \textit{International Journal of Applied Earth Observation and Geoinformation} 112 (2022): 102926.

\bibitem{ref6_1}  X. Wang, Y. Feng, R. Song, Z. Mu and C. Song, "Multi-attentive hierarchical dense fusion net for fusion classification of hyperspectral and LiDAR data," \textit{Inf. Fusion}, vol. 82, pp. 1-18, Jun. 2022.

\bibitem{reff17} Shi, Wenzhong, Dizhou Guo, and Hua Zhang, "A reliable and adaptive spatiotemporal data fusion method for blending multi-spatiotemporal-resolution satellite images," \textit{Remote Sensing of Environment} 268 (2022): 112770.

\bibitem{reff18} Ghamisi, Pedram, \textit{et al.}, "Multisource and multitemporal data fusion in remote sensing: A comprehensive review of the state of the art," \textit{IEEE Geosci. Remote Sens. Mag.}, vol. 7, no. 1, pp. 6-39, March 2019, doi: 10.1109/MGRS.2018.2890023.

\bibitem{reff19} Kakogeorgiou, Ioannis, and Konstantinos Karantzalos, "Evaluating explainable artificial intelligence methods for multi-label deep learning classification tasks in remote sensing," \textit{International Journal of Applied Earth Observation and Geoinformation} 103 (2021): 102520.

\bibitem{reff20} X. He, Y. Chen, and L. Huang, "Toward a trustworthy classifier with deep CNN: uncertainty estimation meets hyperspectral image." \textit{IEEE Trans. Geosci. Remote Sens.},  vol. 60, pp. 1-15, 2022.


\bibitem{reff21} Q. Wang, C. Yin, H. Song, T. Shen and Y. Gu, "UTFNet: Uncertainty-Guided Trustworthy Fusion Network for RGB-Thermal Semantic Segmentation," \textit{IEEE Geosci. Remote Sens. Lett.}, vol. 20, pp. 1-5, 2023, Art no. 7001205, doi: 10.1109/LGRS.2023.3322452


\bibitem{ref20} Jonathan Ho, Nal Kalchbrenner, Dirk Weissenborn, and Tim Salimans, "Axial attention in multidimensional transformers," 2019, \textit{arXiv:1912.12180}.


\bibitem{ref21} Wang H, Zhu Y, Green B, \textit{et al.}, "Axial-deeplab: Stand-alone axial-attention for panoptic segmentation," \textit{Computer Vision–ECCV 2020: 16th European Conference}, Glasgow, UK, August 23–28, 2020, Proceedings, Part IV. Cham: Springer International Publishing, 2020: 108-126.

\bibitem{ref21a} Valanarasu J M J, Oza P, Hacihaliloglu I, \textit{et al.}, "Medical transformer: Gated axial-attention for medical image segmentation," \textit{Medical Image Computing and Computer Assisted Intervention–MICCAI 2021: 24th International Conference}, Strasbourg, France, September 27–October 1, 2021, Proceedings, Part I 24. Springer International Publishing, 2021: 36-46.

\bibitem{ref21b} Wan, Qiang, \textit{et al.}, "Seaformer: Squeeze-enhanced axial transformer for mobile semantic segmentation," 2023, \textit{arXiv:2301.13156}.

\bibitem{refk1} Liu, Ze, \textit{et al.}, "Swin transformer: Hierarchical vision transformer using shifted windows," in \textit{Proc. ICCV}, 2021, pp. 992-1002.

\bibitem{refk2} Dong, Xiaoyi, \textit{et al.}, "Cswin transformer: A general vision transformer backbone with cross-shaped windows," in \textit{Proc. IEEE Conf. Comput. Vis. Pattern Recognit. (CVPR)}, 2022, pp. 12124-12134.

\bibitem{refk3} Han, Kai, \textit{et al.} "Transformer in transformer," in \textit{Proc. Adv. Neural Inf. Process. Syst.}, 2021, 34, pp. 15908-15919.


\bibitem{ref21c} Ramachandran, Prajit, \textit{et al.}, "Stand-alone self-attention in vision models," in  \textit{Proc. Adv. Neural Inf. Process. Syst.}, 2019, 32.

\bibitem{ref23} G. P. Arada and E. P. Dadios, "Partial fingerprint identification through checkerboard sampling method using ANN," \textit{TENCON 2012 IEEE Region 10 Conference}, Cebu, Philippines, 2012, pp. 1-6, doi: 10.1109/TENCON.2012.6412170.


\bibitem{ref24} C. -I. Chang, C. -C. Liang and P. F. Hu, "Iterative Random Training Sampling Convolutional Neural Network for Hyperspectral Image Classification," \textit{IEEE Trans. Geosci. Remote Sens.}, vol. 61, pp. 1-26, 2023.







%
%
%
%
%
%
%
%
%










%
























%
%
%
%
%






\bibitem{ref22} D. P. Kingma and J. Ba, “Adam: A method for stochastic optimization,” 2014, \textit{arXiv:1412.6980}.





\bibitem{ref26} D. Hong, N. Yokoya, J. Chanussot and X. X. Zhu, "An Augmented Linear Mixing Model to Address Spectral Variability for Hyperspectral Unmixing," \textit{IEEE Trans. Image Process.}, vol. 28, no. 4, pp. 1923-1938, Apr. 2019.





\bibitem{ref28} K. Y. Ma and C. -I. Chang, "Kernel-Based Constrained Energy Minimization for Hyperspectral Mixed Pixel Classification," \textit{IEEE Trans. Geosci. Remote Sens.}, vol. 60, pp. 1-23, 2022, Art no. 5510723.



\end{thebibliography}
\end{document}